\documentclass{article}
\makeatletter
\def\input@path{{iclr_formatting/}}
\makeatother
\usepackage{iclr2027_conference,times}

\usepackage{amsmath,amsfonts,bm}

\def\1{\bm{1}}

\DeclareMathAlphabet{\mathsfit}{\encodingdefault}{\sfdefault}{m}{sl}
\SetMathAlphabet{\mathsfit}{bold}{\encodingdefault}{\sfdefault}{bx}{n}

\usepackage{hyperref}
\usepackage{url}
\usepackage{xcolor}
\usepackage{arydshln}
\usepackage{titletoc}
\usepackage{graphicx}
\usepackage{subcaption}
\usepackage{booktabs}
\usepackage{wrapfig}
\usepackage{float}
\newcommand{\methodname}{\textbf{\textsc{IMC-CLINIC}}}

\title{\methodname{}: Coupled Loss-Informed Newton Iterations for Clipping in Analog In-Memory Computing}

\author{Yung-Chin Chen, Chia-Yu Chen, Naveen Verma}

\author{
Yung-Chin Chen\textsuperscript{1,2} \quad
Chia-Yu Chen\textsuperscript{2} \quad
Naveen Verma\textsuperscript{1,2} \\
\textsuperscript{1}Princeton University, NJ, USA
\qquad
\textsuperscript{2}EnCharge AI, CA, USA
}

\iclrfinalcopy 

\begin{document}

\maketitle

\begin{abstract}
Analog in-memory computing (IMC) offers a promising path toward energy-efficient large language model (LLM) inference by executing matrix multiplications (MatMul) directly within memory arrays in the analog domain.
Its efficiency, however, comes with an additional source of error: limited-precision analog-to-digital converters (ADCs) quantize the accumulated analog partial sums and introduce output-side error, which is a different problem from conventional activation and weight quantization at the MatMul inputs.
Clipping can mitigate both operand and ADC quantization errors by reducing their dynamic ranges, but the optimal clipping factors must jointly balance activation rounding and clipping, weight rounding and clipping, and ADC quantization.
Existing clipping methods, designed for digital quantization, do not explicitly optimize these coupled sources of IMC error. Common approaches rely on costly search-based calibration, leading to suboptimal accuracy and long calibration time.
We introduce \methodname{} (\textbf{C}oupled \textbf{L}oss-\textbf{I}nformed \textbf{N}ewton \textbf{I}terations for \textbf{C}lipping), a clipping calibration framework built around an analytical surrogate for IMC MatMul output error.
The loss surrogate models operand quantization, accumulated clipping-induced bias, and ADC quantization jointly, allowing its gradient and approximate curvature to be evaluated directly from only a small calibration set. \methodname{} then jointly optimizes activation and weight clipping factors using a safeguarded Newton-type method.
We show that \methodname{} reduces analog-IMC MatMul output error by effectively balancing operand and ADC quantization errors. Across multiple models and datasets, it improves average zero-shot accuracy by 6.5--11.5 percentage points over the grid search baseline while reducing the calibration time by 10.0$\times$--12.1$\times$. Moreover, its analytical surrogate closely tracks empirical IMC output error, while its optimizer is fast and certified within 1\% of the global optimum under the loss objective across all projections on two representative models.

\end{abstract}

\section{Introduction}
\label{sec:introduction}

As Large Language Models (LLMs) become increasingly capable, their power consumption has become a major concern. To meet the growing power demand of LLM inference workloads, frontier AI infrastructure is scaling toward gigawatt-level power capacity \citep{OpenAI-NVIDIA-10GW,Anthropic-Build-AI-in-America}. This scale of demand creates substantial economic \citep{Anthropic-Electricity} and sustainability \citep{Cooling-for-LLM} challenges, motivating the development of more energy-efficient inference hardware.

\begin{figure*}[t]
    \centering
    \begin{subfigure}[t]{0.52\textwidth}
        \centering
        \includegraphics[height=3.8cm]{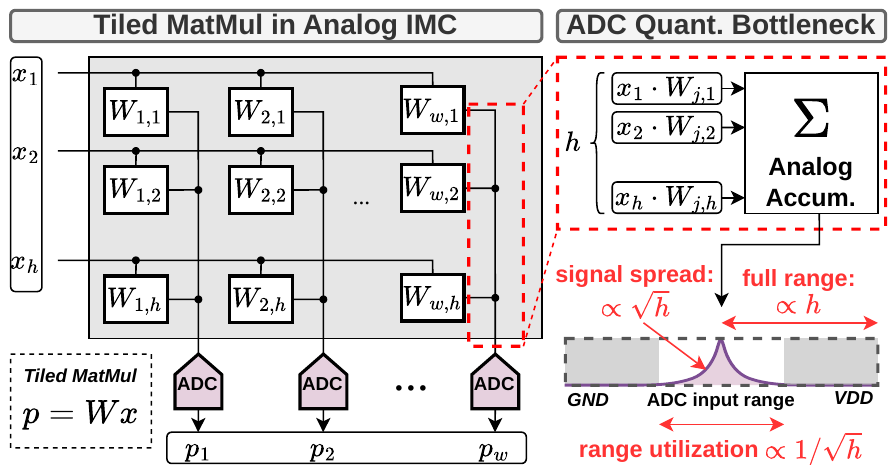}%
        \caption{Analog IMC and ADC range underutilization.}
        \label{fig:imc-intro}
    \end{subfigure}%
    \hfill
    \begin{subfigure}[t]{0.46\textwidth}
        \centering
        \includegraphics[height=3.8cm]{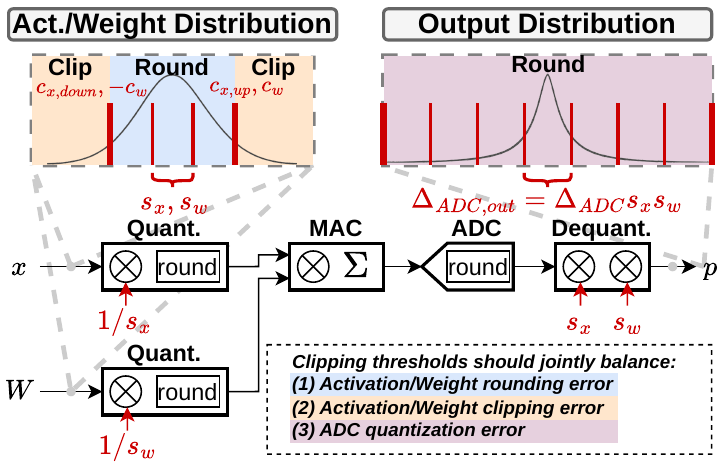}%
        \caption{Coupled operand and ADC quantization errors.}
        \label{fig:motivation}
    \end{subfigure}
    \caption{Motivation for clipping calibration in analog IMC. Clipping changes both operand quantization error and the output-referred ADC quantization error, motivating their joint optimization.}
    \label{fig:imc-intro-and-motivation}
\end{figure*}

Analog in-memory computing (IMC) offers a promising approach toward more energy-efficient LLM inference \citep{verma2019memory,shanbhag2022benchmarking}. By performing matrix multiplications (MatMuls) directly within memory arrays in the analog domain, analog IMC can substantially reduce the energy associated with these compute-intensive operations. This efficiency, however, comes with reduced computational accuracy. While state-of-the-art analog IMC approaches have overcome the effects of analog noise (thermal, device, electronic sources of variability) \citep{valavi2019charge,lee2024switched}, an intrinsic and unavoidable source of noise with analog computation is analog-to-digital quantization at the output, where limited precision of the Analog-to-Digital Converter (ADC) appears as accumulation quantization in MatMul \citep{murmann2020mixed} (Fig.~\ref{fig:imc-intro}). Increasing the ADC resolution can mitigate this error, but this comes at rapidly increasing energy cost, which diminishes the efficiency advantage of analog IMC \citep{adc_survey}. This motivates algorithmic approaches that improve compute accuracy under a fixed ADC resolution.


Weight/activation clipping is a critical technique in digital quantization, and is in fact particularly promising for analog IMC. In digital processors, narrowing the activation and weight ranges reduces rounding error at the cost of clipping extreme values, where well-chosen thresholds can substantially improve quantized-model accuracy. Clipping can be even more beneficial in analog IMC because the reduced operand ranges also shrink the analog partial-sum range, mitigating ADC quantization error at a fixed resolution. This additional benefit, however, makes clipping calibration more challenging: the optimal clipping factors must jointly balance the rounding and clipping errors of weights/activations with the ADC quantization error (Fig.~\ref{fig:motivation}). Existing methods, developed primarily for digital processors, do not optimize the coupling with output quantization error; common approaches often involve costly search, leading to slow calibration and suboptimal accuracy. Analog IMC therefore requires a fast calibration method that jointly optimizes for all error sources.



To address this challenge, we introduce \methodname{} (\textbf{C}oupled \textbf{L}oss-\textbf{I}nformed \textbf{N}ewton \textbf{I}terations for \textbf{C}lipping), which couples an analytical error model with efficient second-order optimization. First, we derive an analytically tractable surrogate for the IMC MatMul output error that jointly models three key error sources: operand rounding and clipping errors, clipping-induced bias across the multiply-accumulate (MAC) accumulation, and ADC quantization error. Importantly, the surrogate preserves the coupling between activation/weight clipping and ADC error, while simplifying higher-order error interactions to make the loss, gradient, and Hessian efficiently computable from calibration statistics. Second, we exploit this analytical structure with a safeguarded Newton optimizer. Since each MatMul requires optimizing only a few clipping variables, Newton updates incur negligible optimization overhead; we further augment the damped Newton method with Hessian-conditioned initialization and positive-curvature safeguards to achieve fast and stable optimization.

We show that \methodname{} substantially improves analog IMC accuracy across multiple models and datasets, improving average zero-shot accuracy by $6.5$--$11.5$ percentage points over the grid search method for a 9-bit ADC IMC accelerator. At the same time, it accelerates clipping calibration by $10.0\times$--$12.1\times$. We additionally validate both components of \methodname{}: the surrogate remains close to measured IMC output error, and the safeguarded Newton solver reaches high-quality solutions rapidly, with all projections in LLaMA-3.2-3B and Qwen3-4B certified to lie within $1\%$ of the global optimum under the calibration objective. Overall, \methodname{} provides an efficient and accurate clipping calibration framework for LLM inference on analog IMC.


\section{Background}
\label{sec:background}

\subsection{Analog IMC}
\label{sec:background-adc}

\paragraph{Promise of Analog IMC.} Analog IMC offers substantially higher energy efficiency than conventional digital architectures for AI inference. Its key advantage comes from performing MAC reductions in analog directly within memory arrays, reducing the energy of both digital computation and data movement.
Consider a linear layer $\mathbf{y}=\mathbf{W}\mathbf{x}$, where an IMC macro contains $h$ physical rows and $w$ output columns. Since a single macro can store only a submatrix of $\mathbf{W}$, the layer is spatially partitioned into $K=\lceil d_{\mathrm{in}}/h\rceil$ row tiles and $L=\lceil d_{\mathrm{out}}/w\rceil$ output groups. For each tile $(k,\ell)$, the macro computes an analog partial sum
\begin{equation*}
\mathbf{p}^{(k,\ell)}
=
\mathbf{W}^{(k,\ell)}\mathbf{x}^{(k)},
\qquad
\mathbf{y}^{(\ell)}
=
\sum_{k=1}^{K}\mathbf{p}^{(k,\ell)},
\end{equation*}
where the $h$ products within each tile are accumulated directly in the analog domain, each partial sum is digitized by an ADC, and the $K$ row-tile contributions are then accumulated digitally.

Thus, the expensive inner-product reduction is performed largely within memory, while only tile-level partial sums require analog-to-digital (A/D) conversion and digital accumulation. As a point of reference, this computing paradigm has enabled analog IMC accelerators to achieve $128$ INT8-equivalent TOPS/W in 28-nm CMOS \citep{lee2024switched}, compared with $5$ TOPS/W for digital INT8 MAC \citep{taco201888} in 28-nm FD-SOI at the same $0.8$ V supply.
Appendix~\ref{appx:imc-llm-inference} further discusses different analog IMC architectures, their application to LLM inference workloads, and the corresponding energy-efficiency benefits.

\paragraph{ADC Quantization Bottleneck.} The efficiency, however, comes at the cost of reduced computational precision, with the analog-to-digital conversion of partial sums being the accuracy bottleneck.

To see why, consider the $h$ products accumulated within an IMC macro. Under a first-order model in which these contributions are independent and similarly distributed, the standard deviation of the partial sum grows as $\mathcal{O}(\sqrt{h})$, while its full accumulation range grows as $\mathcal{O}(h)$. Because this full range must be mapped onto a fixed analog voltage range, the statistical signal swing occupies only an $\mathcal{O}(1/\sqrt{h})$ fraction of the available range. A fixed-resolution ADC must therefore resolve increasingly small signal variations within the same full-scale range \citep{murmann2020mixed}. Consequently, the analog signal increasingly underutilizes the ADC range, degrading effective resolution and compute fidelity.
This ADC range-utilization problem motivates algorithmic techniques that reduce the effective partial-sum dynamic range without increasing ADC resolution.

\subsection{Clipping for LLM Quantization}
\label{sec:background-clipping}

\paragraph{Rounding–Clipping Trade-off.} Clipping reduces quantization error by restricting the dynamic range represented by a fixed number of quantization levels.
For a $b$-bit uniform quantizer over the clipping interval $[c_{\mathrm{down}},c_{\mathrm{up}}]$, the quantization step is $\Delta=(c_{\mathrm{up}}-c_{\mathrm{down}})/(2^b-1)$. Decomposing the quantization error as $e=e_{\mathrm{round}}+e_{\mathrm{clip}}$, the standard uniform quantization-noise approximation \citep{widrow1996statistical} models the in-range rounding error as zero-mean with $\mathbb{E}[e_{\mathrm{round}}^2]\approx \Delta^2/12$. Narrowing the clipping interval therefore decreases the quantization step and reduces rounding error for in-range values, but causes more values to saturate at the interval boundaries and increases clipping error.
The optimal clipping thus needs to balance reduced rounding error against increased distortion from clipped values.

\paragraph{Clipping Parameterization.} For both activations and weights, clipping thresholds can be parameterized as multiplicative retention factors on their corresponding tensor ranges. Based on the typical distributions encountered in LLMs, we consider asymmetric activation clipping and symmetric weight clipping. Let
\begin{equation*}
M_x^+ = \max(\mathbf{X}),
\qquad
M_x^- = \min(\mathbf{X}),
\qquad
M_w = \max |\mathbf{W}|.
\end{equation*}
The clipping thresholds are parameterized as
\begin{equation}
c_{x,\mathrm{up}}=\gamma M_x^+,
\qquad
c_{x,\mathrm{down}}=\beta M_x^-,
\qquad
c_w=\alpha M_w,
\label{eq:clipping-parameterization}
\end{equation}
where $\gamma$, $\beta$, and $\alpha$ are bounded retention factors in $(0,1]$. For dynamically quantized activations, $M_x^+$ and $M_x^-$ are sample-dependent, making $c_{x,\mathrm{up}}$ and $c_{x,\mathrm{down}}$ vary with the input. The activation clipping factors $\gamma$ and $\beta$ remain fixed and specify the retained fractions of the dynamically observed positive and negative ranges. In contrast, the weight range $M_w$ is fixed, so $c_w$ is static for a fixed $\alpha$.

The corresponding activation and weight quantization scales are
\begin{equation}
s_x
=
\frac{\gamma M_x^+-\beta M_x^-}{2^{b_x}-1},
\qquad
s_w
=
\frac{\alpha M_w}{2^{b_w-1}-1}.
\label{eq:operand-scales}
\end{equation}
Thus, the clipping factors control the retained fraction of each operand range and, through the resulting range, its quantization resolution. This parameterization incurs low runtime overhead: under dynamic per-token activation quantization, the static factors $\gamma$ and $\beta$ only rescale the min/max statistics already computed for scale generation, requiring no additional activation scan or reduction, while weight clipping remains entirely static.


\paragraph{Clipping in Analog IMC.} In analog IMC, clipping affects not only operand quantization error but also quantization error incurred in the subsequent A/D conversion.
At a fixed ADC resolution and analog full-scale range, the ADC introduces a fixed quantization error in the quantized partial-sum domain. However, this error is mapped back to the floating-point output domain through the product of the activation and weight scales in Eq.~\eqref{eq:operand-scales}. Reducing either operand range therefore decreases the floating-point magnitude of the same ADC-domain quantization error. Stronger clipping can therefore reduce both operand rounding error and ADC quantization error, but simultaneously increases clipping distortion. Moreover, activation and weight clipping are intrinsically coupled because they jointly determine the scale $s_xs_w$ of the analog partial sum.

Clipping calibration in analog IMC therefore requires jointly balancing operand quantization error, clipping error, and ADC quantization error, rather than optimizing each operand in isolation. Prior work typically treats operand clipping separately or modifies the ADC range directly; Appendix~\ref{appx:existing-clipping} discusses these approaches and their differences from \methodname{}.

\section{Coupled Analytical Output-Error Model}
\label{sec:analytical_error_model}

\paragraph{Operand-Space Quantization Error.} To optimize the clipping factors jointly, we first need an analytical description of how clipping changes the activation and weight quantization errors. Let \(z\) denote an element of an activation or weight tensor, clipped to $[c_{\mathrm{down}},c_{\mathrm{up}}]$. We view the observed activation and weight values as samples from underlying operand distributions, with the required statistics estimated from calibration activations and pretrained weights. The clipping mean-squared error (MSE) is given by the two distribution tails \citep{banner2019posttraining},
\begin{equation*}
\mathbb{E}[e_{\mathrm{clip}}^2] = \int_{-\infty}^{c_{\mathrm{down}}}(c_{\mathrm{down}} - z)^2 p(z)\,dz + \int_{c_{\mathrm{up}}}^{\infty}(c_{\mathrm{up}} - z)^2 p(z)\,dz.
\end{equation*}
Direct evaluation of these integrals requires estimating $p(z)$ and repeatedly integrating it as the clipping thresholds change. Following OCTAV \citep{sakr2022octav}, we instead express them as expectations over indicator functions,
\begin{equation}
\mathbb{E}[e_{\mathrm{clip}}^2] = \mathbb{E}\!\left[(c_{\mathrm{up}} - z)^2 \mathbf{1}_{z>c_{\mathrm{up}}} + (c_{\mathrm{down}} - z)^2 \mathbf{1}_{z<c_{\mathrm{down}}}\right],
\label{eq:clipping-mse-indicator}
\end{equation}
which can be estimated directly from the available operand samples using elementwise operations and reductions. For rounding error, we use the zero-mean $\Delta^2/12$ model introduced in Sec.~\ref{sec:background-clipping}, instantiated with the operand scales in Eq.~\eqref{eq:operand-scales}. Other required error statistics, such as the signed clipping error, can be expressed in similarly efficient forms; their derivations are given in Appendix~\ref{appx:quantization-error-decomposition}. Together, these operand-level statistics provide the quantities needed to model MatMul output error.

\paragraph{Output-Space Operand Error.} We next propagate these clipping-dependent operand errors through the MatMul and derive a tractable surrogate for its output error. Appendix~\ref{appx:matmul-output-error-surrogate} provides additional derivations and empirical validation for the approximations introduced below.

For a MatMul output $y=\sum_i w_i x_i$, $w_i$ denotes the observed pretrained weight, while the required operand-error statistics are modeled as described above. Let $\hat{x}_i=x_i+e_{x,i}$ and $\hat{w}_i=w_i+e_{w,i}$, where $e_{x,i}$ and $e_{w,i}$ denote the activation and weight quantization errors, respectively. The quantization error contributed by the $i$-th MAC is then exactly
\begin{equation}
\delta y_i = x_i e_{w,i} + w_i e_{x,i} + e_{x,i}e_{w,i}.
\label{eq:mac-error}
\end{equation}

The total operand-induced output error is $\sum_i \delta y_i$, whose mean-squared error expands as
\begin{equation*}
\mathbb{E}\!\left[\left(\sum_i \delta y_i\right)^2\right] = \sum_i\mathbb{E}[\delta y_i^2] + \sum_{i\neq j}\mathbb{E}[\delta y_i\delta y_j].
\end{equation*}

Directly modeling the second term requires joint error statistics across MAC coordinates. We therefore assume that the centered MAC errors are approximately pairwise uncorrelated across the reduction dimension, which gives $\mathbb{E}[\delta y_i\delta y_j] \approx \mathbb{E}[\delta y_i]\mathbb{E}[\delta y_j]$ for $i\neq j$. Applying $\sum_{i\neq j}a_i a_j=(\sum_i a_i)^2-\sum_i a_i^2$ with $a_i=\mathbb{E}[\delta y_i]$, the output MSE is approximated by
\begin{equation}
\sum_i\mathbb{E}[\delta y_i^2] + \left(\sum_i\mathbb{E}[\delta y_i]\right)^2 - \sum_i\mathbb{E}[\delta y_i]^2.
\label{eq:operand-output-mse}
\end{equation}
This per-coordinate factorization eliminates explicit cross-coordinate covariance estimation, providing the structure needed for efficient gradient evaluation in Sec.~\ref{sec:clipping_calibration}.

The remaining per-coordinate term $\mathbb{E}[\delta y_i^2]$ still contains mixed and higher-order interactions between activation and weight quantization errors. For a tractable surrogate, we retain the two leading signal--noise contributions and use a second-moment separability approximation,
\begin{equation}
\mathbb{E}[\delta y_i^2] \approx w_i^2\mathbb{E}[e_{x,i}^2] + \mathbb{E}[x_i^2]\mathbb{E}[e_{w,i}^2],
\label{eq:mac-second-moment}
\end{equation}
while omitting the remaining mixed and higher-order terms. In contrast, we retain the signed mean $\mathbb{E}[\delta y_i]$, since clipping can introduce nonzero bias that accumulates coherently across the MAC reduction. Under the zero-mean, signal-independent rounding model,
\begin{equation}
\mathbb{E}[\delta y_i] = w_i\mathbb{E}[e_{x,\mathrm{clip},i}] + e_{w,\mathrm{clip},i}\left(\mathbb{E}[x_i] + \mathbb{E}[e_{x,\mathrm{clip},i}]\right).
\label{eq:mac-signed-mean}
\end{equation}

By expressing operand quantization errors in the MatMul output space, the same formulation also allows them to be combined directly with the output-referred ADC quantization error.

\paragraph{Inter-related IMC Output Error.} Finally, we incorporate ADC quantization error to obtain the complete analog-IMC output-error objective optimized by \methodname{}. At fixed ADC resolution and analog full-scale range, each slice-level ADC conversion has a fixed quantization step. Our hardware mapping, following \citet{lee2024switched} and \citet{cambricon-cim}, decomposes each 8-bit operand into 4-bit slices and digitally recombines the resulting ADC-quantized partial products with their corresponding significance weights. We therefore define $\Delta_{\mathrm{ADC}}$ as the \emph{effective ADC step size}, after accounting for these slice-level conversions and recombination weights, such that the ADC error of one reconstructed IMC partial sum has variance $\Delta_{\mathrm{ADC}}^2/12$; its derivation from the physical ADC step is provided in Appendix~\ref{appx:matmul-output-error-surrogate}.

Following the statistical theory of quantization \citep{widrow1996statistical}, the individual slice-level ADC errors can also be modeled as independent, zero-mean uniform noise. Through the operand scales in Eq.~\eqref{eq:operand-scales}, the reconstructed ADC error is mapped to the floating-point output domain by the product $s_x s_w$. For $K$ independently digitized row-tile partial sums, the variances therefore add, giving
\begin{equation}
\mathcal{L}_{\mathrm{ADC}}
=
K\frac{\Delta_{\mathrm{ADC}}^2}{12}(s_x s_w)^2.
\label{eq:adc-loss}
\end{equation}
Because $\Delta_{\mathrm{ADC}}$ is fixed by the ADC configuration and bit-sliced hardware mapping, while $s_x$ and $s_w$ depend on the clipping factors, the output-referred ADC error remains jointly controlled by activation and weight clipping.

Assuming that the operand-induced output error and ADC quantization error are approximately uncorrelated, their MSE contributions approximately add. Combining the operand-error surrogate above with the ADC term gives the complete calibration objective for $\boldsymbol{\theta}=(\gamma,\beta,\alpha)$,
\begin{equation}
\mathcal{L}(\boldsymbol{\theta})
= \underbrace{\sum_i\left(
w_i^2\mathbb{E}[e_{x,i}^2]
+ \mathbb{E}[x_i^2]\mathbb{E}[e_{w,i}^2]\right)}_{\mathcal{L}_{\mathrm{diag}}}
+ \underbrace{\left(\sum_i\mathbb{E}[\delta y_i]\right)^2 - \sum_i\mathbb{E}[\delta y_i]^2}_{\mathcal{L}_{\mathrm{bias}}}
+ \underbrace{K\frac{\Delta_{\mathrm{ADC}}^2}{12}(s_x s_w)^2}_{\mathcal{L}_{\mathrm{ADC}}}.
\label{eq:surrogate-objective}
\end{equation}

Here, the activation error statistics and scale, $\mathbb{E}[e_{x,i}^2]$ and $s_x$, depend on $(\gamma,\beta)$, while the weight error statistics and scale, $\mathbb{E}[e_{w,i}^2]$ and $s_w$, depend on $\alpha$. The signed error $\mathbb{E}[\delta y_i]$ depends jointly on all three clipping factors.

The resulting surrogate captures the coupled effect of activation and weight rounding, clipping, and ADC quantization on MatMul output error while retaining a structure amenable to efficient optimization, as will be illustrated in the following sections.

\section{Clipping Calibration with Safeguarded Newton Iterations}
\label{sec:clipping_calibration}

Building on the surrogate in Eq.~\eqref{eq:surrogate-objective}, we optimize the clipping factors using a safeguarded Newton-type method. The low-dimensional clipping problem admits analytical gradients and an efficient curvature approximation from calibration statistics. Moreover, the relevant optimization region is empirically locally convex: across all evaluated projections, we observe a single connected positive-semidefinite (PSD) region of the full Hessian of the surrogate containing the initialization, accepted trajectory, and calibrated solution (Appendix~\ref{app:psd-region-validation}). In this section, we derive the full analytical gradient and an approximate Hessian. We then use these derivatives in safeguarded Newton updates.

\paragraph{Analytical Gradient Evaluation.} The indicator-based error statistics in Eq.~\eqref{eq:clipping-mse-indicator} permit analytical differentiation with respect to the clipping factors.

For a single MatMul, we optimize $\boldsymbol{\theta}=(\gamma,\beta,\alpha)$. When one activation feeds multiple branches, e.g., the $q/k/v$ or $up$/$gate$ projections, we jointly optimize $\boldsymbol{\theta}=(\gamma,\beta,\alpha_1,\ldots,\alpha_M)$, where $(\gamma,\beta)$ are shared activation clipping factors and each branch has its own weight clipping factor $\alpha_m$. We denote the gradient of the surrogate objective by
$\mathbf{g}(\boldsymbol{\theta})=\nabla_{\boldsymbol{\theta}}\mathcal{L}(\boldsymbol{\theta})$.

To illustrate the gradient computation, differentiating the activation clipping-error second moment with respect to the upper clipping factor $\gamma$ using Leibniz's rule gives
\begin{equation*}
\frac{\partial}{\partial\gamma}
\mathbb{E}[e_{x,\mathrm{clip}}^2]
=
2M_x^+\!\int_{c_{x,\mathrm{up}}}^{\infty}
(c_{x,\mathrm{up}}-x)p(x)\,dx
=
2M_x^+\mathbb{E}\!\left[
e_{x,\mathrm{clip}}\mathbf{1}_{x>c_{x,\mathrm{up}}}
\right],
\end{equation*}
where the boundary term vanishes because the clipping residual is zero at $x=c_{x,\mathrm{up}}$. The final expression can be evaluated directly from calibration samples using a threshold comparison, elementwise multiplication, and reduction \citep{sakr2022octav}. Analogous analytical expressions apply to the remaining activation, weight, bias, and ADC terms, as detailed in Appendix~\ref{appx:gradient-hessian}.
Consequently, the full gradient of the coupled clipping objective can be evaluated directly from calibration statistics without dense clipping search or repeated MatMul reconstruction.

\paragraph{Approximate Hessian Evaluation.} Most second-order terms retain the same efficient empirical structure as the gradient. For example, the activation clipping-error second moment satisfies
\begin{equation*}
\frac{\partial^2}{\partial\gamma^2}
\mathbb{E}[e_{x,\mathrm{clip}}^2]
=
2(M_x^+)^2
\mathbb{E}\!\left[
\mathbf{1}_{x>c_{x,\mathrm{up}}}
\right],
\end{equation*}
which can still be evaluated directly from calibration samples. In contrast, the signed clipping mean, which enters the accumulated bias term of the output-error objective, satisfies
\begin{equation*}
\frac{\partial^2}{\partial\gamma^2}
\mathbb{E}[e_{x,\mathrm{clip}}]
=
-(M_x^+)^2
p_x(c_{x,\mathrm{up}}),
\end{equation*}
and therefore requires the probability density at the moving clipping boundary. Such pointwise density values cannot be obtained by a simple empirical reduction and instead require density estimation, which is more costly and can be noisy near distribution tails. We therefore omit these density-sensitive second-order terms and construct an approximate Hessian, denoted $\widetilde{\mathbf H}$, from the remaining analytical curvature terms, while leaving the surrogate loss and analytical gradient unchanged. The approximate Hessian affects only the proposed Newton direction; descent and sufficient decrease are still verified using the exact surrogate and gradient. The complete Hessian derivation and omitted terms are detailed in Appendix~\ref{appx:gradient-hessian}.

\paragraph{Safeguarded Newton-Type Optimization.} We combine the analytical gradient and approximate Hessian to construct the approximate Newton direction at each calibration iteration.

At iteration $t$, we compute the approximate Newton direction and projected trial update
\begin{equation}
\mathbf{d}_t=-\widetilde{\mathbf{H}}_t^{-1}\mathbf{g}_t,
\qquad
\boldsymbol{\theta}_{\mathrm{trial}}
=
\Pi_{\Theta}\!\left(\boldsymbol{\theta}_t+\eta_t\mathbf{d}_t\right),
\label{eq:safeguarded-newton-update}
\end{equation}
where $\eta_t\in(0,1]$ is selected by backtracking line search and $\Pi_{\Theta}$ projects onto the feasible clipping-factor range. We initialize from a coarse grid by selecting the lowest-loss point whose approximate Hessian has positive minimum eigenvalue, placing the optimizer in a locally well-conditioned region (Appendix~\ref{appx:newton-initialization}). Because projection can alter the Newton direction, we accept a trial step only if the actual projected step is a descent direction, satisfies the Armijo sufficient-decrease condition, and preserves positive curvature; otherwise, we reduce the step size and retry. Consequently, every accepted update decreases the surrogate objective. Within the empirically observed locally convex region, convergence to a stationary point therefore corresponds to a local minimum.
We further validate in Sec.~\ref{sec:post-hoc-epsilon-optimality-validation} and Appendix~\ref{appx:optimality-validation} that the resulting solutions are globally $\epsilon$-optimal under the calibration objective across all evaluated projections.

\section{Experiments}
\label{sec:experiments}

In this section, we first describe the common analog IMC configuration and clipping baselines in Sec.~\ref{sec:imc-configuration-and-baselines}. We then show why clipping must jointly account for operand and ADC quantization errors in Sec.~\ref{sec:sqnr-analysis-across-error-sources}, followed by the resulting model-accuracy and calibration-time improvements of \methodname{} in Sec.~\ref{sec:model-accuracy-and-calibration-time}. In Sec.~\ref{sec:surrogate-loss-fidelity}, we show that the analytical surrogate, while enabling efficient optimization, remains accurate. Finally, Sec.~\ref{sec:post-hoc-epsilon-optimality-validation} demonstrates that the safeguarded Newton optimizer is efficient and converges to near-optimal solutions under the surrogate objective.

\subsection{IMC Configuration and Baselines}
\label{sec:imc-configuration-and-baselines}

We evaluate all methods under a common analog IMC inference setting. MatMuls are mapped to 512-row IMC arrays; larger reductions are partitioned across arrays and their digitized partial sums accumulated digitally. Operands are rotated \citep{Quarot} then quantized to 8-bit precision and represented using 4-bit operand slices, with each analog partial sum quantized by a 9-bit ADC. The ADC full-scale range is fixed to the worst-case analog partial-sum range implied by the operand representation and IMC reduction dimension, and is shared across all methods. Appendix~\ref{appx:extended-results} studies sensitivity to ADC precision and IMC reduction dimension.

We compare two IMC baselines across all four models. \textbf{No Clipping} uses the dynamic quantization range without additional clipping. Following PrefixQuant \citep{chen2024prefixquant}, \textbf{W/A Grid Search} applies a 1D weight grid search followed by a 2D activation grid search. It selects weight clipping using activation-aware linear-output reconstruction MSE, then searches separate upper and lower activation clipping factors using decoder-block output MSE. This accommodates asymmetric activation distributions without the cost of a joint 3D search. Designed for digital quantization, W/A Grid Search accounts for analog IMC ADC error only indirectly. FP16 serves as the full-precision accuracy reference. Unless otherwise specified, these configurations apply throughout.

\subsection{Output-Error Analysis Across Error Sources}
\label{sec:sqnr-analysis-across-error-sources}

\begin{figure*}[t]
    \centering
    \includegraphics[width=\textwidth]{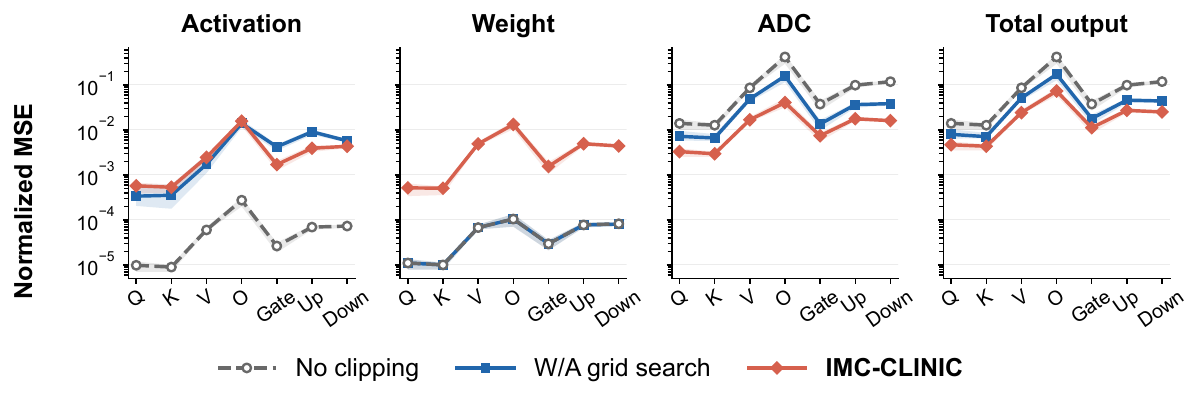}
    \caption{Output-error analysis for LLaMA-3.2-3B across projections. Each point reports the median normalized MSE across 28 decoder layers, and shaded regions denote the 25th--75th percentiles. The first three panels isolate activation-, weight-, and ADC-induced output errors, while the final panel measures the total IMC output error directly. Lower is better.}
    \label{fig:sqnr-analysis}
\end{figure*}

We first examine how different clipping methods balance the error sources that determine IMC MatMul accuracy. Figure~\ref{fig:sqnr-analysis} evaluates LLaMA-3.2-3B \citep{LLaMA3} using WikiText-2 \citep{WikiText-2} validation activations under the IMC configuration in Sec.~\ref{sec:imc-configuration-and-baselines}. We separately measure the output-referred errors induced by activation quantization, weight quantization, and ADC quantization, together with the resulting total output MSE. Each error is normalized by the corresponding full-precision output signal power. For each projection type, we report the median across all 28 decoder layers, with the shaded region indicating the interquartile range. The precise definitions of the isolated error terms are provided in Appendix~\ref{appx:sqnr-experiment-setup}.

Without clipping, activation and weight quantization errors remain small, but ADC quantization dominates the total output error. W/A grid search reduces the ADC error by accepting more activation quantization error, while keeping the weight quantization error close to that of no clipping. In contrast, \methodname{} better balances all three error sources, allowing somewhat larger operand quantization errors to further reduce the ADC error. This results in the lowest total output MSE across all projection types.

These results show that minimizing activation and weight quantization errors individually does not yield the best IMC output accuracy. By jointly accounting for operand and ADC quantization errors, \methodname{} finds clipping factors that achieve a better overall error trade-off.
The corresponding optimized clipping factors are reported in Appendix~\ref{appx:clipping-factors}, where they typically fall in the $0.5$--$0.7$ range, indicating that balancing operand and ADC errors favors relatively aggressive clipping.

\subsection{Model Accuracy and Calibration Time}
\label{sec:model-accuracy-and-calibration-time}

\newsavebox{\mainresultstablebox}
\begin{wraptable}{r}{0.50\textwidth}
    \vspace{-8pt}
    \centering
    \scriptsize
    \setlength{\tabcolsep}{2.8pt}
    \renewcommand{\arraystretch}{0.84}
    \begin{lrbox}{\mainresultstablebox}
    \begin{tabular}{@{}llr:r:r@{}}
        \hline
        Model & Method
        & PPL $\downarrow$
        & Avg. $\uparrow$
        & Time $\downarrow$ \\
        \hline

        LLaMA-3.2-3B
            & No clip.
            & 175.70
            & 0.384
            & -- \\
            & W/A grid
            & 19.12
            & 0.488
            & 32.9 \\
            & \textbf{\methodname{}}
            & \textbf{12.27}
            & \textbf{0.575}
            & \textbf{3.3} \\
        \hdashline
            & FP16
            & 7.81
            & 0.650
            & -- \\
        \hline

        LLaMA-3.1-8B
            & No clip.
            & 127.28
            & 0.397
            & -- \\
            & W/A grid
            & 13.68
            & 0.567
            & 73.7 \\
            & \textbf{\methodname{}}
            & \textbf{9.17}
            & \textbf{0.646}
            & \textbf{6.1} \\
        \hdashline
            & FP16
            & 6.24
            & 0.716
            & -- \\
        \hline

        Qwen3-4B
            & No clip.
            & 2156.59
            & 0.359
            & -- \\
            & W/A grid
            & 71.68
            & 0.419
            & 44.4 \\
            & \textbf{\methodname{}}
            & \textbf{30.17}
            & \textbf{0.534}
            & \textbf{4.3} \\
        \hdashline
            & FP16
            & 13.64
            & 0.667
            & -- \\
        \hline

        Qwen3-8B
            & No clip.
            & 38.61
            & 0.442
            & -- \\
            & W/A grid
            & 13.94
            & 0.582
            & 74.6 \\
            & \textbf{\methodname{}}
            & \textbf{11.72}
            & \textbf{0.647}
            & \textbf{6.5} \\
        \hdashline
            & FP16
            & 9.72
            & 0.694
            & -- \\
        \hline
    \end{tabular}
    \end{lrbox}
    \normalsize
    \captionsetup{width=\dimexpr\wd\mainresultstablebox\relax,skip=10pt}
    \caption{WikiText-2 PPL, mean seven-task accuracy (Avg.), and calibration time (min). Appendix~\ref{appx:full-task-results} reports task-level results.}
    \label{tab:accuracy-and-calibration-time}
    \usebox{\mainresultstablebox}
    \vspace{-6pt}
\end{wraptable}

We next evaluate \methodname{} on LLaMA-3.2-3B \citep{LLaMA-3.2}, LLaMA-3.1-8B \citep{LLaMA3}, Qwen3-4B, and Qwen3-8B \citep{yang2025qwen3}. We report WikiText-2 perplexity \citep{WikiText-2}, zero-shot accuracy (\texttt{acc}) on WinoGrande \citep{winogrande} and BoolQ \citep{BoolQ}, and zero-shot normalized accuracy (\texttt{acc\_norm}) on OpenBookQA \citep{openbookQA}, PIQA \citep{PIQA}, ARC-Challenge and ARC-Easy \citep{ARC}, and HellaSwag \citep{HellaSwag}. All clipping methods use the same eight WikiText-2 calibration sequences of length 2048.

Table~\ref{tab:accuracy-and-calibration-time} compares no clipping, W/A grid search, and \methodname{} under the same IMC configuration, reporting perplexity (denoted as PPL) and average zero-shot accuracy; the complete task-level results are provided in Appendix~\ref{appx:full-task-results}. Compared with W/A grid search, \methodname{} reduces perplexity by 15.9--57.9\% and improves average zero-shot accuracy by 6.5--11.5 percentage points across the four models. It also outperforms the grid-search baseline on all evaluated model-task pairs, consistently narrowing the gap to FP16 inference.

The final column of Table~\ref{tab:accuracy-and-calibration-time} reports end-to-end clipping calibration time measured using one NVIDIA A100 80 GB GPU per run. W/A grid search sequentially evaluates candidate weight and activation clipping factors through repeated model execution, whereas \methodname{} directly optimizes the surrogate objective using its analytical gradient and approximate Hessian. As a result, \methodname{} is $10.0\times$--$12.1\times$ faster while simultaneously achieving higher model accuracy.

\subsection{Fidelity of Surrogate Loss}
\label{sec:surrogate-loss-fidelity}

We next evaluate whether the analytical surrogate accurately captures the empirical IMC output error. To provide a direct comparison, we construct an ADC-aware alternating-search benchmark that uses empirical MatMul output MSE as its objective and performs coordinate search over the activation and weight clipping factors (details in Appendix~\ref{appx:adc-aware-alternating}). Unlike the W/A grid-search baseline, both activation and weight candidates are evaluated through the ADC-enabled IMC forward path, so all clipping-factor updates directly account for ADC quantization.

\begin{figure}[t]
    \centering
    \includegraphics[width=\textwidth]{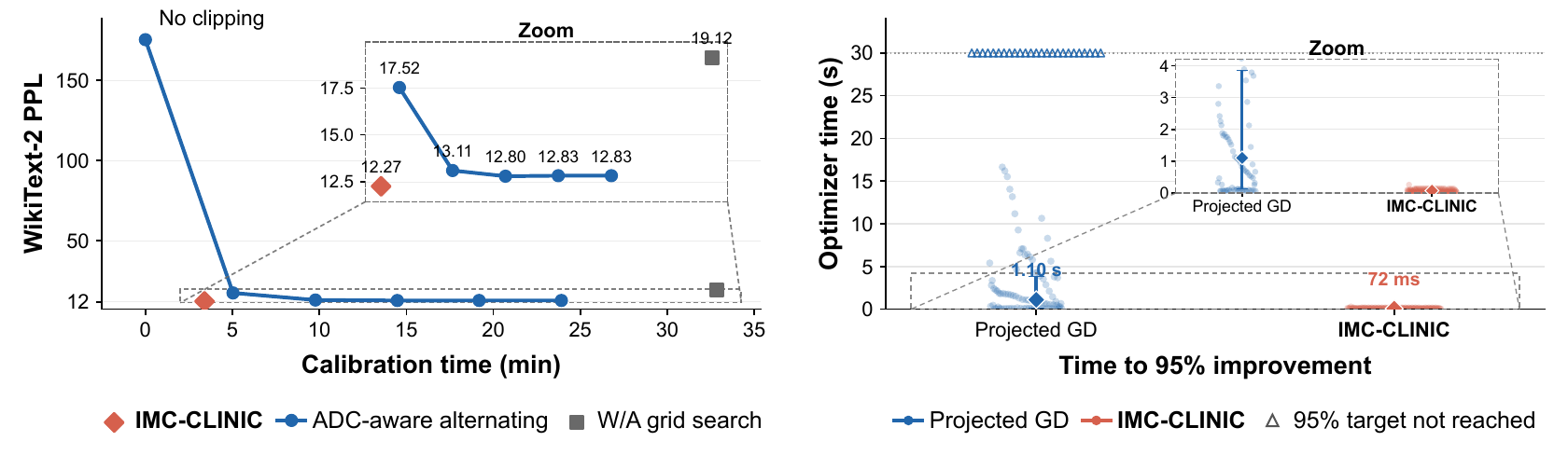}
    \caption{\textbf{Calibration and optimization efficiency on LLaMA-3.2-3B.} \textbf{Left:} \methodname{} achieves a better WikiText-2 PPL--calibration-time trade-off than direct ADC-aware alternating search. \textbf{Right:} Its safeguarded Newton optimizer reaches $95\%$ of the post-initialization loss improvement faster than projected gradient descent ($72$\,ms vs.\ $1.10$\,s median).}
    \label{fig:surrogate-loss-fidelity}
\end{figure}

As shown in Fig.~\ref{fig:surrogate-loss-fidelity} (left), this direct empirical search requires substantially longer calibration while achieving slightly worse perplexity than \methodname{}. Thus, although \methodname{} optimizes an analytical approximation rather than repeatedly evaluating empirical MSE, it achieves a better perplexity--calibration-time trade-off. We further compare the surrogate directly against empirical output MSE in Appendix~\ref{appx:surrogate-fidelity}, where the median mismatch remains only about $2$--$4\%$ across all projection types in LLaMA-3.2-3B.

Together, these results show that the surrogate closely tracks the empirical objective while enabling much more efficient optimization with analytical first- and second-order information.

\subsection{Optimization Efficiency and Solution Optimality}
\label{sec:post-hoc-epsilon-optimality-validation}

Here, we evaluate the efficiency and solution quality of the safeguarded Newton optimizer in \methodname{}. We compare against projected gradient descent (GD) from the same initialization and measure the optimizer time required to achieve $95\%$ of the post-initialization surrogate-loss improvement (details in Appendix~\ref{appx:optimizer-comparison}). As shown in Fig.~\ref{fig:surrogate-loss-fidelity} (right), \methodname{} reaches this target in a median of $72\,\mathrm{ms}$, compared with $1.10\,\mathrm{s}$ for projected gradient descent, while some projected-gradient runs do not reach the target within the time limit.

Fast convergence alone does not guarantee a high-quality solution. We therefore additionally perform full-domain branch-and-bound validation of its global $\epsilon$-optimality in Appendix~\ref{appx:optimality-criterion}, asking whether any feasible clipping factors can improve the calibrated surrogate objective by more than $1\%$. Across all evaluated projections in LLaMA-3.2-3B and Qwen3-4B, the solutions found by \methodname{} are certified to be globally $1\%$-optimal; the certification procedure is detailed in Appendix~\ref{appx:optimality-validation}.

Together, these results show that the safeguarded Newton optimizer is both fast and near-optimal under the surrogate objective.

\section{Conclusion}
\label{sec:conclusion}

We introduced \methodname{}, a clipping-calibration framework for analog IMC that jointly accounts for activation, weight, and ADC quantization effects through an analytical MatMul output-error surrogate. By combining this surrogate with safeguarded Newton-type optimization, \methodname{} avoids expensive grid search while retaining the coupling that is critical for IMC clipping. Across four LLMs, it consistently improves model accuracy over W/A grid search while reducing calibration time by 10.0$\times$--12.1$\times$. We further show that the analytical surrogate closely tracks empirical IMC output error, while the optimizer converges rapidly and is certified within 1\% of the global optimum under the calibration objective across all projections on two representative models. These results show that clipping can be calibrated efficiently and reliably for analog IMC when operand and ADC errors are optimized jointly.

\subsection*{AI use statement}

\begin{itemize}
    \item \textbf{Writing assistance.} Generative AI tools were used to aid and polish manuscript writing, including improving clarity, concision, and presentation.

    \item \textbf{Retrieval and discovery.} Generative AI tools were used to help identify and summarize relevant prior work and references.

    \item \textbf{Research ideation and execution.} Generative AI tools were used to support research discussions, code implementation, writing scripts to run sweep and ablation experiments, and collecting experimental results.

    \item \textbf{Drafting.} Generative AI tools were used to draft portions of the manuscript, which were subsequently reviewed and revised by the authors.
\end{itemize}


\subsection*{Reproducibility statement}

We provide code in the supplementary material to reproduce the main results of this work, including \methodname{}, the analog IMC simulator, baseline clipping methods, model evaluation, sweep and ablation experiments, and the branch-and-bound optimality validation. The paper specifies the evaluated models and datasets, quantization and IMC configurations, calibration settings, evaluation metrics, and baseline procedures, with additional experimental details and sensitivity analyses provided in the appendix. The released code contains the configurations and scripts used to reproduce the reported tables and figures. All evaluated models and datasets are publicly available.




\bibliography{references}
\bibliographystyle{iclr_formatting/iclr2027_conference}

\appendix
\clearpage
\startcontents[appendix]

\section*{Appendix Contents}
\label{appx:contents}
\printcontents[appendix]{}{1}{}

\clearpage
\section{Analog IMC for LLM Inference}
\label{appx:imc-llm-inference}

\subsection{Analog IMC Architectures}
\label{appx:imc-architectures}

Analog IMC has been demonstrated across diverse memory technologies, including RRAM \citep{RRAM-IMC}, MRAM \citep{MRAM-IMC}, PCM \citep{PCM-IMC}, eDRAM \citep{edram-IMC}, and SRAM, as well as across different analog accumulation mechanisms such as current-domain \citep{current-domain-IMC}, time-domain \citep{time-domain-IMC}, and charge-domain computation \citep{valavi2019charge,lee2021fully}. These approaches offer different tradeoffs in density, programmability, and efficiency, but analog variation, circuit nonlinearity, and noise can limit the precision of high-dimensional MatMul computation.

For high-precision LLM inference, we therefore focus on switched-capacitor SRAM IMC \citep{valavi2019charge,lee2021fully,lee2024switched}. Switched-capacitor designs perform accumulation in the charge domain using lithographically defined metal capacitors, whose high matching precision enables low-noise analog computation over large reduction dimensions. Consequently, ADC quantization, rather than analog compute noise within the array, becomes the primary precision limitation \citep{lee2024switched}. Prior capacitor-based silicon measurements \citep{jia2021scalable,lee2024switched} further show close agreement between measured chip outputs and bit-true simulation, supporting an accurate algorithm-level abstraction of the hardware behavior. This high-SNR (signal-to-noise ratio) and accurately modelable regime provides the hardware basis for the clipping optimization studied in this work.

At the system level, analog IMC provides efficiency benefits in two ways. First, storing weights within the compute arrays reduces weight movement between memory and processing elements. Second, performing the inner-product reduction directly in the analog domain substantially reduces the energy of MatMul computation. For modern LLMs, however, the limited capacity of on-chip IMC arrays prevents the full model from being kept locally, requiring weights to be repeatedly streamed from higher levels of the memory hierarchy. The compute-energy reduction therefore remains beneficial even when full weight localization is infeasible.

\subsection{LLM Inference Workloads}
\label{appx:llm-inference-workloads}

Analog IMC benefits the two phases of LLM inference differently. Prefill is typically compute-bound, allowing the low-energy MatMul computation of IMC to translate directly into system-level efficiency gains. Decode, in contrast, is traditionally memory-bound, so the compute-energy advantage of IMC can be diluted by weight movement and other memory-system costs. However, emerging LLM workloads increasingly shift decode and overall inference toward more compute-intensive regimes: agentic workloads involve long, multi-turn contexts that increase the importance of input-side computation \citep{vllm2026agentx}, while continuous batching increases weight reuse across concurrent decode sequences \citep{yu2022orca} and speculative decoding evaluates multiple candidate tokens per target-model pass \citep{leviathan2023fast}. Together, these trends increase the relevance of energy-efficient MatMul computation across both prefill and decode.

This compute-energy benefit becomes particularly important on specialized accelerators that aggressively optimize data movement through tiling, buffering, and reuse. In such systems, memory-side overhead can be substantially amortized, shifting a larger fraction of system energy toward the compute cores. For example, the TransFusion study \citep{zhang2025transfusion} reports that PE-array computation accounts for the majority of energy in its cloud accelerator configuration after aggressive data-movement optimization. This is our targeted regime in which the low-energy MatMul computation of analog IMC can translate into substantial system-level efficiency gains.

\subsection{Energy-Efficiency Comparison}
\label{appx:energy-efficiency-comparison}

We compare the switched-capacitor SRAM IMC macro of ~\citet{lee2024switched} with the $8\times8$-bit digital MAC of \citet{taco201888}. Both operate at $0.8$ V in nominal 28-nm technologies, using CMOS and FD-SOI, respectively. \citet{lee2024switched} report $8161$ TOPS/W normalized to 1-bit computation for Config.~2. Since an INT8 MAC contains $8\times8=64$ 1-bit products,
\begin{equation*}
\eta_{\mathrm{IMC,INT8}}
=
\frac{8161}{64}
=
127.52~\mathrm{TOPS/W}.
\end{equation*}

This is an INT8-equivalent efficiency derived from the reported 1-bit-normalized metric.

For \citet{taco201888}, the $8\times8$-bit MAC consumes $0.390$ pJ/MAC at $0.8$ V, giving
\begin{equation*}
\eta_{\mathrm{digital,INT8}}
=
\frac{2}{0.390~\mathrm{pJ}}
=
5.13~\mathrm{TOPS/W},
\end{equation*}
where one multiplication and one accumulation are counted as two operations. Thus, under this normalization, the switched-capacitor SRAM IMC macro achieves approximately $24.9\times$ higher energy efficiency, highlighting the substantial energy-efficiency potential of analog IMC for MatMul.

\section{Existing Clipping Methods for LLM and Analog IMC Quantization}
\label{appx:existing-clipping}

Existing clipping methods span both operand-range calibration for digital LLM quantization and range optimization for analog IMC. Prior operand-clipping approaches typically optimize activation and weight ranges independently or sequentially, while IMC-specific methods either incorporate clipping through hardware-aware retraining or directly narrow the ADC input range. \methodname{} instead jointly optimizes activation and weight clipping through their coupled effect on MatMul output error using an efficient analytical calibration procedure. We summarize the relevant prior work below.

\subsection{Clipping Calibration for LLM Quantization}
\label{appx:llm-clipping}

\paragraph{Activation Clipping.}

ACIQ \citep{banner2019posttraining} derives analytical clipping thresholds for activations by modeling their distributions and minimizing the resulting quantization MSE, while PACT \citep{choi2018pact} learns an activation clipping threshold during quantization-aware training. For Transformer quantization, Outlier Suppression \citep{wei2022outlier} introduces token-wise clipping to better handle the highly nonuniform activation ranges across tokens. More recent LLM quantization methods combine dynamic per-token quantization with static clipping factors. QuaRot \citep{Quarot} and SpinQuant \citep{Spinquant} determine the activation range dynamically for each token and apply a fixed clipping ratio to the resulting range. This formulation of dynamic ranges with static clipping factors is also used in our activation parameterization.

\paragraph{Weight Clipping.}

For weight clipping, the clipping ranges can be calibrated entirely offline. HPTQ \citep{habi2021hptq} performs per-channel threshold search to minimize weight quantization MSE. For LLMs, OmniQuant \citep{shao2024omniquant} introduces learnable weight clipping optimized through block-wise reconstruction, while SignRound \citep{SignRound} jointly optimizes weight clipping and rounding parameters. Weight clipping is also used in QuaRot \citep{Quarot} and SpinQuant \citep{Spinquant} via MSE-based grid search over candidate weight ranges.

\paragraph{Weight and Activation Clipping.}

Several methods optimize quantization ranges for both weights and activations. QIL \citep{QIL} and TQT \citep{TQT} learn quantization intervals or thresholds for both operands through quantization-aware training. OCTAV \citep{sakr2022octav} derives a Newton--Raphson procedure for computing MSE-optimal clipping scalars for both weight and activation tensors, while optimizing the clipping of each tensor independently. For post-training quantization, EasyQuant \citep{EasyQuant} alternates weight and activation range optimization using layer-output reconstruction, while PrefixQuant \citep{chen2024prefixquant} performs LLM-specific weight and activation calibration through reconstruction-based search. These methods address both operands, but do not directly optimize their coupled contribution to MatMul output error, which is critical for IMC-based hardware.

\subsection{Clipping and Range Optimization for Analog IMC}
\label{appx:imc-clipping}

\paragraph{Hardware-Aware Operand Clipping.}

Operand clipping is not commonly used as an optimization knob in prior IMC quantization work. For example, RAOQ \citep{RAOQ} and Cambricon-CIM \citep{cambricon-cim} mitigate ADC-related error through operand reshaping and coding-base reformulation, respectively, rather than clipping. One notable exception is \citet{rasch2023hardware}, who improve PCM-based analog IMC accuracy by optimizing activation and weight ranges through hardware-aware retraining. Their analog model includes ADC/DAC (digital-to-analog converter) quantization together with PCM-specific nonidealities such as programming variation and analog noise. While this allows clipping to adapt to multiple hardware effects, it requires iterative forward and backward optimization and therefore incurs a high calibration cost.

\paragraph{ADC-Input Clipping.}

Another line of work directly narrows the ADC input range to improve the effective resolution of partial-sum quantization. The Optimal Clipping Criterion (OCC) \citep{OCC} selects an ADC clipping range that balances partial-sum clipping error against ADC rounding error. CIMQ \citep{CIMQ} similarly optimizes partial-sum clipping thresholds through a reparameterized clipping function to reduce the required ADC resolution. At the circuit level, \citet{lee2024switched} support configurable ADC quantization that concentrates quantization levels around the high-probability region of the IMC output distribution.

ADC-input clipping provides a promising way to improve ADC range utilization, but it has two limitations for algorithm-level optimization. First, the ADC range is controlled by a limited number of circuit-level reference voltages and therefore offers substantially fewer degrees of freedom than digital operand clipping, which can use separate activation and weight factors across layers and projections. Second, narrowing the ADC reference range reduces the voltage represented by each quantization level. Analog noise therefore occupies an increasingly large fraction of the ADC step, making the resulting error increasingly difficult to represent with a simple deterministic quantization model. Constructing a reliable algorithmic abstraction for aggressive ADC-range clipping therefore requires incorporating detailed circuit-dependent noise characteristics \citep{grimm2022neural}.

\section{Analytical Error Model}
\label{appx:analytical-error-model}

\subsection{Quantization-Error Decomposition}
\label{appx:quantization-error-decomposition}

Eq.~\eqref{eq:clipping-mse-indicator} introduces the clipping-error second moment and its indicator-based evaluation from calibration samples. Here, we complete the operand-level formulation by deriving the rounding contributions for asymmetric activations and symmetric weights, together with the signed clipping-error first moment required by the MatMul output model. These quantities provide the first- and second-order operand-error statistics propagated through the MatMul in Appendix~\ref{appx:matmul-output-error-surrogate}.

\paragraph{Activation Rounding Error.} Asymmetric activation quantization requires special treatment because rounding the integer zero-point generally displaces the clipping thresholds from the extreme reconstruction levels. The activation quantization scale is
\begin{equation*}
s_x = \frac{c_{x,\mathrm{up}}-c_{x,\mathrm{down}}}{2^{b_x}-1}.
\end{equation*}
Let $z_p^\star$ denote the ideal real-valued zero-point and $z_p=\operatorname{round}(z_p^\star)$ the implemented integer zero-point, with residual
\begin{equation*}
\epsilon_{z_p} = z_p-z_p^\star \in[-1/2,1/2].
\end{equation*}
For in-range activations, the standard high-resolution approximation gives rounding MSE $s_x^2/12$. For values clipped to either boundary, zero-point rounding displaces the corresponding reconstruction level by $\epsilon_{z_p}s_x$, producing boundary rounding error $\epsilon_{z_p}^2s_x^2$. The rounding contribution for a fixed quantization grid can therefore be written as
\begin{equation*}
\begin{aligned}
\mathbb{E}[e_{x,\mathrm{round}}^2]
\approx{}& \frac{s_x^2}{12}\int_{c_{x,\mathrm{down}}}^{c_{x,\mathrm{up}}} p(x)\,dx \\
&+ \epsilon_{z_p}^2s_x^2\left[\int_{-\infty}^{c_{x,\mathrm{down}}} p(x)\,dx + \int_{c_{x,\mathrm{up}}}^{\infty} p(x)\,dx\right].
\end{aligned}
\end{equation*}
Directly retaining $\epsilon_{z_p}$ would introduce discontinuities whenever the rounded zero-point changes. To obtain a smooth analytical model, we average over this fractional zero-point offset by modeling $\epsilon_{z_p}\sim\mathcal{U}(-1/2,1/2)$, for which $\mathbb{E}[\epsilon_{z_p}^2]=1/12$. The two probability masses then sum to one, yielding
\begin{equation*}
\mathbb{E}[e_{x,\mathrm{round}}^2] \approx \frac{s_x^2}{12}
\end{equation*}
over the full activation distribution, including values mapped to the clipping boundaries.

\paragraph{Weight Rounding Error.} Symmetric weight quantization has a simpler structure because its fixed zero-point of zero makes the clipping thresholds coincide exactly with the extreme reconstruction levels. The weight quantization scale is
\begin{equation*}
s_w = \frac{c_w}{2^{b_w-1}-1}.
\end{equation*}
Consequently, clipped weights incur no additional boundary rounding error, and rounding applies only to weights within $[-c_w,c_w]$. Its contribution is therefore
\begin{equation*}
\mathbb{E}[e_{w,\mathrm{round}}^2] \approx \frac{s_w^2}{12}\int_{-c_w}^{c_w} p(w)\,dw,
\end{equation*}
or equivalently,
\begin{equation*}
\mathbb{E}[e_{w,\mathrm{round}}^2] = \frac{s_w^2}{12}\mathbb{E}\!\left[\mathbf{1}_{|w|\leq c_w}\right].
\end{equation*}
Thus, unlike activation rounding, the weight-rounding term is explicitly weighted by the retained probability mass.

\paragraph{Signed Clipping Error.} While rounding is modeled as zero-mean, clipping generally introduces a nonzero signed error that cannot be characterized by its second moment alone. For a generic operand $z$ clipped to $[c_{\mathrm{down}},c_{\mathrm{up}}]$, its signed clipping mean is
\begin{equation*}
\mathbb{E}[e_{\mathrm{clip}}] = \int_{-\infty}^{c_{\mathrm{down}}}(c_{\mathrm{down}} - z)p(z)\,dz + \int_{c_{\mathrm{up}}}^{\infty}(c_{\mathrm{up}} - z)p(z)\,dz.
\end{equation*}
Following the same transformation used for the clipping second moment in Eq.~\eqref{eq:clipping-mse-indicator}, this becomes
\begin{equation*}
\mathbb{E}[e_{\mathrm{clip}}] = \mathbb{E}\!\left[(c_{\mathrm{down}} - z)\mathbf{1}_{z<c_{\mathrm{down}}} + (c_{\mathrm{up}} - z)\mathbf{1}_{z>c_{\mathrm{up}}}\right],
\end{equation*}
which can be evaluated directly from calibration samples. For activations, this expectation is taken over the calibration distribution. For an observed weight sample $w_i$, the corresponding signed clipping error is
\begin{equation*}
e_{w,\mathrm{clip},i} = (c_w-w_i)\mathbf{1}_{w_i>c_w} + (-c_w-w_i)\mathbf{1}_{w_i<-c_w}.
\end{equation*}
These signed quantities are retained because clipping bias can accumulate coherently across the MatMul reduction.

\paragraph{Combined Operand Statistics.} Combining the rounding and clipping contributions gives the operand-error moments required by the output-space surrogate, which are estimated from the available activation and weight samples. Under the zero-mean rounding model and the phase-averaged activation approximation,
\begin{equation*}
\begin{aligned}
\mathbb{E}[e_x]
&\approx \mathbb{E}[e_{x,\mathrm{clip}}], \\
\mathbb{E}[e_x^2]
&\approx \frac{s_x^2}{12} + \mathbb{E}[e_{x,\mathrm{clip}}^2],
\end{aligned}
\end{equation*}
while for symmetric weights,
\begin{equation*}
\begin{aligned}
\mathbb{E}[e_{w,i}]
&\approx e_{w,\mathrm{clip},i}, \\
\mathbb{E}[e_{w,i}^2]
&\approx \frac{s_w^2}{12}\mathbf{1}_{|w_i|\leq c_w} + e_{w,\mathrm{clip},i}^2.
\end{aligned}
\end{equation*}
Finally, substituting the clipping parameterization in Eq.~\eqref{eq:clipping-parameterization},
\begin{equation*}
\begin{aligned}
(c_{x,\mathrm{down}},c_{x,\mathrm{up}})
&= (\beta M_x^-,\gamma M_x^+), \\
c_w
&= \alpha M_w,
\end{aligned}
\end{equation*}
gives
\begin{equation*}
\begin{aligned}
s_x
&= \frac{\gamma M_x^+-\beta M_x^-}{2^{b_x}-1}, \\
s_w
&= \frac{\alpha M_w}{2^{b_w-1}-1},
\end{aligned}
\end{equation*}
making all required operand-error statistics explicit functions of $(\gamma,\beta,\alpha)$. These statistics form the inputs to the MatMul output-error surrogate derived next.

\subsection{MatMul Output-Error Surrogate}
\label{appx:matmul-output-error-surrogate}

Eq.~\eqref{eq:surrogate-objective} gives the MatMul output-error objective optimized by \methodname{}. Here, we provide additional details for the approximations underlying the operand-induced output error and validate them empirically. Throughout this analysis, expectations and covariances refer to the underlying operand distributions and quantization-noise model, and are estimated from calibration activations and pretrained weights.

\paragraph{Cross-Coordinate Error Decorrelation.} The exact operand-induced MatMul error satisfies
\begin{equation*}
\mathbb{E}\!\left[\left(\sum_i \delta y_i\right)^2\right] = \sum_i \mathbb{E}[\delta y_i^2] + 2\sum_{i<j}\mathbb{E}[\delta y_i\delta y_j].
\end{equation*}
The operand-output approximation in Eq.~\eqref{eq:operand-output-mse} assumes that the centered MAC errors are approximately pairwise uncorrelated across the reduction dimension,
\begin{equation*}
\operatorname{Cov}(\delta y_i,\delta y_j) = \mathbb{E}[\delta y_i\delta y_j] - \mathbb{E}[\delta y_i]\mathbb{E}[\delta y_j] \approx 0, \qquad i\neq j.
\end{equation*}
Under this assumption,
\begin{equation*}
\mathbb{E}\!\left[\left(\sum_i \delta y_i\right)^2\right] \approx \sum_i \mathbb{E}[\delta y_i^2] + \left(\sum_i\mathbb{E}[\delta y_i]\right)^2 - \sum_i\mathbb{E}[\delta y_i]^2,
\end{equation*}
which yields the diagonal and signed-bias contributions in Eq.~\eqref{eq:surrogate-objective}.

\begin{figure*}[t]
    \centering
    \includegraphics[width=\textwidth]{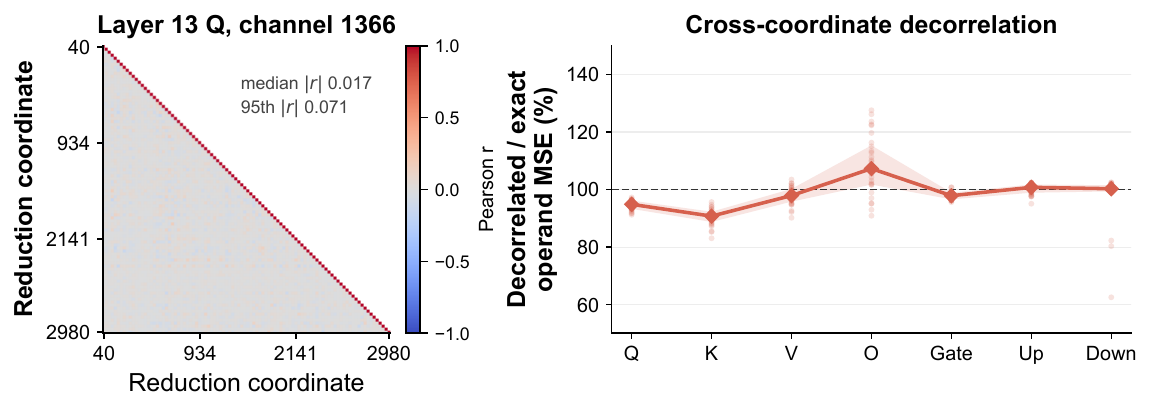}
    \caption{Empirical validation of cross-coordinate error decorrelation on LLaMA-3.2-3B. \textbf{Left:} Pairwise Pearson correlations between centered MAC errors for a representative output channel in the layer-13 $q$ projection. \textbf{Right:} Ratio of the decorrelated approximation to the exact operand MSE across decoder layers. Faint points denote individual layers, diamonds denote the median, shaded regions indicate the 25th--75th percentiles, and the dashed line denotes exact agreement.}
    \label{fig:cross-coordinate-validation}
\end{figure*}

We empirically validate this approximation on LLaMA-3.2-3B. As shown in Fig.~\ref{fig:cross-coordinate-validation} (left), pairwise correlations between centered MAC errors are small for the representative layer-13 $q$ projection, with median $|r|=0.017$ and 95th-percentile $|r|=0.071$. We further measure
\begin{equation*}
100\times
\frac{
\sum_i \mathbb{E}[\delta y_i^2]
+
\left(\sum_i\mathbb{E}[\delta y_i]\right)^2
-
\sum_i\mathbb{E}[\delta y_i]^2
}{
\mathbb{E}\!\left[\left(\sum_i\delta y_i\right)^2\right]
},
\end{equation*}
i.e., the decorrelated approximation relative to the exact operand MSE. Figure~\ref{fig:cross-coordinate-validation} (right) shows that this ratio remains reasonably close to $100\%$ across all projection types and decoder layers, supporting the use of the pairwise-decorrelation approximation. Crucially, this approximation replaces the $O(d^2)$ cross-coordinate joint statistics with per-coordinate moments and reductions, enabling efficient evaluation of the surrogate and its derivatives during calibration.

\paragraph{Per-Coordinate Second-Moment Approximation.} For the per-MAC error in Eq.~\eqref{eq:mac-error},
\begin{equation*}
\delta y_i = x_i e_{w,i} + w_i e_{x,i} + e_{x,i}e_{w,i},
\end{equation*}
the exact second moment contains six terms,
\begin{equation*}
\begin{aligned}
\mathbb{E}[\delta y_i^2]
={}&
w_i^2\mathbb{E}[e_{x,i}^2]
+ \mathbb{E}[x_i^2e_{w,i}^2]
+ \mathbb{E}[e_{x,i}^2e_{w,i}^2]
\\
&+ 2w_i\mathbb{E}[x_i e_{x,i}e_{w,i}]
+ 2\mathbb{E}[x_i e_{x,i}e_{w,i}^2]
+ 2w_i\mathbb{E}[e_{x,i}^2e_{w,i}].
\end{aligned}
\end{equation*}

To obtain a tractable analytical objective, we retain the two leading signal--noise contributions summarized in Eq.~\eqref{eq:mac-second-moment},
\begin{equation*}
\mathbb{E}[\delta y_i^2]
\approx
w_i^2\mathbb{E}[e_{x,i}^2]
+
\mathbb{E}[x_i^2]\mathbb{E}[e_{w,i}^2].
\end{equation*}
The remaining four terms contain mixed activation--weight error products or higher-order error moments whose direct evaluation would require additional joint statistics.

\begin{figure*}[t]
    \centering
    \includegraphics[width=\textwidth]{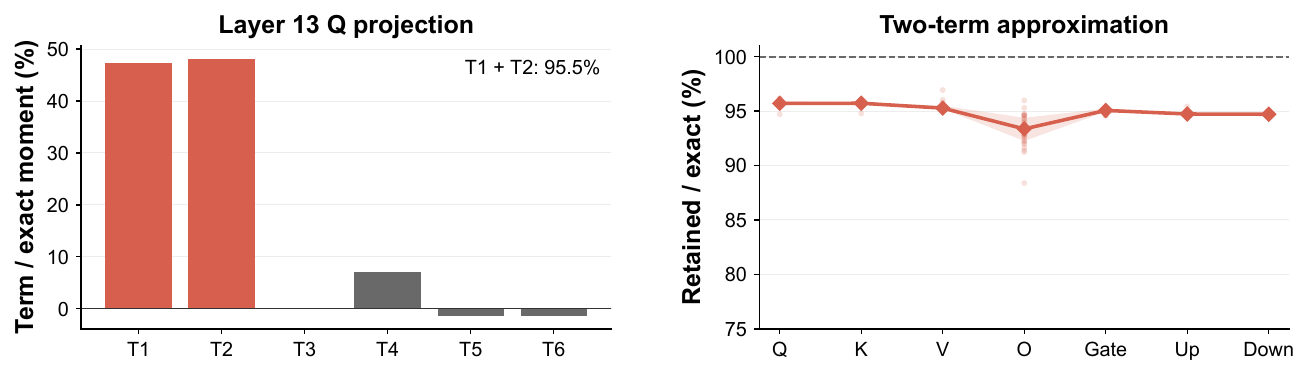}
    \caption{Empirical validation of the two-term second-moment approximation on LLaMA-3.2-3B. \textbf{Left:} Signed contributions of the six exact terms for the representative layer-13 $q$ projection, where the two retained terms account for $95.5\%$ of the total. \textbf{Right:} Ratio of the retained two-term sum to the exact six-term sum across decoder layers. Faint points denote individual layers, diamonds denote the median, shaded regions indicate the 25th--75th percentiles, and the dashed line denotes exact agreement.}
    \label{fig:two-term-approximation}
\end{figure*}

To evaluate whether retaining only the two leading terms provides a sufficiently accurate approximation, we directly compare their empirical sum against the exact six-term second moment on LLaMA-3.2-3B. As shown in Fig.~\ref{fig:two-term-approximation} (left), the retained terms account for $95.5\%$ of the exact second moment in the representative layer-13 $q$ projection. Across all 28 decoder layers, Fig.~\ref{fig:two-term-approximation} (right) shows that the retained-to-exact ratio remains close to $100\%$ for every projection type, with median values of approximately $93$--$96\%$. This roughly $5\%$ mismatch is an acceptable trade-off for a substantially simpler loss surrogate with a more benign landscape for optimization.

\begin{figure*}[t]
    \centering
    \includegraphics[width=\textwidth]{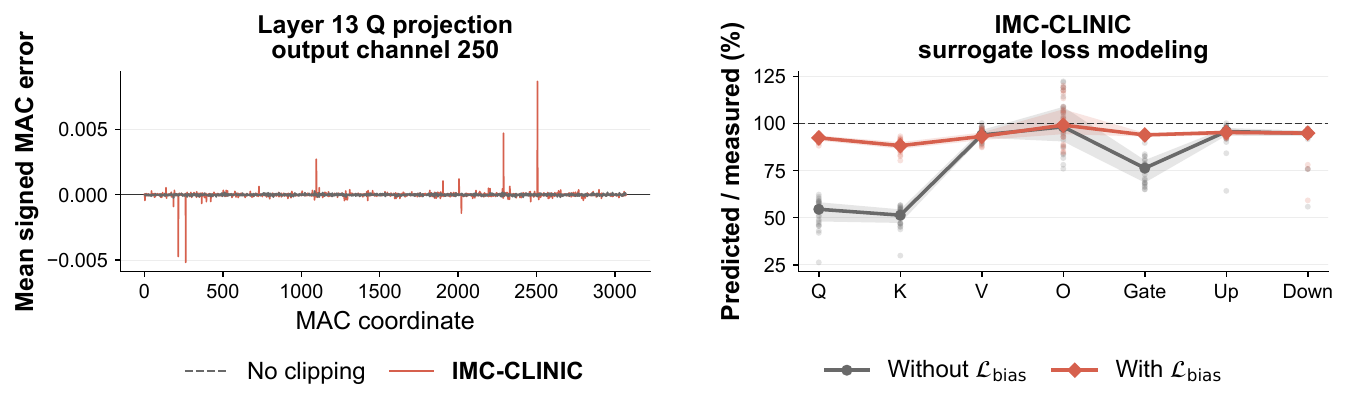}
    \caption{Empirical validation of the signed-bias term on LLaMA-3.2-3B. \textbf{Left:} Mean signed MAC error $\mathbb{E}[\delta y_i]$ across MAC coordinates for a representative output channel in the layer-13 $q$ projection. \textbf{Right:} Ratio of predicted to measured operand-output MSE across decoder layers, comparing the surrogate without and with $\mathcal{L}_{\mathrm{bias}}$. Faint points denote individual layers, diamonds denote the median, shaded regions indicate the 25th--75th percentiles, and the dashed line denotes exact agreement.}
    \label{fig:clip-bias}
\end{figure*}

\paragraph{Signed-Bias Contribution.} Clipping errors are not necessarily zero mean: structured LLM outliers can cause a small number of MAC coordinates to be repeatedly clipped in the same direction, producing large signed mean errors. As shown in Fig.~\ref{fig:clip-bias} (left), most MAC coordinates remain near zero, while a few exhibit pronounced clipping-induced bias. These signed errors accumulate across the MatMul reduction and can materially affect the output error.

Figure~\ref{fig:clip-bias} (right) confirms this effect at the model scale. Omitting $\mathcal{L}_{\mathrm{bias}}$ substantially underestimates the measured operand-output MSE, especially for the $q$ and $k$ projections, where the prediction is lower by roughly $45$--$50\%$. Including $\mathcal{L}_{\mathrm{bias}}$ brings the analytical prediction much closer to the measured error across all projection types. We therefore retain $\mathcal{L}_{\mathrm{bias}}$ in the surrogate objective.

\paragraph{Bit-Sliced ADC Error.} Our IMC mapping follows \citet{lee2024switched,cambricon-cim} to decompose each 8-bit activation and weight operand into high and low 4-bit slices, producing four independently digitized $4\mathrm{b}\times4\mathrm{b}$ partial products $(HH,HL,LH,LL)$. For a fixed IMC row tile, let $\Delta_{\mathrm{slice}}$ denote the physical ADC quantization step for each slice-level conversion. The digitized slice products are recombined according to their binary significance as
\begin{equation*}
p_{-1,1}
=
2^8 p_{HH}
+
2^4\left(p_{HL}+p_{LH}\right)
+
p_{LL},
\end{equation*}
where $p_{-1,1}$ denotes the reconstructed partial sum in the circuit's $+1/-1$ representation \citep{programmableIMC}. Under the standard high-resolution quantization model, the four slice-level ADC errors are independent and zero mean with variance $\Delta_{\mathrm{slice}}^2/12$. Their variances therefore add according to the squared reconstruction coefficients,
\begin{equation*}
\operatorname{Var}(n_{-1,1})
=
\frac{\Delta_{\mathrm{slice}}^2}{12}
\left(2^{16}+2\cdot 2^8+1\right)
=
\frac{\left(257\,\Delta_{\mathrm{slice}}\right)^2}{12}.
\end{equation*}
The circuit represents integer operands through
\begin{equation*}
x_{-1,1}=2x_{\mathrm{int}}+1,
\qquad
w_{-1,1}=2w_{\mathrm{int}}+1,
\end{equation*}
so recovering the conventional integer dot product scales the reconstructed partial sum by $1/4$, with the remaining cross terms corrected digitally. Consequently, the ADC error itself is also scaled by $1/4$. We therefore define the effective reconstructed ADC-noise scale
\begin{equation}
\Delta_{\mathrm{ADC}}
\equiv
\frac{257}{4}\Delta_{\mathrm{slice}},
\label{eq:effective-adc-step}
\end{equation}
such that one reconstructed row-tile partial sum has ADC-error variance $\Delta_{\mathrm{ADC}}^2/12$. For a MatMul partitioned across $K$ independently digitized row tiles, these variances add, and mapping the accumulated error back to the floating-point domain introduces the operand scale product $s_xs_w$, yielding
\begin{equation*}
\mathcal{L}_{\mathrm{ADC}}
=
K\frac{\Delta_{\mathrm{ADC}}^2}{12}(s_xs_w)^2,
\end{equation*}
which recovers the ADC term in Eq.~\eqref{eq:adc-loss}.

\begin{figure*}[t]
    \centering
    \includegraphics[width=\textwidth]{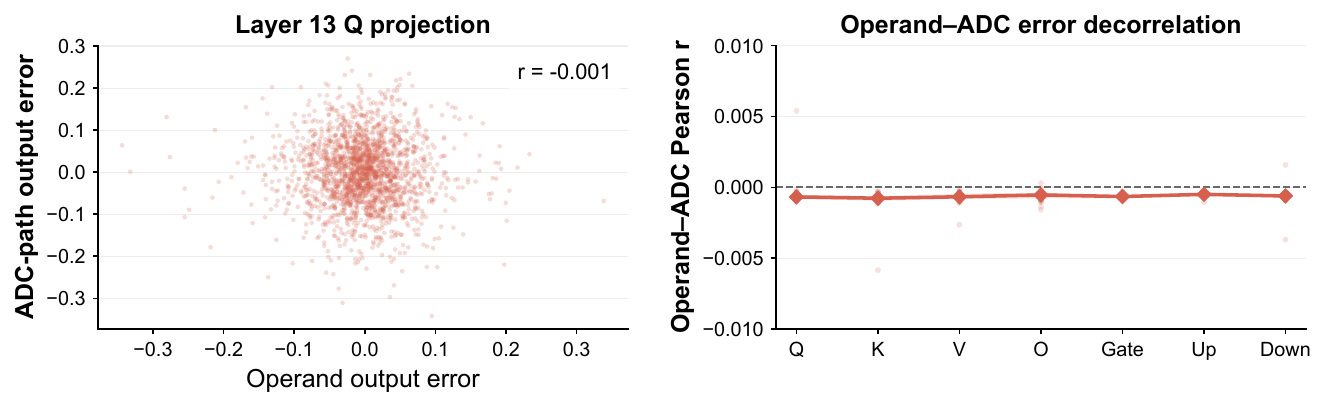}
    \caption{Empirical validation of operand--ADC error decorrelation on LLaMA-3.2-3B. \textbf{Left:} Operand-induced output error versus the additional ADC error for the representative layer-13 $q$ projection, with Pearson correlation $r=-0.001$. \textbf{Right:} Pearson correlation across all 28 decoder layers and projection types. Faint points denote individual layers, diamonds denote the median, shaded regions indicate the 25th--75th percentiles, and the dashed line denotes zero correlation.}
    \label{fig:adc-error-correlation}
\end{figure*}

\paragraph{Operand--ADC Error Decorrelation.} The additive treatment of operand and ADC quantization errors assumes that their correlation is negligible. We therefore directly measure the Pearson correlation between the operand-induced output error and the additional ADC error on LLaMA-3.2-3B. As shown in Fig.~\ref{fig:adc-error-correlation} (left), the representative layer-13 $q$ projection has correlation $r=-0.001$. Across all 28 decoder layers, Fig.~\ref{fig:adc-error-correlation} (right) shows that the correlation remains close to zero for every projection type. These results support the assumption that operand and ADC quantization errors are approximately uncorrelated, allowing their MSE contributions to be added in the surrogate objective.

Together, these validations assess the approximations used to construct the analytical output-error objective while avoiding explicit estimation of cross-coordinate and higher-order joint statistics.

\subsection{Gradient and Hessian Derivations}
\label{appx:gradient-hessian}

Eq.~\eqref{eq:surrogate-objective} defines the analytical output-error objective optimized by \methodname{}. Here, we derive the corresponding first- and second-order derivatives required for second-order calibration. We model the calibration activations and pretrained weights as samples from underlying operand distributions. We first derive the derivatives of the corresponding distributions with respect to the clipping factors, and then estimate the resulting expectations, tail probabilities, and boundary densities from the available samples (to avoid differentiating finite-sample indicator functions directly).

\paragraph{Derivative Setup.} The objective is
\begin{equation*}
\mathcal{L} = \mathcal{L}_{\mathrm{diag}} + \mathcal{L}_{\mathrm{bias}} + \mathcal{L}_{\mathrm{ADC}},
\end{equation*}
parameterized by the clipping factors $\boldsymbol{\theta}=(\gamma,\beta,\alpha)$. Throughout this subsection, subscripts denote partial derivatives,
\begin{equation*}
\begin{aligned}
(\cdot)_p
&\equiv \frac{\partial(\cdot)}{\partial p},
\\
(\cdot)_{pq}
&\equiv \frac{\partial^2(\cdot)}{\partial p\,\partial q},
\qquad
p,q\in\{\gamma,\beta,\alpha\}.
\end{aligned}
\end{equation*}
We otherwise reuse the operand-error quantities introduced in Appendix~\ref{appx:quantization-error-decomposition}, avoiding additional intermediate notation.

\paragraph{Operand-Error Derivatives.} The operand-error statistics in Appendix~\ref{appx:quantization-error-decomposition} depend on the clipping factors through the clipping thresholds and quantization scales. Since these quantities are linear in $(\gamma,\beta,\alpha)$, their derivatives can be obtained analytically using Leibniz's integral rule.

For the signed activation clipping error,
\begin{equation*}
\begin{aligned}
\left(\mathbb{E}[e_{x,\mathrm{clip}}]\right)_{\gamma}
&= M_x^+\mathbb{E}\!\left[\mathbf{1}_{x>c_{x,\mathrm{up}}}\right],
\\
\left(\mathbb{E}[e_{x,\mathrm{clip}}]\right)_{\beta}
&= M_x^-\mathbb{E}\!\left[\mathbf{1}_{x<c_{x,\mathrm{down}}}\right],
\\
\left(\mathbb{E}[e_{x,\mathrm{clip}}]\right)_{\gamma\gamma}
&= -(M_x^+)^2 p_x(c_{x,\mathrm{up}}),
\\
\left(\mathbb{E}[e_{x,\mathrm{clip}}]\right)_{\beta\beta}
&= (M_x^-)^2 p_x(c_{x,\mathrm{down}}),
\\
\left(\mathbb{E}[e_{x,\mathrm{clip}}]\right)_{\gamma\beta}
&= 0.
\end{aligned}
\end{equation*}
where $p_x(\cdot)$ denotes the activation density. Thus, the first derivatives depend on the probability mass beyond the clipping thresholds, whereas the second derivatives depend on the density at the moving boundaries.

For the activation clipping-error second moment,
\begin{equation*}
\begin{aligned}
\left(\mathbb{E}[e_{x,\mathrm{clip}}^2]\right)_{\gamma}
&= 2M_x^+\mathbb{E}\!\left[e_{x,\mathrm{clip}}\mathbf{1}_{x>c_{x,\mathrm{up}}}\right],
\\
\left(\mathbb{E}[e_{x,\mathrm{clip}}^2]\right)_{\beta}
&= 2M_x^-\mathbb{E}\!\left[e_{x,\mathrm{clip}}\mathbf{1}_{x<c_{x,\mathrm{down}}}\right],
\\
\left(\mathbb{E}[e_{x,\mathrm{clip}}^2]\right)_{\gamma\gamma}
&= 2(M_x^+)^2\mathbb{E}\!\left[\mathbf{1}_{x>c_{x,\mathrm{up}}}\right],
\\
\left(\mathbb{E}[e_{x,\mathrm{clip}}^2]\right)_{\beta\beta}
&= 2(M_x^-)^2\mathbb{E}\!\left[\mathbf{1}_{x<c_{x,\mathrm{down}}}\right],
\\
\left(\mathbb{E}[e_{x,\mathrm{clip}}^2]\right)_{\gamma\beta}
&= 0.
\end{aligned}
\end{equation*}
The boundary terms vanish because the clipping residual is zero at the corresponding threshold. Using $\mathbb{E}[e_x^2] \approx s_x^2/12 + \mathbb{E}[e_{x,\mathrm{clip}}^2]$, we obtain, for $p,q\in\{\gamma,\beta\}$,
\begin{equation*}
\begin{aligned}
\left(\mathbb{E}[e_x^2]\right)_p
&=
\frac{s_x s_{x,p}}{6}
+
\left(\mathbb{E}[e_{x,\mathrm{clip}}^2]\right)_p, \\
\left(\mathbb{E}[e_x^2]\right)_{pq}
&=
\frac{s_{x,p}s_{x,q}}{6}
+
\left(\mathbb{E}[e_{x,\mathrm{clip}}^2]\right)_{pq},
\end{aligned}
\end{equation*}
where $s_{x,\gamma}=M_x^+/(2^{b_x}-1)$ and $s_{x,\beta}=-M_x^-/(2^{b_x}-1)$.

For symmetric weights, the first derivative of the signed clipping error for an observed weight sample is
\begin{equation*}
(e_{w,\mathrm{clip},i})_{\alpha} = M_w\left(\mathbf{1}_{w_i>c_w}-\mathbf{1}_{w_i<-c_w}\right).
\end{equation*}
Averaging over the underlying weight distribution gives
\begin{equation*}
\left(\mathbb{E}_w[e_{w,\mathrm{clip}}]\right)_{\alpha} = M_w\left[\mathbb{P}(w>c_w)-\mathbb{P}(w<-c_w)\right],
\end{equation*}
and
\begin{equation*}
\left(\mathbb{E}_w[e_{w,\mathrm{clip}}]\right)_{\alpha\alpha} = M_w^2\left[p_w(-c_w)-p_w(c_w)\right],
\end{equation*}
where $p_w(\cdot)$ denotes the weight density for the corresponding quantization group. Thus, as with activations, the first derivative depends on the probability mass beyond the clipping boundaries, while the second-order curvature depends on the density at the moving boundaries.

For the weight-error second moment, in-range rounding is modeled using the statistical theory of quantization as zero-mean noise with variance $s_w^2/12$, while weights outside $[-c_w,c_w]$ incur clipping error. Care is required when differentiating this model because the zero-mean noise model and the exact quantizer differ at the clipping thresholds. In the underlying symmetric quantizer, each clipping threshold is itself an outer quantization level, so the rounding residual is zero at $w=\pm c_w$. Under an infinitesimal threshold change $\delta c_w$, newly entering weights have $O(\delta c_w)$ rounding residual and, for a locally bounded density, occupy $O(\delta c_w)$ probability mass. Their total squared-error contribution is therefore $O(\delta c_w^3)$, which contributes neither to the first nor the second derivative as $\delta c_w \to 0$. We therefore differentiate the in-range rounding variance with respect to $s_w$ before estimating the resulting quantities from the pretrained weight.

This gives
\begin{equation*}
\left(\mathbb{E}[e_{w,i}^2]\right)_{\alpha}
\approx
\frac{s_w s_{w,\alpha}}{6}
\mathbf{1}_{|w_i|\leq c_w}
+
2e_{w,\mathrm{clip},i}
(e_{w,\mathrm{clip},i})_{\alpha},
\end{equation*}
and
\begin{equation*}
\begin{aligned}
\left(\mathbb{E}[e_{w,i}^2]\right)_{\alpha\alpha}
&\approx
\frac{s_{w,\alpha}^2}{6}
\mathbf{1}_{|w_i|\leq c_w}
+
2(e_{w,\mathrm{clip},i})_{\alpha}^2, \\
s_{w,\alpha}
&=
\frac{M_w}{2^{b_w-1}-1}.
\end{aligned}
\end{equation*}

\paragraph{Signed Mean-Error Derivatives.} The signed mean error contributed by the $i$-th MAC is defined in Eq.~\eqref{eq:mac-signed-mean} as
\begin{equation*}
\mathbb{E}[\delta y_i] = w_i\mathbb{E}[e_{x,\mathrm{clip},i}] + e_{w,\mathrm{clip},i}\left(\mathbb{E}[x_i] + \mathbb{E}[e_{x,\mathrm{clip},i}]\right).
\end{equation*}
The activation clipping statistics depend only on $(\gamma,\beta)$, whereas the weight clipping error depends only on $\alpha$. The first derivatives are therefore
\begin{equation*}
\begin{aligned}
\left(\mathbb{E}[\delta y_i]\right)_{\gamma}
&= \left(w_i+e_{w,\mathrm{clip},i}\right)\left(\mathbb{E}[e_{x,\mathrm{clip},i}]\right)_{\gamma}, \\
\left(\mathbb{E}[\delta y_i]\right)_{\beta}
&= \left(w_i+e_{w,\mathrm{clip},i}\right)\left(\mathbb{E}[e_{x,\mathrm{clip},i}]\right)_{\beta}, \\
\left(\mathbb{E}[\delta y_i]\right)_{\alpha}
&= (e_{w,\mathrm{clip},i})_{\alpha}\left(\mathbb{E}[x_i] + \mathbb{E}[e_{x,\mathrm{clip},i}]\right).
\end{aligned}
\end{equation*}

Applying the product rule gives the six unique second derivatives,
\begin{equation*}
\begin{aligned}
\left(\mathbb{E}[\delta y_i]\right)_{\gamma\gamma}
&= \left(w_i+e_{w,\mathrm{clip},i}\right)\left(\mathbb{E}[e_{x,\mathrm{clip},i}]\right)_{\gamma\gamma}, \\
\left(\mathbb{E}[\delta y_i]\right)_{\beta\beta}
&= \left(w_i+e_{w,\mathrm{clip},i}\right)\left(\mathbb{E}[e_{x,\mathrm{clip},i}]\right)_{\beta\beta}, \\
\left(\mathbb{E}[\delta y_i]\right)_{\gamma\beta}
&= \left(w_i+e_{w,\mathrm{clip},i}\right)\left(\mathbb{E}[e_{x,\mathrm{clip},i}]\right)_{\gamma\beta} = 0, \\
\left(\mathbb{E}[\delta y_i]\right)_{\gamma\alpha}
&= (e_{w,\mathrm{clip},i})_{\alpha}\left(\mathbb{E}[e_{x,\mathrm{clip},i}]\right)_{\gamma}, \\
\left(\mathbb{E}[\delta y_i]\right)_{\beta\alpha}
&= (e_{w,\mathrm{clip},i})_{\alpha}\left(\mathbb{E}[e_{x,\mathrm{clip},i}]\right)_{\beta}, \\
\left(\mathbb{E}[\delta y_i]\right)_{\alpha\alpha}
&= M_w^2\left[p_w(-c_w)-p_w(c_w)\right]\left(\mathbb{E}[x_i] + \mathbb{E}[e_{x,\mathrm{clip},i}]\right).
\end{aligned}
\end{equation*}
The mixed $\gamma$--$\alpha$ and $\beta$--$\alpha$ derivatives explicitly capture the coupling between activation and weight clipping. The density-dependent curvatures of the underlying activation and weight distributions propagate through these expressions into the Hessian of the bias term.

\paragraph{Diagonal and Bias Loss Derivatives.} The diagonal loss is
\begin{equation*}
\mathcal{L}_{\mathrm{diag}} = \sum_i\left(w_i^2\mathbb{E}[e_{x,i}^2] + \mathbb{E}[x_i^2]\mathbb{E}[e_{w,i}^2]\right).
\end{equation*}
Because the activation-error statistics depend only on $(\gamma,\beta)$ and the weight-error statistics depend only on $\alpha$, its first derivatives are
\begin{equation*}
\begin{aligned}
(\mathcal{L}_{\mathrm{diag}})_{\gamma}
&= \sum_i w_i^2\left(\mathbb{E}[e_{x,i}^2]\right)_{\gamma}, \\
(\mathcal{L}_{\mathrm{diag}})_{\beta}
&= \sum_i w_i^2\left(\mathbb{E}[e_{x,i}^2]\right)_{\beta}, \\
(\mathcal{L}_{\mathrm{diag}})_{\alpha}
&= \sum_i \mathbb{E}[x_i^2]\left(\mathbb{E}[e_{w,i}^2]\right)_{\alpha}.
\end{aligned}
\end{equation*}
The nonzero second derivatives are
\begin{equation*}
\begin{aligned}
(\mathcal{L}_{\mathrm{diag}})_{\gamma\gamma}
&= \sum_i w_i^2\left(\mathbb{E}[e_{x,i}^2]\right)_{\gamma\gamma}, \\
(\mathcal{L}_{\mathrm{diag}})_{\beta\beta}
&= \sum_i w_i^2\left(\mathbb{E}[e_{x,i}^2]\right)_{\beta\beta}, \\
(\mathcal{L}_{\mathrm{diag}})_{\gamma\beta}
&= \sum_i w_i^2\left(\mathbb{E}[e_{x,i}^2]\right)_{\gamma\beta}, \\
(\mathcal{L}_{\mathrm{diag}})_{\alpha\alpha}
&= \sum_i \mathbb{E}[x_i^2]\left(\mathbb{E}[e_{w,i}^2]\right)_{\alpha\alpha},
\end{aligned}
\end{equation*}
with
\begin{equation*}
(\mathcal{L}_{\mathrm{diag}})_{\gamma\alpha} = (\mathcal{L}_{\mathrm{diag}})_{\beta\alpha} = 0.
\end{equation*}

The bias term is
\begin{equation*}
\mathcal{L}_{\mathrm{bias}} = \left(\sum_i\mathbb{E}[\delta y_i]\right)^2 - \sum_i\mathbb{E}[\delta y_i]^2.
\end{equation*}
For $p\in\{\gamma,\beta,\alpha\}$, its first derivative is
\begin{equation*}
\begin{aligned}
(\mathcal{L}_{\mathrm{bias}})_p
={}& 2\left(\sum_i\mathbb{E}[\delta y_i]\right)\left(\sum_i\left(\mathbb{E}[\delta y_i]\right)_p\right) \\
&- 2\sum_i\mathbb{E}[\delta y_i]\left(\mathbb{E}[\delta y_i]\right)_p.
\end{aligned}
\end{equation*}
For $p,q\in\{\gamma,\beta,\alpha\}$, its second derivatives are
\begin{equation*}
\begin{aligned}
(\mathcal{L}_{\mathrm{bias}})_{pq}
={}& 2\left(\sum_i\left(\mathbb{E}[\delta y_i]\right)_p\right)\left(\sum_i\left(\mathbb{E}[\delta y_i]\right)_q\right) \\
&+ 2\left(\sum_i\mathbb{E}[\delta y_i]\right)\left(\sum_i\left(\mathbb{E}[\delta y_i]\right)_{pq}\right) \\
&- 2\sum_i\left[\left(\mathbb{E}[\delta y_i]\right)_p\left(\mathbb{E}[\delta y_i]\right)_q + \mathbb{E}[\delta y_i]\left(\mathbb{E}[\delta y_i]\right)_{pq}\right].
\end{aligned}
\end{equation*}
Unlike $\mathcal{L}_{\mathrm{diag}}$, the bias term generally has nonzero $\gamma$--$\alpha$ and $\beta$--$\alpha$ Hessian entries through the mixed derivatives of $\mathbb{E}[\delta y_i]$, thereby coupling activation and weight clipping.

\paragraph{ADC Gradient and Hessian.} The ADC contribution is
\begin{equation*}
\mathcal{L}_{\mathrm{ADC}} = K\frac{\Delta_{\mathrm{ADC}}^2}{12}s_x^2 s_w^2.
\end{equation*}
Since $s_x$ is linear in $(\gamma,\beta)$ and $s_w$ is linear in $\alpha$, their second derivatives vanish. The gradient is therefore
\begin{equation*}
\begin{aligned}
(\mathcal{L}_{\mathrm{ADC}})_{\gamma}
&= K\frac{\Delta_{\mathrm{ADC}}^2}{6}s_x s_{x,\gamma}s_w^2, \\
(\mathcal{L}_{\mathrm{ADC}})_{\beta}
&= K\frac{\Delta_{\mathrm{ADC}}^2}{6}s_x s_{x,\beta}s_w^2, \\
(\mathcal{L}_{\mathrm{ADC}})_{\alpha}
&= K\frac{\Delta_{\mathrm{ADC}}^2}{6}s_x^2s_ws_{w,\alpha}.
\end{aligned}
\end{equation*}
The corresponding Hessian entries are
\begin{equation*}
\begin{aligned}
(\mathcal{L}_{\mathrm{ADC}})_{\gamma\gamma}
&= K\frac{\Delta_{\mathrm{ADC}}^2}{6}s_{x,\gamma}^2s_w^2, \\
(\mathcal{L}_{\mathrm{ADC}})_{\beta\beta}
&= K\frac{\Delta_{\mathrm{ADC}}^2}{6}s_{x,\beta}^2s_w^2, \\
(\mathcal{L}_{\mathrm{ADC}})_{\gamma\beta}
&= K\frac{\Delta_{\mathrm{ADC}}^2}{6}s_{x,\gamma}s_{x,\beta}s_w^2, \\
(\mathcal{L}_{\mathrm{ADC}})_{\gamma\alpha}
&= K\frac{\Delta_{\mathrm{ADC}}^2}{3}s_xs_{x,\gamma}s_ws_{w,\alpha}, \\
(\mathcal{L}_{\mathrm{ADC}})_{\beta\alpha}
&= K\frac{\Delta_{\mathrm{ADC}}^2}{3}s_xs_{x,\beta}s_ws_{w,\alpha}, \\
(\mathcal{L}_{\mathrm{ADC}})_{\alpha\alpha}
&= K\frac{\Delta_{\mathrm{ADC}}^2}{6}s_x^2s_{w,\alpha}^2.
\end{aligned}
\end{equation*}
Equivalently, ordering the parameters as $(\gamma,\beta,\alpha)$,
\begin{equation*}
\nabla^2\mathcal{L}_{\mathrm{ADC}} = K\frac{\Delta_{\mathrm{ADC}}^2}{6}
\begin{bmatrix}
s_{x,\gamma}^2s_w^2 & s_{x,\gamma}s_{x,\beta}s_w^2 & 2s_xs_{x,\gamma}s_ws_{w,\alpha} \\
s_{x,\gamma}s_{x,\beta}s_w^2 & s_{x,\beta}^2s_w^2 & 2s_xs_{x,\beta}s_ws_{w,\alpha} \\
2s_xs_{x,\gamma}s_ws_{w,\alpha} & 2s_xs_{x,\beta}s_ws_{w,\alpha} & s_x^2s_{w,\alpha}^2
\end{bmatrix}.
\end{equation*}
Unlike the diagonal operand-error term, the ADC contribution directly introduces $\gamma$--$\alpha$ and $\beta$--$\alpha$ curvature because its output-referred error depends multiplicatively on the activation and weight scales.

\paragraph{Full and Optimization Hessians.} Combining the three loss components gives the analytical gradient
\begin{equation*}
\nabla\mathcal{L} = \nabla\mathcal{L}_{\mathrm{diag}} + \nabla\mathcal{L}_{\mathrm{bias}} + \nabla\mathcal{L}_{\mathrm{ADC}},
\end{equation*}
and the full Hessian of the analytical surrogate
\begin{equation*}
H \equiv \nabla^2\mathcal{L} = \nabla^2\mathcal{L}_{\mathrm{diag}} + \nabla^2\mathcal{L}_{\mathrm{bias}} + \nabla^2\mathcal{L}_{\mathrm{ADC}}.
\end{equation*}
The full Hessian includes the density-sensitive curvature of the signed clipping errors,
\begin{equation*}
\begin{aligned}
\left(\mathbb{E}[e_{x,\mathrm{clip}}]\right)_{\gamma\gamma}
&= -(M_x^+)^2p_x(c_{x,\mathrm{up}}), \\
\left(\mathbb{E}[e_{x,\mathrm{clip}}]\right)_{\beta\beta}
&= (M_x^-)^2p_x(c_{x,\mathrm{down}}),
\end{aligned}
\end{equation*}
and
\begin{equation*}
\left(\mathbb{E}_w[e_{w,\mathrm{clip}}]\right)_{\alpha\alpha} = M_w^2\left[p_w(-c_w)-p_w(c_w)\right],
\end{equation*}
which propagate through the signed mean-error derivatives into $\nabla^2\mathcal{L}_{\mathrm{bias}}$.

During calibration, we use an approximate Hessian $\widetilde H$ obtained by omitting these density-sensitive second-order terms,
\begin{equation}
\widetilde H = H\Bigg|_{
\substack{
(\mathbb{E}[e_{x,\mathrm{clip}}])_{\gamma\gamma}=0,\\
(\mathbb{E}[e_{x,\mathrm{clip}}])_{\beta\beta}=0,\\
(\mathbb{E}_w[e_{w,\mathrm{clip}}])_{\alpha\alpha}=0
}},
\label{eq:approximate-hessian}
\end{equation}
while retaining the full analytical gradient and all remaining curvature, including the clipping-error second moments, rounding-error terms, mixed activation--weight derivatives, and ADC curvature. This avoids relying on pointwise density estimates at moving clipping boundaries during optimization, while preserving the dominant second-order structure of the objective. The full density-aware Hessian $H$ is retained for the loss-landscape analysis in Appendix~\ref{app:psd-region-validation}.

All expectations, tail probabilities, and boundary densities above are estimated from the finite calibration activations and pretrained weights.

\subsection{Empirical Validation of the PSD Region}
\label{app:psd-region-validation}

Our safeguarded calibration method is designed to operate within a locally well-behaved region of the surrogate objective. While the optimizer uses the approximate Hessian derived in Appendix~\ref{appx:gradient-hessian}, we evaluate the full Hessian of the analytical surrogate, \(H\), post hoc to characterize the surrogate loss landscape on real LLM calibration activations and weights. We define a configuration as PSD when
\begin{equation*}
\lambda_{\min}(H) \geq 0.
\end{equation*}

\begin{figure*}[t]
    \centering
    \includegraphics[width=\textwidth]{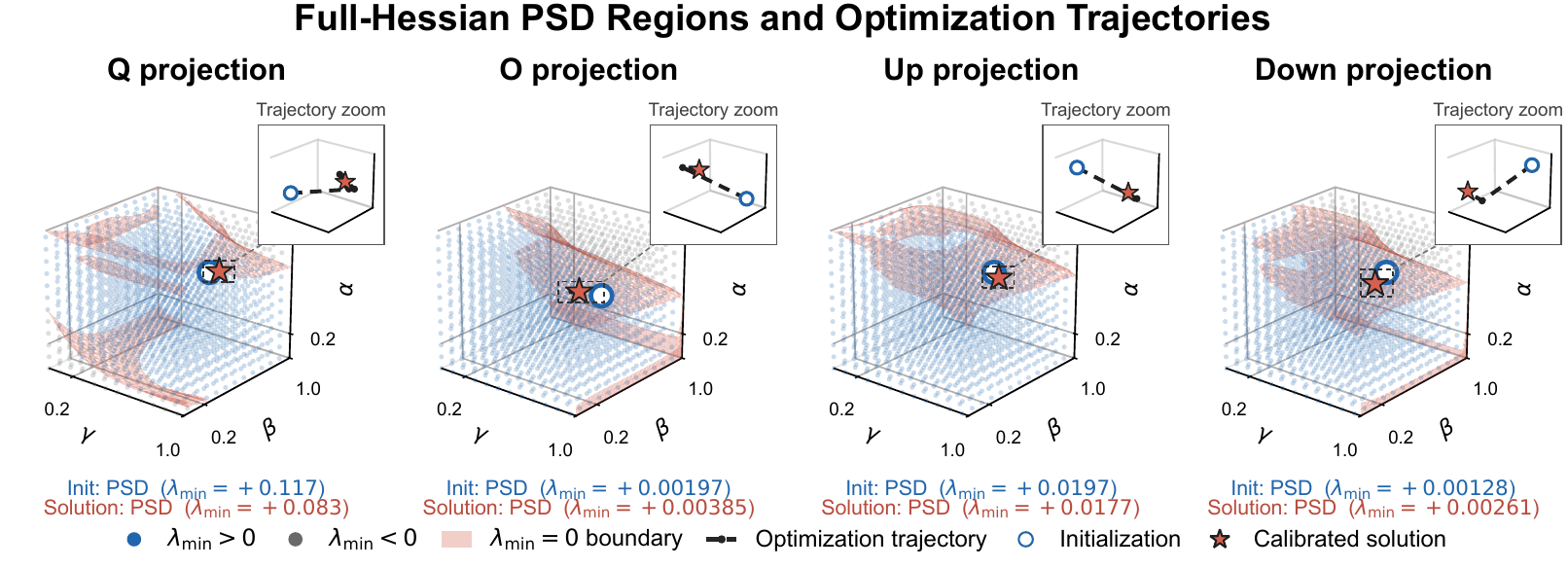}
    \caption{PSD regions of the full surrogate Hessian for the $q$, $o$, $up$, and $down$ projections in decoder block 13 of LLaMA-3.2-3B. Blue points are sampled clipping-factor configurations with $\lambda_{\min}>0$, gray points have $\lambda_{\min}<0$, and the red surface marks the $\lambda_{\min}=0$ boundary. The insets show the optimization trajectories (black dashed lines) from initialization (blue circles) to the calibrated solutions (red stars); both endpoints lie in the PSD region in all four projections.}
    \label{fig:psd-region-layer13}
\end{figure*}

Figure~\ref{fig:psd-region-layer13} visualizes representative PSD regions for different projections in LLaMA-3.2-3B decoder block 13. Across these examples, the sampled PSD configurations form a single connected component that contains the initialization, accepted optimization trajectory, and calibrated solution. The $\lambda_{\min}=0$ boundary separates the PSD region from configurations with indefinite curvature.

To complement the representative landscapes, Table~\ref{tab:psd-region-summary} summarizes the full-Hessian validation across all evaluated projections of LLaMA-3.2-3B and Qwen3-4B. Across both models, every evaluated projection exhibits a single connected PSD component containing the initialization, optimization trajectory, and calibrated solution, providing empirical support for the landscape assumption used by our safeguarded calibration method.

\begin{table}[t]
    \centering
    \caption{Full-Hessian PSD-region validation across all projections of LLaMA-3.2-3B and Qwen3-4B. Entries report the number of projections satisfying each property.}
    \label{tab:psd-region-summary}
    \begin{tabular}{lcc}
        \hline
        \textbf{Property}
        & \textbf{LLaMA-3.2-3B}
        & \textbf{Qwen3-4B} \\
        \hline
        Single connected PSD component
        & 196/196 projs & 252/252 projs \\
        Initialization in PSD component
        & 196/196 projs & 252/252 projs \\
        Calibrated solution in PSD component
        & 196/196 projs & 252/252 projs \\
        Accepted trajectory remains in PSD component
        & 196/196 projs & 252/252 projs \\
        \hline
    \end{tabular}
\end{table}

\section{Safeguarded Newton Optimization}
\label{appx:newton-optimization}

Eq.~\eqref{eq:safeguarded-newton-update} summarizes the safeguarded Newton update used by \methodname{}: we combine the analytical gradient with the approximate Hessian $\widetilde H$, initialize from a coarse set of candidate points, and apply projected Newton updates with backtracking line search. Here, we provide the complete optimization procedure, including the initialization criterion, curvature safeguards, projected-step acceptance conditions, and stopping criterion. The gradient and Hessian expressions themselves are derived in Appendix~\ref{appx:gradient-hessian}; throughout this section, $\widetilde H$ denotes the approximate Hessian used during calibration, while the full density-aware Hessian $H$ is used for the landscape analysis in Appendix~\ref{app:psd-region-validation}.

\subsection{Initialization and Curvature Safeguards}
\label{appx:newton-initialization}

For a single MatMul, the optimization variables are $\boldsymbol{\theta}=(\gamma,\beta,\alpha)$, constrained to the feasible box
\begin{equation*}
\mathcal{D} = [\theta_{\min},1]^3.
\end{equation*}
When one activation is shared by $M$ projections, the same procedure applies to
\begin{equation*}
\boldsymbol{\theta} = (\gamma,\beta,\alpha_1,\ldots,\alpha_M), \qquad \mathcal{D} = [\theta_{\min},1]^{M+2},
\end{equation*}
where $(\gamma,\beta)$ are shared activation clipping factors and each projection has its own weight clipping factor.

\paragraph{Hessian-Conditioned Initialization.} To avoid starting from an indefinite local curvature model, we evaluate a small coarse bank of candidate points. To keep this initialization search inexpensive, we set the two activation clipping factors equal, $\gamma=\beta=s$, and construct
\begin{equation*}
\mathcal{S} = \operatorname{linspace}(0.5,1,4), \qquad \mathcal{A} = \operatorname{linspace}(0.5,1,4).
\end{equation*}
For a single MatMul, the initialization bank is
\begin{equation*}
\mathcal{G} = \left\{ (s,s,\alpha): s\in\mathcal{S},\; \alpha\in\mathcal{A} \right\},
\end{equation*}
containing only $4\times4=16$ candidates. For jointly optimized branches, each candidate is extended as $(s,s,\alpha,\ldots,\alpha)$, using the same initial weight clipping factor for all branches.

For each candidate, we evaluate the approximate Hessian $\widetilde H(\boldsymbol{\theta})$. Among candidates satisfying
\begin{equation*}
\lambda_{\min} \left( \widetilde H(\boldsymbol{\theta}) \right)>0,
\end{equation*}
we initialize from the point with the smallest surrogate loss,
\begin{equation*}
\boldsymbol{\theta}_0 = \underset{ \boldsymbol{\theta}\in\mathcal{G}: \lambda_{\min}(\widetilde H(\boldsymbol{\theta}))>0 }{\arg\min} \; \mathcal{L}(\boldsymbol{\theta}).
\end{equation*}
This provides a low-cost initialization at a low-loss point with a positive-definite local curvature model.

\paragraph{Newton Direction and Curvature Safeguard.} At iteration $t$, let
\begin{equation*}
g_t = \nabla\mathcal{L}(\boldsymbol{\theta}_t), \qquad \widetilde H_t = \widetilde H(\boldsymbol{\theta}_t).
\end{equation*}
When $\widetilde H_t\succ0$, the approximate Newton direction is
\begin{equation*}
d_t = -\widetilde H_t^{-1}g_t.
\end{equation*}
For $g_t\neq0$, this is a strict descent direction \citep{nocedal2006numerical}, since
\begin{equation*}
g_t^\top d_t = -g_t^\top\widetilde H_t^{-1}g_t < 0.
\end{equation*}
We therefore enforce positive definiteness of the approximate Hessian along the accepted optimization trajectory. Specifically, a trial point is eligible for acceptance only if
\begin{equation*}
\lambda_{\min} \left( \widetilde H(\boldsymbol{\theta}_{\mathrm{trial}}) \right) > 0.
\end{equation*}
This safeguard maintains a positive-definite local curvature model at successive accepted iterates.

Positive definiteness of $\widetilde H_t$ guarantees that the unprojected Newton direction is locally descending, but a full Newton step need not decrease the nonlinear objective, particularly after enforcing the box constraints. We therefore combine this curvature safeguard with projected backtracking and an explicit sufficient-decrease condition, as described next.

\subsection{Projected Updates, Backtracking, and Stopping}
\label{appx:newton-projected-updates}

\paragraph{Projected Backtracking.} We enforce the box constraints using a projected Newton update \citep{bertsekas1982projected}. Since $\mathcal{D}$ is a box, projection onto $\mathcal{D}$ reduces to componentwise clipping:
\begin{equation*}
\boldsymbol{\theta}_{\mathrm{trial}} = \operatorname{clip} \left( \boldsymbol{\theta}_t+\eta_t d_t, \mathcal{D} \right),
\end{equation*}
where $\eta_t\in(0,1]$ is the line-search step size and $\operatorname{clip}(\cdot,\mathcal{D})$ clamps each clipping factor to its feasible interval. Because this operation can alter both the magnitude and direction of the nominal Newton update, we define the actual trial step as
\begin{equation*}
s_t = \boldsymbol{\theta}_{\mathrm{trial}} - \boldsymbol{\theta}_t.
\end{equation*}
We require this actual step to remain a descent direction,
\begin{equation*}
g_t^\top s_t < 0.
\end{equation*}
The trial point must further satisfy the Armijo sufficient-decrease condition \citep{armijo1966minimization,nocedal2006numerical},
\begin{equation*}
\mathcal{L}(\boldsymbol{\theta}_{\mathrm{trial}}) \leq \mathcal{L}(\boldsymbol{\theta}_t) + c\,g_t^\top s_t,
\end{equation*}
where $c\in(0,1)$ controls the required decrease. Finally, we retain the positive-curvature safeguard from the previous subsection,
\begin{equation*}
\lambda_{\min} \left( \widetilde H(\boldsymbol{\theta}_{\mathrm{trial}}) \right) > 0.
\end{equation*}
A trial point is accepted only when all three conditions hold:
\begin{equation}
g_t^\top s_t<0, \qquad \mathcal{L}(\boldsymbol{\theta}_{\mathrm{trial}}) \leq \mathcal{L}(\boldsymbol{\theta}_t) + c\,g_t^\top s_t, \qquad \lambda_{\min} \left( \widetilde H(\boldsymbol{\theta}_{\mathrm{trial}}) \right)>0.
\label{eq:newton-acceptance}
\end{equation}
The first two conditions in Eq.~\eqref{eq:newton-acceptance} ensure that the accepted step decreases the surrogate objective, while the third preserves a positive-definite local curvature model.

We initialize the line search with $\eta_t=1$. If any acceptance condition fails, we shrink the step geometrically,
\begin{equation*}
\eta_t \leftarrow \rho\eta_t, \qquad 0<\rho<1,
\end{equation*}
and recompute the clipped trial point. Backtracking continues until a trial point is accepted or $\eta_t<\eta_{\min}$. In the latter case, no admissible step is found and optimization terminates at the current iterate. We use $c=10^{-4}$, $\rho=0.5$, and $\eta_{\min}=10^{-4}$ in all experiments.

\paragraph{Stopping Criterion.} Because both box clipping and backtracking can substantially modify the raw Newton direction, convergence is assessed using the accepted iterates rather than the undamped Newton step. After accepting $\boldsymbol{\theta}_{t+1}$, we compute the relative loss change
\begin{equation*}
r_t = \frac{ \left| \mathcal{L}(\boldsymbol{\theta}_{t+1}) - \mathcal{L}(\boldsymbol{\theta}_t) \right| }{ \max\!\left( |\mathcal{L}(\boldsymbol{\theta}_t)|, \varepsilon_{\mathrm{num}} \right) },
\end{equation*}
where $\varepsilon_{\mathrm{num}}>0$ prevents numerical instability when the loss is close to zero. We also measure the accepted parameter displacement,
\begin{equation*}
\delta_t = \left\| \boldsymbol{\theta}_{t+1} - \boldsymbol{\theta}_t \right\|_{\infty}.
\end{equation*}
After an initial warmup of $T_{\mathrm{warmup}}$ iterations, we terminate when
\begin{equation*}
r_t<\tau_{\mathcal L} \qquad\text{and}\qquad \delta_t<\tau_{\theta}
\end{equation*}
hold for $P$ consecutive accepted iterations. In all experiments, we use
\begin{equation*}
T_{\mathrm{warmup}}=5, \qquad \tau_{\mathcal L}=10^{-7}, \qquad \tau_{\theta}=5\times10^{-5}, \qquad P=3, \qquad \varepsilon_{\mathrm{num}}=10^{-30}.
\end{equation*}
Requiring both the objective decrease and the accepted parameter displacement to remain small avoids declaring convergence when a large raw Newton step is repeatedly reduced by box clipping or backtracking.

\section{Evaluation Protocol}
\label{appx:evaluation-protocol}

This appendix provides the detailed evaluation protocol used throughout our experiments. We first describe the LLM evaluation and calibration setup, followed by the error-source analysis setup, the quantization and IMC configuration, and the clipping baselines.

\subsection{LLM Evaluation Setup}
\label{appx:llm-evaluation-setup}

We evaluate \methodname{} on LLaMA-3.2-3B and LLaMA-3.1-8B \citep{LLaMA3}, and Qwen3-4B and Qwen3-8B \citep{yang2025qwen3}. We report perplexity on WikiText-2 \citep{WikiText-2} and zero-shot accuracy on WinoGrande \citep{winogrande}, OpenBookQA \citep{openbookQA}, PIQA \citep{PIQA}, ARC-Challenge and ARC-Easy \citep{ARC}, BoolQ \citep{BoolQ}, and HellaSwag \citep{HellaSwag}. Perplexity is evaluated over the full WikiText-2 test split using non-overlapping sequences of length 2048, while all downstream tasks are evaluated in the zero-shot setting.

For clipping calibration, we sample 8 random contiguous windows of length 2048 from the WikiText-2 training split, following the small-sample calibration setting used in PrefixQuant \citep{chen2024prefixquant}. We use the same calibration samples for all clipping methods, including \methodname{} and the grid-search baselines.

All quantized comparison methods use the same Hadamard rotation, following QuaRot \citep{Quarot}, before clipping calibration to suppress activation and weight outliers. We apply Hadamard rotations at the R1--R4 locations defined in SpinQuant \citep{Spinquant}, while holding the rotation configuration fixed across all clipping methods.

\subsection{Error-Source Analysis Setup}
\label{appx:sqnr-experiment-setup}

Figure~\ref{fig:sqnr-analysis} analyzes how different clipping methods trade off activation, weight, and ADC quantization errors using real LLaMA-3.2-3B activations. We use eight non-overlapping WikiText-2 validation sequences of length 2048 and evaluate every attention and MLP projection across all 28 decoder layers. All clipping methods are evaluated using the same full-precision inputs and rotated weights.

\paragraph{Output-Space Error Measurement.} To compare the different error sources on a common scale, we measure all of them after projection into the MatMul output space rather than comparing operand-space errors directly. For an input matrix $\mathbf{X}$ and weight matrix $\mathbf{W}$, let
\begin{equation*}
\mathbf{Y} = \mathbf{X}\mathbf{W}^{\top}
\end{equation*}
denote the full-precision MatMul output, and let $\mathbf{X}_q$ and $\mathbf{W}_q$ denote the corresponding dequantized activations and weights after quantization. We define
\begin{equation*}
\mathbf{Y}_{A}
=
\mathbf{X}_q\mathbf{W}^{\top},
\qquad
\mathbf{Y}_{W}
=
\mathbf{X}\mathbf{W}_q^{\top},
\end{equation*}
which isolate the output errors caused by activation and weight quantization, respectively. We further define
\begin{equation*}
\mathbf{Y}_{Q}
=
\mathbf{X}_q\mathbf{W}_q^{\top},
\qquad
\mathbf{Y}_{I}
=
\operatorname{IMC}(\mathbf{X}_q,\mathbf{W}_q),
\end{equation*}
where $\mathbf{Y}_{Q}$ is the digital MatMul output using both quantized operands and $\mathbf{Y}_{I}$ is the corresponding output from the IMC emulation.

For $N$ output elements, the four output-space MSEs are
\begin{equation*}
\begin{aligned}
E_{\mathrm{act}}
&=
\frac{1}{N}
\left\|
\mathbf{Y}_{A}-\mathbf{Y}
\right\|_F^2,
&
E_{\mathrm{weight}}
&=
\frac{1}{N}
\left\|
\mathbf{Y}_{W}-\mathbf{Y}
\right\|_F^2,
\\
E_{\mathrm{ADC}}
&=
\frac{1}{N}
\left\|
\mathbf{Y}_{I}-\mathbf{Y}_{Q}
\right\|_F^2,
&
E_{\mathrm{total}}
&=
\frac{1}{N}
\left\|
\mathbf{Y}_{I}-\mathbf{Y}
\right\|_F^2.
\end{aligned}
\end{equation*}
Thus, $E_{\mathrm{act}}$ and $E_{\mathrm{weight}}$ isolate the effect of quantizing one operand at a time, while $E_{\mathrm{ADC}}$ isolates the additional error introduced by ADC quantization after both operands have already been quantized. $E_{\mathrm{total}}$ is measured directly from the complete IMC output and therefore also includes interactions and cross terms between error sources; consequently, the first three terms do not form an additive decomposition of the total error.

All four errors are measured in the same MatMul output space and can therefore be compared directly. Activation and weight quantization errors are propagated through the corresponding MatMul, while ADC error is already output-referred, avoiding comparisons between errors defined in different operand spaces.

\paragraph{Normalized MSE.} The scale of the MatMul output MSE varies across layers and projection types, so we normalize each error by the corresponding full-precision output signal power,
\begin{equation*}
P_Y
=
\frac{1}{N}
\left\|
\mathbf{Y}
\right\|_F^2,
\qquad
\operatorname{NMSE}_c
=
\frac{E_c}{P_Y},
\end{equation*}
where $c\in\{\mathrm{act},\mathrm{weight},\mathrm{ADC},\mathrm{total}\}$. Normalized MSE is exactly the reciprocal of the widely used linear SQNR (signal-to-quantization-noise ratio),
\begin{equation*}
\operatorname{SQNR}_c
=
\frac{P_Y}{E_c}
=
\frac{1}{\operatorname{NMSE}_c},
\end{equation*}
or equivalently,
\begin{equation*}
\operatorname{SQNR}_{c,\mathrm{dB}}
=
-10\log_{10}\operatorname{NMSE}_c.
\end{equation*}
We report normalized MSE because it removes differences in output scale while preserving a one-to-one correspondence with SQNR, and directly exposes the relative magnitudes and trade-offs among the different error sources.

For each layer and projection, the squared errors and signal power are accumulated over all eight validation sequences and output channels before normalization. For each projection type, Fig.~\ref{fig:sqnr-analysis} reports the median normalized MSE across the 28 decoder layers, with the shaded region indicating the 25th--75th percentiles.

\subsection{Quantization and IMC Configuration}
\label{appx:quantization-imc-configuration}

Unless otherwise specified, all methods use dynamic asymmetric per-token A8 quantization and static symmetric per-channel W8 quantization. We additionally quantize the KV cache to 8 bits using dynamic asymmetric quantization with group size 128.

Our analog IMC emulation follows Config.~2 of \citet{lee2024switched}, including its operand encoding and accumulation scheme. We choose this switched-capacitor SRAM implementation because it achieves state-of-the-art energy efficiency in 28-nm CMOS while maintaining sufficiently high analog compute precision for an algorithm-level abstraction. Each 8-bit activation and weight operand is decomposed into two 4-bit slices, producing four 4-bit $\times$ 4-bit partial products. MatMuls are mapped to IMC arrays with 512 rows and 32 output columns, and each analog partial sum is quantized by a 9-bit ADC before the digitized partial products are recombined.

The ADC full-scale range is set by the maximum analog accumulation range determined by the circuit's operand encoding and 512-row accumulation depth, and is held fixed across all clipping methods. MatMuls with larger reduction dimensions are partitioned across multiple 512-row arrays, whose digitized outputs are accumulated digitally. ADC precision and IMC reduction dimension are varied separately in Appendix~\ref{appx:imc-array-dimensions}.

\subsection{Grid-Search Baseline Implementation}
\label{appx:baselines}

We implement the W/A grid-search baseline following the MSE-based clipping search procedure and hyperparameters of PrefixQuant \citep{chen2024prefixquant}, with one deliberate modification for our target hardware setting: quantized block forwards are executed through our analog IMC emulation rather than ordinary digital quantized inference. Consequently, the reconstruction loss used by the grid search is indirectly affected by ADC quantization, making the baseline partially ADC-aware. However, unlike \methodname{}, it does not explicitly model or directly optimize the coupled operand- and ADC-quantization errors.

For dynamic asymmetric activation quantization, the upper and lower clipping factors are independently searched over \{0.60,0.65,\ldots,1.00\}, following PrefixQuant \citep{chen2024prefixquant}. This gives a $9\times9$ candidate grid. Each candidate pair is evaluated using the MSE between the quantized decoder-block output and its corresponding full-precision output, and the pair with the lowest reconstruction error is selected. In our implementation, these quantized block forwards use the IMC emulation described in Appendix~\ref{appx:quantization-imc-configuration}.

For weight clipping, we retain the activation-aware, group-wise weight-scale search used in PrefixQuant \citep{chen2024prefixquant}, including $n_{\mathrm{grid}}=20$ and $\mathrm{max\_shrink}=0.5$. The search performs ten iterations with nominal shrink factors $1-i/20$, $i=0,\ldots,9$, and selects the weight scales that minimize the MSE between the original and quantized linear-layer outputs on the cached calibration activations. This local weight search is otherwise unchanged from the digital baseline.

Calibration proceeds block by block. For each block, we first obtain its full-precision output and cache the required intermediate activations. The clipping parameters are then searched sequentially across the quantizers in the block, after which the fully quantized block output is propagated as the calibration input to the next block. All clipping methods use the same calibration samples, rotation configuration, quantization precision, and IMC configuration.

\subsection{ADC-Aware Alternating Search Benchmark Implementation}
\label{appx:adc-aware-alternating}

While the W/A grid-search baseline directly follows prior clipping work \citep{chen2024prefixquant}, its optimization objective differs from the MatMul output-error objective optimized by \methodname{}, and therefore does not directly indicate how accurately the analytical surrogate represents that objective. We therefore introduce an ADC-aware alternating-search benchmark that replaces the surrogate with direct empirical IMC output MSE while retaining the same clipping variables.

Let $Y=XW^\top$ denote the floating-point MatMul output and let
\begin{equation*}
Y_I(\boldsymbol{\theta})
=
\operatorname{IMC}\!\left(
X_q(\gamma,\beta), W_q(\alpha)
\right)
\end{equation*}
denote the corresponding output from the production quantizers and IMC forward path. The benchmark directly minimizes
\begin{equation*}
\mathcal{L}_{\mathrm{emp}}(\boldsymbol{\theta})
=
\operatorname{MSE}\!\left(
Y_I(\boldsymbol{\theta}),Y
\right).
\end{equation*}
When one activation is shared across multiple projections, such as Q/K/V or Gate/Up, we use the same parameterization as \methodname{},
\begin{equation*}
\boldsymbol{\theta}
=
(\gamma,\beta,\alpha_1,\ldots,\alpha_M),
\end{equation*}
and evaluate a shared activation candidate using the summed empirical output MSE across the corresponding projections.

Starting from no clipping, we alternately search the activation clipping factors $(\gamma,\beta)$ and weight clipping factors $\alpha_m$ over
\begin{equation*}
\{0.001,0.10,0.20,0.30,0.40,0.50,0.60,0.70,0.80,0.90,1.00\}.
\end{equation*}
Every candidate is evaluated through the complete ADC-enabled IMC forward path, making both activation and weight clipping directly aware of ADC quantization. The finite grid trades search resolution for calibration cost: a finer grid provides a more precise empirical search but requires proportionally more IMC evaluations. We repeat the alternating search for a fixed number of iterations, using 5 iterations for the benchmark reported in Fig.~\ref{fig:surrogate-loss-fidelity}.

\subsection{Surrogate Fidelity Evaluation}
\label{appx:surrogate-fidelity}

We further verify that the analytical surrogate closely approximates the empirical MatMul output MSE. Using the same WikiText-2 activations and output-space measurements as Appendix~\ref{appx:sqnr-experiment-setup}, we evaluate every projection across all 28 layers of LLaMA-3.2-3B under the clipping factors produced by \methodname{}. For each error component, we compare its analytical surrogate $\mathcal{L}$ with the corresponding empirically measured MSE $M$, and report the relative mismatch
\begin{equation*}
100\frac{|\mathcal{L}-M|}{M}.
\end{equation*}

\begin{figure}[H]
    \centering
    \includegraphics[width=\textwidth]{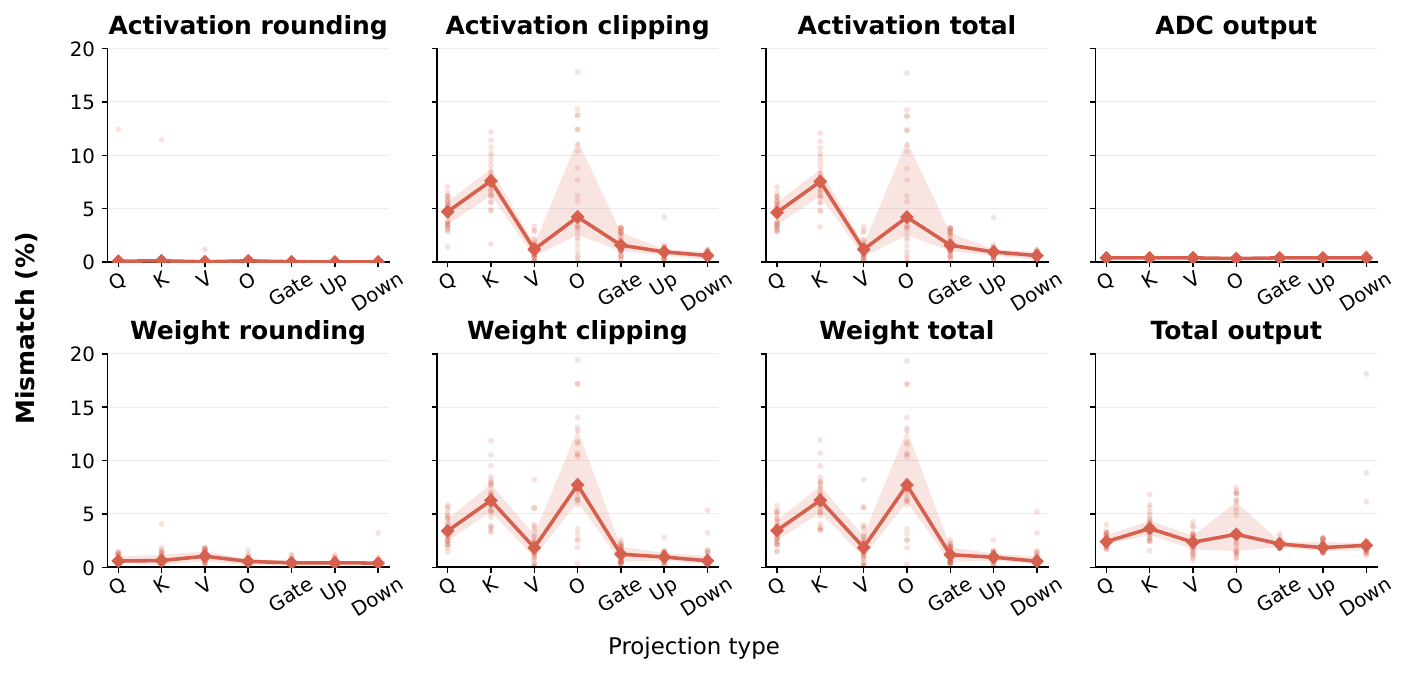}
    \caption{Relative mismatch between the analytical surrogate and empirically measured MatMul output MSE for LLaMA-3.2-3B at the calibrated solutions. Points denote individual layers, solid lines show the median across all 28 layers, and shaded regions indicate the 25th--75th percentiles. Each panel reports the indicated error component across projection types.}
    \label{fig:surrogate-mse-mismatch}
\end{figure}

Figure~\ref{fig:surrogate-mse-mismatch} reports this mismatch for activation rounding, clipping, and total error; weight rounding, clipping, and total error; ADC error; and the complete IMC output error. The component-wise diagnostics show that most approximation error arises from clipping, while the rounding and ADC terms are modeled particularly closely. Most importantly, the complete surrogate achieves only approximately $2$--$4\%$ median mismatch across all projection types. This confirms that the surrogate closely tracks empirical IMC output MSE while enabling analytical first- and second-order information for efficient optimization.

\subsection{Projected Gradient Descent Implementation}
\label{appx:optimizer-comparison}

To isolate the optimization benefit of safeguarded Newton updates, we compare \methodname{} against projected GD while holding the surrogate objective, initialization, feasible domain, and calibration data fixed. We evaluate 112 optimization problems from all projection layers in LLaMA-3.2-3B (shared-input projections are treated as one optimization problem). For each problem, both optimizers start from the same coarse initialization and optimize the same parameter vector
\begin{equation*}
\boldsymbol{\theta}
=
(\gamma,\beta,\alpha_1,\ldots,\alpha_M),
\qquad
10^{-3}\leq\theta_j\leq1.
\end{equation*}
Projected GD uses the analytical surrogate gradient with box projection and Armijo backtracking, while \methodname{} uses the production approximate Hessian and safeguarded Newton updates. The shared initialization cost is excluded from optimizer time.

We compare the optimizers using the time required to recover $95\%$ of the available post-initialization loss improvement. For each problem, let
\begin{equation*}
L_0=\mathcal{L}(\boldsymbol{\theta}_0)
\end{equation*}
be the common post-initialization loss and $L_\star$ the lowest feasible loss observed across the compared runs. We define
\begin{equation*}
L_{95}
=
L_\star+0.05(L_0-L_\star),
\end{equation*}
and report the time at which each optimizer first evaluates a feasible point satisfying
\begin{equation*}
\mathcal{L}(\boldsymbol{\theta})\leq L_{95}.
\end{equation*}
The timing includes objective evaluations performed during line search and therefore measures the time to find a solution meeting the target, rather than only accepted iterations. Both methods use the same 30-second optimizer budget.

As shown in Fig.~\ref{fig:surrogate-loss-fidelity} (right), \methodname{} reaches the target on all $112/112$ problems with a median time of $72\,\mathrm{ms}$, whereas projected GD reaches it on $87/112$ problems with a median of $1.10\,\mathrm{s}$ among successful runs. This comparison shows that the second-order information exploited by \methodname{} substantially accelerates optimization of the same surrogate objective.

\section{Extended Results and Ablation Studies}
\label{appx:extended-results}

This appendix provides additional sweep and ablation results that complement the main experiments. We evaluate the sensitivity of \methodname{} to hardware and calibration settings and further examine its robustness and optimization behavior.

\subsection{Error-Source Analysis Across Models}
\label{appx:error-source-across-models}

\begin{figure*}[p]
    \centering
    \begin{subfigure}{\textwidth}
        \centering
        \includegraphics[width=\linewidth]{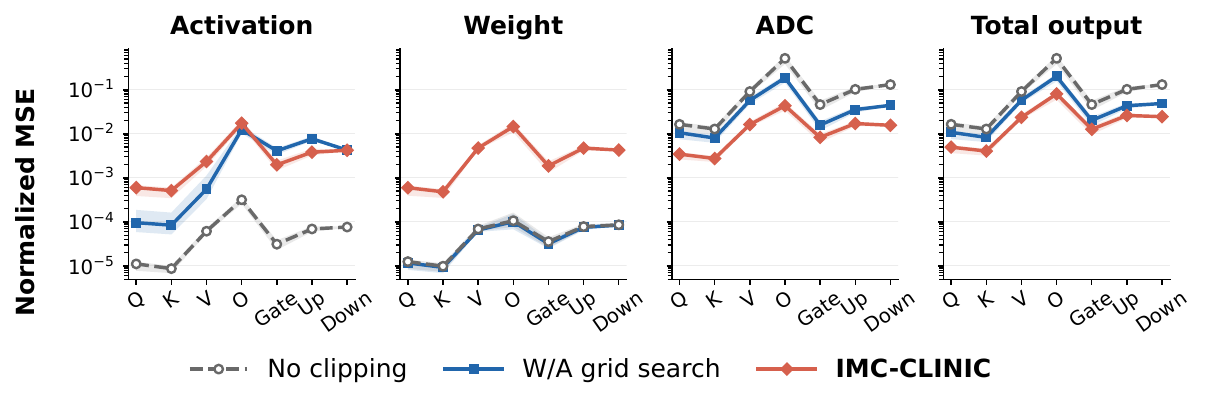}
        \caption{LLaMA-3.1-8B.}
        \label{fig:error-sources-llama31-8b}
    \end{subfigure}

    \medskip
    \begin{subfigure}{\textwidth}
        \centering
        \includegraphics[width=\linewidth]{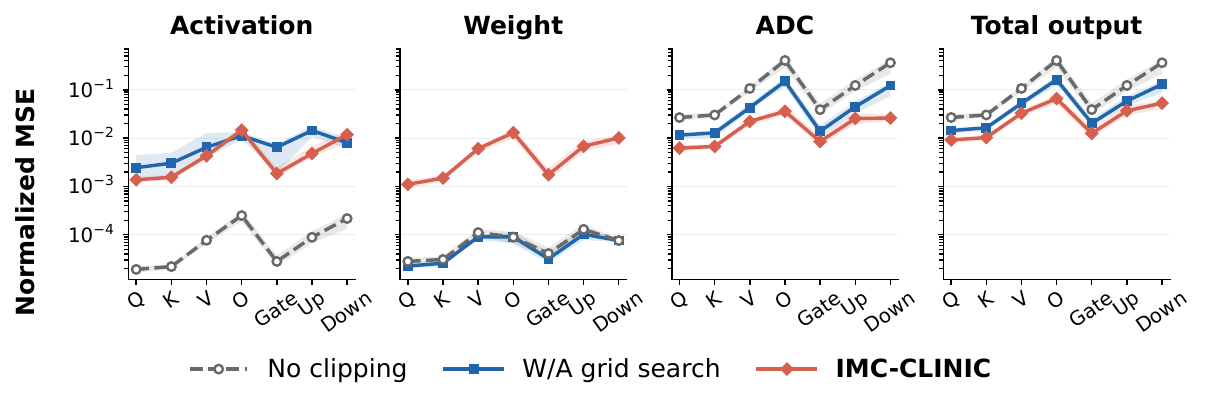}
        \caption{Qwen3-4B.}
        \label{fig:error-sources-qwen3-4b}
    \end{subfigure}

    \medskip
    \begin{subfigure}{\textwidth}
        \centering
        \includegraphics[width=\linewidth]{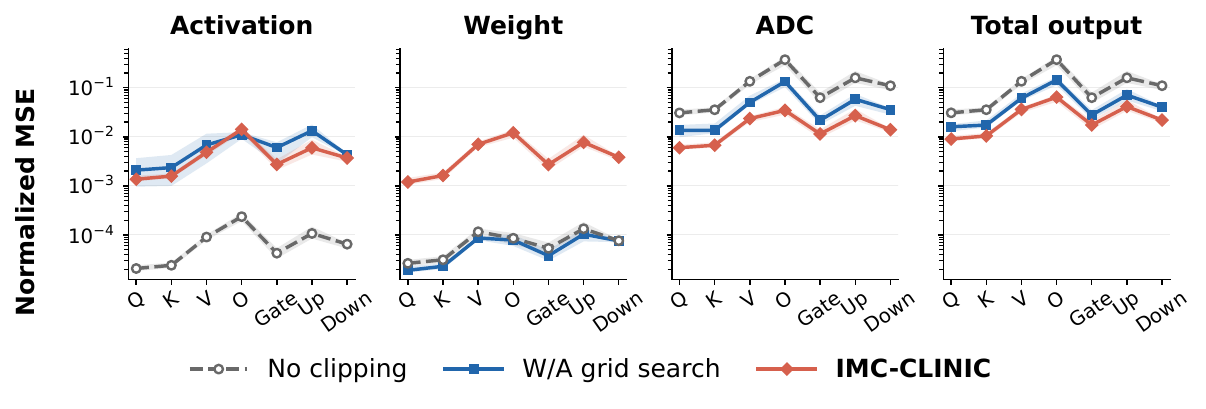}
        \caption{Qwen3-8B.}
        \label{fig:error-sources-qwen3-8b}
    \end{subfigure}
    \caption{Output-error analysis for the three additional models. Each subfigure shows normalized MSE from activation quantization, weight quantization, and ADC quantization, followed by the total IMC output error across projection types. Points report the median across decoder layers, and shaded bands show the interquartile range. Lower is better.}
    \label{fig:error-source-across-models}
\end{figure*}

Figure~\ref{fig:error-source-across-models} extends the error-source analysis in Sec.~\ref{sec:sqnr-analysis-across-error-sources} to LLaMA-3.1-8B, Qwen3-4B, and Qwen3-8B. Across these models and projection types, \methodname{} accepts somewhat larger operand quantization error to reduce ADC error, resulting in lower total output error than the clipping baselines.

\subsection{Optimized Clipping Factors}
\label{appx:clipping-factors}

\begin{figure*}[t]
    \centering
    \includegraphics[width=\textwidth]
    {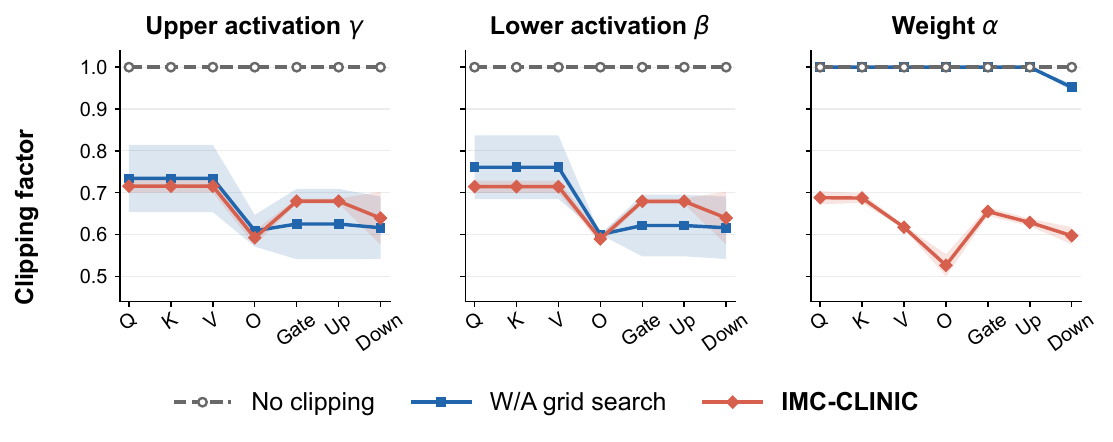}
    \caption{Clipping factors for LLaMA-3.2-3B under the default 9-bit ADC configuration. Results are shown for the upper activation factor $\gamma$, lower activation factor $\beta$, and weight factor $\alpha$ across projection types. Markers denote the mean across decoder layers, and shaded regions indicate $\pm 1$ standard deviation. A clipping factor of $1$ corresponds to no clipping.}
    \label{fig:clipping-factors}
\end{figure*}

Figure~\ref{fig:clipping-factors} shows the clipping factors found by \methodname{} for LLaMA-3.2-3B under the default 9-bit ADC configuration. Across projection types, the optimized activation and weight clipping factors are typically in the range of approximately $0.5$--$0.7$. Thus, balancing operand quantization error against ADC quantization error requires relatively aggressive clipping, reducing the retained operand range by roughly $30$--$50\%$ compared with no clipping.

\subsection{Full Task-Level Accuracy Results}
\label{appx:full-task-results}

Table~\ref{tab:full-task-results} expands the average zero-shot accuracy reported in Table~\ref{tab:accuracy-and-calibration-time} into the seven constituent tasks. We report length-normalized accuracy (\texttt{acc\_norm}) for OpenBookQA, PIQA, ARC-Challenge, ARC-Easy, and HellaSwag, and standard accuracy (\texttt{acc}) for WinoGrande and BoolQ. Avg. is the unweighted mean across all seven tasks, computed before rounding. W8A8 no-IMC denotes the INT8 digital implementation, which helps isolate the degradation caused by ADC quantization in analog IMC. Because its results remain very close to FP16 across all four models and configurations we tested, we omit this reference from the remaining experiments for clarity.

\begin{table}[h!]
    \centering
    \caption{Full WikiText-2 perplexity (PPL), seven-task zero-shot accuracy, and calibration time across models and inference configurations. Avg. is the unweighted mean task accuracy; a dash indicates that calibration time is unavailable or inapplicable.}
    \label{tab:full-task-results}
    \scriptsize
    \setlength{\tabcolsep}{3.5pt}
    \begin{tabular}{llr:rrrrrrrr:r}
        \hline
        Model & Method
        & PPL ($\downarrow$)
        & Wino & OBQA & PIQA & ARC-C & BoolQ & ARC-E & Hella
        & Avg. ($\uparrow$)
        & Calib. Time ($\downarrow$) \\
        \hline

        LLaMA-3.2-3B
            & No clipping
            & 175.70
            & 0.515 & 0.268 & 0.533 & 0.241 & 0.527 & 0.293 & 0.310
            & 0.384
            & -- \\
            & W/A grid search
            & 19.12
            & 0.566 & 0.294 & 0.672 & 0.317 & 0.515 & 0.528 & 0.525
            & 0.488
            & 32.9 min \\
            & \textbf{\methodname{}}
            & \textbf{12.27}
            & \textbf{0.622} & \textbf{0.368} & \textbf{0.722} & \textbf{0.367}
            & \textbf{0.665} & \textbf{0.630} & \textbf{0.651}
            & \textbf{0.575}
            & \textbf{3.3 min} \\
        \hdashline
            & W8A8 no-IMC
            & 7.82
            & 0.698 & 0.402 & 0.778 & 0.464 & 0.745 & 0.723 & 0.740
            & 0.650
            & -- \\
            & FP16 baseline
            & 7.81
            & 0.694 & 0.408 & 0.781 & 0.462 & 0.742 & 0.721 & 0.741
            & 0.650
            & -- \\
        \hline

        LLaMA-3.1-8B
            & No clipping
            & 127.28
            & 0.460 & 0.274 & 0.572 & 0.263 & 0.489 & 0.363 & 0.357
            & 0.397
            & -- \\
            & W/A grid search
            & 13.68
            & 0.591 & 0.380 & 0.718 & 0.378 & 0.642 & 0.615 & 0.649
            & 0.567
            & 73.7 min \\
            & \textbf{\methodname{}}
            & \textbf{9.17}
            & \textbf{0.680} & \textbf{0.386} & \textbf{0.763} & \textbf{0.440}
            & \textbf{0.778} & \textbf{0.734} & \textbf{0.744}
            & \textbf{0.646}
            & \textbf{6.1 min} \\
        \hdashline
            & W8A8 no-IMC
            & 6.25
            & 0.747 & 0.456 & 0.809 & 0.543 & 0.828 & 0.826 & 0.793
            & 0.714
            & -- \\
            & FP16 baseline
            & 6.24
            & 0.746 & 0.454 & 0.812 & 0.549 & 0.831 & 0.826 & 0.793
            & 0.716
            & -- \\
        \hline

        Qwen3-4B
            & No clipping
            & 2156.59
            & 0.485 & 0.272 & 0.514 & 0.247 & 0.446 & 0.271 & 0.279
            & 0.359
            & -- \\
            & W/A grid search
            & 71.68
            & 0.498 & 0.276 & 0.557 & 0.275 & 0.580 & 0.364 & 0.382
            & 0.419
            & 44.4 min \\
            & \textbf{\methodname{}}
            & \textbf{30.17}
            & \textbf{0.548} & \textbf{0.326} & \textbf{0.688} & \textbf{0.362}
            & \textbf{0.735} & \textbf{0.545} & \textbf{0.536}
            & \textbf{0.534}
            & \textbf{4.3 min} \\
        \hdashline
            & W8A8 no-IMC
            & 13.61
            & 0.665 & 0.408 & 0.748 & 0.533 & 0.853 & 0.785 & 0.684
            & 0.668
            & -- \\
            & FP16 baseline
            & 13.64
            & 0.660 & 0.400 & 0.749 & 0.540 & 0.851 & 0.786 & 0.684
            & 0.667
            & -- \\
        \hline

        Qwen3-8B
            & No clipping
            & 38.61
            & 0.537 & 0.266 & 0.626 & 0.276 & 0.530 & 0.421 & 0.438
            & 0.442
            & -- \\
            & W/A grid search
            & 13.94
            & 0.545 & 0.374 & 0.724 & 0.417 & 0.746 & 0.652 & 0.616
            & 0.582
            & 74.6 min \\
            & \textbf{\methodname{}}
            & \textbf{11.72}
            & \textbf{0.646} & \textbf{0.400} & \textbf{0.747} & \textbf{0.485}
            & \textbf{0.825} & \textbf{0.729} & \textbf{0.695}
            & \textbf{0.647}
            & \textbf{6.5 min} \\
        \hdashline
            & W8A8 no-IMC
            & 9.70
            & 0.680 & 0.418 & 0.777 & 0.563 & 0.867 & 0.807 & 0.750
            & 0.695
            & -- \\
            & FP16 baseline
            & 9.72
            & 0.676 & 0.414 & 0.777 & 0.565 & 0.866 & 0.809 & 0.750
            & 0.694
            & -- \\
        \hline
    \end{tabular}
\end{table}

\subsection{ADC Precision}
\label{appx:adc-precision}
We first evaluate the effect of ADC precision by sweeping the resolution from 8 to 10 bits, covering a typical operating range for switched-capacitor IMC \citep{lee2021fully,lee2024switched}. Higher-resolution ADCs are not considered because their conversion energy increases sharply with resolution \citep{adc_survey}, while the smaller least-significant-bit (LSB) step makes them increasingly sensitive to analog noise. As shown in Table~\ref{tab:adc-precision-sweep}, the benefit of clipping is largest at lower ADC precision, where ADC quantization is most severe. At 8 and 9 bits, \methodname{} substantially improves inference accuracy over No Clipping and W/A Grid Search. At 10 bits, the performance gap narrows as ADC quantization becomes less dominant, although \methodname{} still achieves the best perplexity and accuracy results.

\begin{table*}[t]
    \centering
    \caption{LLaMA-3.2-3B results across ADC precision.}
    \label{tab:adc-precision-sweep}
    \scriptsize
    \setlength{\tabcolsep}{3.5pt}
    \begin{tabular}{rlr:rrrrrrrr}
        \hline
        ADC bits & Method
        & PPL ($\downarrow$)
        & Wino & OBQA & PIQA & ARC-C & BoolQ & ARC-E & Hella
        & Avg. ($\uparrow$) \\
        \hline

        8
        & No clipping
        & 29717.68
        & 0.502 & \textbf{0.294} & 0.505 & \textbf{0.271} & 0.412 & 0.256 & 0.262
        & 0.358 \\
        & W/A grid search
        & 1572.10
        & 0.510 & 0.262 & 0.504 & 0.268 & 0.395 & 0.259 & 0.263
        & 0.352 \\
        & \methodname{}
        & \textbf{42.63}
        & \textbf{0.530} & 0.244 & \textbf{0.585} & 0.242
        & \textbf{0.531} & \textbf{0.402} & \textbf{0.376}
        & \textbf{0.416} \\

        \hdashline
        9$^{*}$
        & No clipping
        & 175.70
        & 0.515 & 0.268 & 0.533 & 0.241 & 0.527 & 0.293 & 0.310
        & 0.384 \\
        & W/A grid search
        & 19.12
        & 0.566 & 0.294 & 0.672 & 0.317 & 0.515 & 0.528 & 0.525
        & 0.488 \\
        & \methodname{}
        & \textbf{12.27}
        & \textbf{0.622} & \textbf{0.368} & \textbf{0.722} & \textbf{0.367}
        & \textbf{0.665} & \textbf{0.630} & \textbf{0.651}
        & \textbf{0.575} \\

        \hdashline
        10
        & No clipping
        & 10.25
        & 0.626 & 0.362 & \textbf{0.752} & 0.393 & 0.677 & 0.656 & 0.681
        & 0.592 \\
        & W/A grid search
        & 9.33
        & 0.651 & 0.398 & \textbf{0.752} & 0.431 & 0.671 & 0.683 & 0.708
        & 0.614 \\
        & \methodname{}
        & \textbf{9.04}
        & \textbf{0.665} & \textbf{0.408} & 0.748 & \textbf{0.433}
        & \textbf{0.688} & \textbf{0.692} & \textbf{0.715}
        & \textbf{0.621} \\

        \hdashline
        & FP16 baseline
        & 7.81
        & 0.694 & 0.408 & 0.781 & 0.463 & 0.742 & 0.721 & 0.741
        & 0.650 \\

        \hline
        \multicolumn{11}{l}{\scriptsize $^{*}$ denotes default configuration.}
    \end{tabular}
\end{table*}

\subsection{IMC Array Dimensions}
\label{appx:imc-array-dimensions}
In the main experiments, we use a fixed IMC array with 512 rows and 32 columns. Here, we vary both dimensions to evaluate the sensitivity of clipping calibration to the array shape. The number of rows determines the analog reduction length and therefore directly affects the accumulated partial-sum range and ADC quantization error. This dependency is explicitly captured in our surrogate objective, allowing \methodname{} to adapt its clipping factors to different reduction lengths. As shown in Table~\ref{tab:imc-row-sweep}, \methodname{} consistently outperforms the clipping baselines across row counts from 128 to 2048, with its advantage becoming particularly pronounced at larger reduction dimensions where ADC quantization is more severe.

In contrast, the number of columns only changes how many output channels are computed in parallel and does not affect the per-column analog reduction or ADC quantization. Accordingly, Table~\ref{tab:imc-column-sweep} shows nearly identical results across different column counts.

\begin{table*}[t]
    \centering
    \caption{LLaMA-3.2-3B results across IMC row counts.}
    \label{tab:imc-row-sweep}
    \scriptsize
    \setlength{\tabcolsep}{3.5pt}
    \begin{tabular}{rlr:rrrrrrrr}
        \hline
        IMC rows & Method
        & PPL ($\downarrow$)
        & Wino & OBQA & PIQA & ARC-C & BoolQ & ARC-E & Hella
        & Avg. ($\uparrow$) \\
        \hline

        128
        & No clipping
        & 10.36
        & 0.627 & 0.404 & 0.737 & 0.419 & 0.647 & 0.660 & 0.696
        & 0.599 \\
        & W/A grid search
        & 9.39
        & 0.669 & 0.400 & \textbf{0.751} & 0.406 & 0.657 & 0.677 & 0.699
        & 0.608 \\
        & \methodname{}
        & \textbf{9.04}
        & \textbf{0.680} & \textbf{0.408} & 0.749 & \textbf{0.434}
        & \textbf{0.706} & \textbf{0.692} & \textbf{0.712}
        & \textbf{0.626} \\

        \hdashline
        256
        & No clipping
        & 15.86
        & 0.564 & 0.326 & 0.681 & 0.331 & 0.518 & 0.528 & 0.593
        & 0.506 \\
        & W/A grid search
        & 11.42
        & 0.647 & 0.358 & 0.742 & 0.398 & 0.607 & 0.657 & 0.661
        & 0.581 \\
        & \methodname{}
        & \textbf{10.03}
        & \textbf{0.648} & \textbf{0.384} & \textbf{0.744} & \textbf{0.401}
        & \textbf{0.654} & \textbf{0.660} & \textbf{0.688}
        & \textbf{0.597} \\

        \hdashline
        512$^{*}$
        & No clipping
        & 175.70
        & 0.515 & 0.268 & 0.533 & 0.241 & 0.527 & 0.293 & 0.310
        & 0.384 \\
        & W/A grid search
        & 19.12
        & 0.566 & 0.294 & 0.672 & 0.317 & 0.515 & 0.528 & 0.525
        & 0.488 \\
        & \methodname{}
        & \textbf{12.27}
        & \textbf{0.622} & \textbf{0.368} & \textbf{0.722} & \textbf{0.367}
        & \textbf{0.665} & \textbf{0.630} & \textbf{0.651}
        & \textbf{0.575} \\

        \hdashline
        1024
        & No clipping
        & 4408.64
        & 0.506 & 0.272 & 0.501 & 0.247 & 0.404 & 0.269 & 0.269
        & 0.352 \\
        & W/A grid search
        & 116.04
        & 0.504 & 0.224 & 0.547 & 0.214 & 0.522 & 0.332 & 0.326
        & 0.381 \\
        & \methodname{}
        & \textbf{19.35}
        & \textbf{0.584} & \textbf{0.296} & \textbf{0.668} & \textbf{0.294}
        & \textbf{0.564} & \textbf{0.522} & \textbf{0.548}
        & \textbf{0.497} \\

        \hdashline
        2048
        & No clipping
        & 19971.77
        & 0.508 & \textbf{0.294} & 0.518 & 0.261 & 0.413 & 0.245 & 0.260
        & 0.357 \\
        & W/A grid search
        & 800.85
        & 0.517 & 0.266 & 0.497 & \textbf{0.276} & 0.422 & 0.264 & 0.268
        & 0.358 \\
        & \methodname{}
        & \textbf{37.89}
        & \textbf{0.542} & 0.282 & \textbf{0.603} & 0.243
        & \textbf{0.487} & \textbf{0.400} & \textbf{0.406}
        & \textbf{0.424} \\

        \hdashline
        & FP16 baseline
        & 7.81
        & 0.694 & 0.408 & 0.781 & 0.463 & 0.742 & 0.721 & 0.741
        & 0.650 \\

        \hline
        \multicolumn{11}{l}{\scriptsize $^{*}$ denotes default configuration.}
    \end{tabular}
\end{table*}

\begin{table*}[t]
    \centering
    \caption{LLaMA-3.2-3B results across IMC column counts.}
    \label{tab:imc-column-sweep}
    \scriptsize
    \setlength{\tabcolsep}{3.5pt}
    \begin{tabular}{rlr:rrrrrrrr}
        \hline
        IMC columns & Method
        & PPL ($\downarrow$)
        & Wino & OBQA & PIQA & ARC-C & BoolQ & ARC-E & Hella
        & Avg. ($\uparrow$) \\
        \hline

        8
        & No clipping
        & 175.32
        & 0.505 & 0.228 & 0.542 & 0.223 & 0.538 & 0.299 & 0.311
        & 0.378 \\
        & W/A grid search
        & 19.12
        & 0.565 & 0.296 & 0.672 & 0.317 & 0.515 & 0.529 & 0.525
        & 0.488 \\
        & \methodname{}
        & \textbf{12.27}
        & \textbf{0.622} & \textbf{0.368} & \textbf{0.722} & \textbf{0.367}
        & \textbf{0.665} & \textbf{0.630} & \textbf{0.651}
        & \textbf{0.575} \\

        \hdashline
        16
        & No clipping
        & 175.32
        & 0.504 & 0.228 & 0.542 & 0.223 & 0.538 & 0.299 & 0.311
        & 0.378 \\
        & W/A grid search
        & 19.12
        & 0.565 & 0.296 & 0.672 & 0.317 & 0.515 & 0.529 & 0.525
        & 0.488 \\
        & \methodname{}
        & \textbf{12.27}
        & \textbf{0.622} & \textbf{0.368} & \textbf{0.722} & \textbf{0.367}
        & \textbf{0.665} & \textbf{0.630} & \textbf{0.651}
        & \textbf{0.575} \\

        \hdashline
        32$^{*}$
        & No clipping
        & 175.70
        & 0.515 & 0.268 & 0.533 & 0.241 & 0.527 & 0.293 & 0.310
        & 0.384 \\
        & W/A grid search
        & 19.12
        & 0.566 & 0.294 & 0.672 & 0.317 & 0.515 & 0.528 & 0.525
        & 0.488 \\
        & \methodname{}
        & \textbf{12.27}
        & \textbf{0.622} & \textbf{0.368} & \textbf{0.722} & \textbf{0.367}
        & \textbf{0.665} & \textbf{0.630} & \textbf{0.651}
        & \textbf{0.575} \\

        \hdashline
        64
        & No clipping
        & 175.32
        & 0.505 & 0.228 & 0.542 & 0.223 & 0.538 & 0.299 & 0.311
        & 0.378 \\
        & W/A grid search
        & 19.12
        & 0.565 & 0.296 & 0.672 & 0.317 & 0.515 & 0.529 & 0.525
        & 0.488 \\
        & \methodname{}
        & \textbf{12.27}
        & \textbf{0.622} & \textbf{0.368} & \textbf{0.722} & \textbf{0.367}
        & \textbf{0.665} & \textbf{0.630} & \textbf{0.651}
        & \textbf{0.575} \\

        \hdashline
        128
        & No clipping
        & 175.32
        & 0.505 & 0.228 & 0.542 & 0.223 & 0.538 & 0.299 & 0.311
        & 0.378 \\
        & W/A grid search
        & 19.12
        & 0.565 & 0.296 & 0.672 & 0.317 & 0.515 & 0.529 & 0.525
        & 0.488 \\
        & \methodname{}
        & \textbf{12.27}
        & \textbf{0.622} & \textbf{0.368} & \textbf{0.722} & \textbf{0.367}
        & \textbf{0.665} & \textbf{0.630} & \textbf{0.651}
        & \textbf{0.575} \\

        \hdashline
        & FP16 baseline
        & 7.81
        & 0.694 & 0.408 & 0.781 & 0.463 & 0.742 & 0.721 & 0.741
        & 0.650 \\

        \hline
        \multicolumn{11}{l}{\scriptsize $^{*}$ denotes default configuration.}
    \end{tabular}
\end{table*}

\subsection{Robustness to ADC Analog Noise}

\label{sec:appendix_adc_noise}

Although charge-domain switched-capacitor IMC can provide high-precision analog partial-sum computation even over large accumulation dimensions \citep{valavi2019charge, lee2021fully}, the subsequent ADC remains subject to random analog noise in addition to quantization error. Important contributors include sampling $kT/C$ noise, comparator noise, reference-path noise, and related circuit effects \citep{shen201816, zhong2015thermal}. To evaluate the robustness of \methodname{} to these nonidealities, we inject additive zero-mean Gaussian noise independently into each of the four slice-level ADC conversions \citep{ASIM}. For $j\in\{HH,HL,LH,LL\}$,

\begin{equation*}
    \tilde{p}_j = p_j + n_j, \qquad
    n_j \sim \mathcal{N}\!\left(
        0,
        \left(\sigma_{\mathrm{A}}\Delta_{\mathrm{slice}}\right)^2
    \right),
\end{equation*}

where $p_j$ is the ideal analog partial sum for slice $j$, $\Delta_{\mathrm{slice}}$ is one physical ADC LSB, and $\sigma_{\mathrm{A}}$ specifies the root-mean-square (RMS) analog-noise magnitude in LSB units. Noise is injected before ADC quantization,

\begin{equation*}
    p_{j,q} = Q_{\mathrm{ADC}}(\tilde{p}_j),
\end{equation*}

after which the four digitized slice-level partial sums are recombined using their corresponding significance weights as described in Appendix~\ref{appx:matmul-output-error-surrogate}.

As a robustness study, we evaluate $\sigma_{\mathrm{A}}\in\{0.1,0.2,0.3,0.4\}\,\mathrm{LSB}_{\mathrm{RMS}}$. Table~\ref{tab:analog-noise-sweep} reports the resulting inference performance. As the analog-noise magnitude increases, the additional ADC-input uncertainty reduces the achievable inference accuracy for all clipping methods. Nevertheless, \methodname{} consistently outperforms the baseline clipping methods across the evaluated noise levels, demonstrating that its benefit is preserved in the presence of random ADC analog noise.

\begin{table*}[t]
    \centering
    \caption{LLaMA-3.2-3B robustness to analog noise.}
    \label{tab:analog-noise-sweep}
    \scriptsize
    \setlength{\tabcolsep}{3.5pt}
    \begin{tabular}{rlr:rrrrrrrr}
        \hline
        Noise (LSB) & Method
        & PPL ($\downarrow$)
        & Wino & OBQA & PIQA & ARC-C & BoolQ & ARC-E & Hella
        & Avg. ($\uparrow$) \\
        \hline

        0$^{*}$
        & No clipping
        & 175.70
        & 0.515 & 0.268 & 0.533 & 0.241 & 0.527 & 0.293 & 0.310
        & 0.384 \\
        & W/A grid search
        & 19.12
        & 0.566 & 0.294 & 0.672 & 0.317 & 0.515 & 0.528 & 0.525
        & 0.488 \\
        & \methodname{}
        & \textbf{12.27}
        & \textbf{0.622} & \textbf{0.368} & \textbf{0.722} & \textbf{0.367}
        & \textbf{0.665} & \textbf{0.630} & \textbf{0.651}
        & \textbf{0.575} \\

        \hdashline
        0.1
        & No clipping
        & 288.45
        & 0.514 & 0.256 & 0.503 & 0.242 & 0.478 & 0.294 & 0.287
        & 0.368 \\
        & W/A grid search
        & 21.70
        & 0.552 & 0.300 & 0.599 & 0.299 & 0.536 & 0.485 & 0.490
        & 0.466 \\
        & \methodname{}
        & \textbf{12.58}
        & \textbf{0.614} & \textbf{0.352} & \textbf{0.679} & \textbf{0.345}
        & \textbf{0.643} & \textbf{0.601} & \textbf{0.636}
        & \textbf{0.553} \\

        \hdashline
        0.2
        & No clipping
        & 1299.73
        & 0.516 & 0.270 & 0.511 & 0.239 & 0.424 & 0.264 & 0.265
        & 0.356 \\
        & W/A grid search
        & 42.63
        & 0.516 & 0.250 & 0.552 & 0.243 & 0.480 & 0.384 & 0.385
        & 0.401 \\
        & \methodname{}
        & \textbf{14.18}
        & \textbf{0.577} & \textbf{0.336} & \textbf{0.646} & \textbf{0.336}
        & \textbf{0.608} & \textbf{0.576} & \textbf{0.608}
        & \textbf{0.527} \\

        \hdashline
        0.3
        & No clipping
        & 6049.69
        & 0.473 & 0.262 & 0.502 & 0.274 & 0.412 & 0.258 & 0.259
        & 0.348 \\
        & W/A grid search
        & 151.48
        & 0.489 & 0.220 & 0.503 & 0.234 & 0.483 & 0.303 & 0.299
        & 0.362 \\
        & \methodname{}
        & \textbf{18.28}
        & \textbf{0.552} & \textbf{0.318} & \textbf{0.608} & \textbf{0.306}
        & \textbf{0.552} & \textbf{0.524} & \textbf{0.540}
        & \textbf{0.486} \\

        \hdashline
        0.4
        & No clipping
        & 16822.71
        & 0.500 & \textbf{0.270} & 0.505 & 0.253 & 0.406 & 0.254 & 0.264
        & 0.350 \\
        & W/A grid search
        & 697.77
        & 0.488 & 0.254 & 0.500 & 0.252 & 0.429 & 0.269 & 0.270
        & 0.352 \\
        & \methodname{}
        & \textbf{30.54}
        & \textbf{0.511} & 0.260 & \textbf{0.564} & \textbf{0.265}
        & \textbf{0.492} & \textbf{0.421} & \textbf{0.436}
        & \textbf{0.421} \\

        \hdashline
        & FP16 baseline
        & 7.81
        & 0.694 & 0.408 & 0.781 & 0.463 & 0.742 & 0.721 & 0.741
        & 0.650 \\

        \hline
        \multicolumn{11}{l}{\scriptsize $^{*}$ denotes default configuration.}
    \end{tabular}
\end{table*}

\subsection{Calibration Robustness}
\label{appx:calibration-robustness}
We first vary the number of calibration samples from 8 to 128 while fixing the sequence length at 2048. We additionally report clipping-calibration time and cap each calibration run at one hour. As shown in Table~\ref{tab:calib-sample-sweep}, \methodname{} remains stable across the full sweep, with perplexity between 12.21 and 12.31 and average zero-shot accuracy between 0.565 and 0.579, while calibration time grows from 3.4 minutes at 8 samples to 10.7 minutes at 128 samples. In contrast, W/A Grid Search takes 32.9 minutes even with 8 samples and exceeds the one-hour limit for all larger calibration sets. We note that the calibration time has run-to-run variability, so the reported times can differ slightly from Table.~\ref{tab:accuracy-and-calibration-time} (e.g., 3.3 min and 3.4 min).

\begin{table*}[t]
    \centering
    \caption{LLaMA-3.2-3B results across calibration-set sizes. N/E indicates metrics not evaluated because calibration timed out; T/O marks a run exceeding the 60-minute limit.}
    \label{tab:calib-sample-sweep}
    \scriptsize
    \setlength{\tabcolsep}{2.8pt}
    \begin{tabular}{rlr:rrrrrrrr:r}
        \hline
        Samples & Method
        & PPL ($\downarrow$)
        & Wino & OBQA & PIQA & ARC-C & BoolQ & ARC-E & Hella
        & Avg. ($\uparrow$)
        & Calib. Time ($\downarrow$) \\
        \hline

        8$^{*}$
        & W/A grid search
        & 19.12
        & 0.566 & 0.294 & 0.672 & 0.317 & 0.515 & 0.528 & 0.525
        & 0.488
        & 32.9 min \\
        & \methodname{}
        & \textbf{12.27}
        & \textbf{0.622} & \textbf{0.368} & \textbf{0.722} & \textbf{0.367}
        & \textbf{0.665} & \textbf{0.630} & \textbf{0.651}
        & \textbf{0.575}
        & \textbf{3.4 min} \\

        \hdashline
        16
        & W/A grid search
        & N/E
        & N/E & N/E & N/E & N/E & N/E & N/E & N/E
        & N/E
        & $>60$ min (T/O) \\
        & \methodname{}
        & 12.21
        & 0.631 & 0.350 & 0.716 & 0.373 & 0.623 & 0.642 & 0.654
        & 0.570
        & 3.9 min \\

        \hdashline
        32
        & W/A grid search
        & N/E
        & N/E & N/E & N/E & N/E & N/E & N/E & N/E
        & N/E
        & $>60$ min (T/O) \\
        & \methodname{}
        & 12.31
        & 0.631 & 0.370 & 0.723 & 0.384 & 0.650 & 0.654 & 0.641
        & 0.579
        & 5.0 min \\

        \hdashline
        64
        & W/A grid search
        & N/E
        & N/E & N/E & N/E & N/E & N/E & N/E & N/E
        & N/E
        & $>60$ min (T/O) \\
        & \methodname{}
        & 12.31
        & 0.615 & 0.352 & 0.712 & 0.369 & 0.626 & 0.643 & 0.640
        & 0.565
        & 6.8 min \\

        \hdashline
        128
        & W/A grid search
        & N/E
        & N/E & N/E & N/E & N/E & N/E & N/E & N/E
        & N/E
        & $>60$ min (T/O) \\
        & \methodname{}
        & 12.26
        & 0.629 & 0.372 & 0.716 & 0.358 & 0.601 & 0.630 & 0.645
        & 0.565
        & 10.7 min \\

        \hdashline
        & FP16 baseline
        & 7.81
        & 0.694 & 0.408 & 0.781 & 0.463 & 0.742 & 0.721 & 0.741
        & 0.650
        & -- \\

        \hline
        \multicolumn{12}{l}{\scriptsize $^{*}$ denotes default configuration.}
    \end{tabular}
\end{table*}

We next vary the calibration sequence length from 512 to 2048 while keeping the number of samples fixed at 8. As shown in Table~\ref{tab:calib-seqlen-sweep}, \methodname{} again shows little sensitivity to this choice, maintaining perplexity around 12.2 and average zero-shot accuracy between 0.571 and 0.584. Its calibration time remains nearly constant at 3.0--3.4 minutes, whereas W/A Grid Search increases from 10.9 to 32.9 minutes as the sequence length grows. This highlights the better scaling of our analytical calibration procedure with calibration sequence length.

\begin{table*}[t]
    \centering
    \caption{LLaMA-3.2-3B results across calibration sequence lengths.}
    \label{tab:calib-seqlen-sweep}
    \scriptsize
    \setlength{\tabcolsep}{2.8pt}
    \begin{tabular}{rlr:rrrrrrrr:r}
        \hline
        Seq. length & Method
        & PPL ($\downarrow$)
        & Wino & OBQA & PIQA & ARC-C & BoolQ & ARC-E & Hella
        & Avg. ($\uparrow$)
        & Calib. Time ($\downarrow$) \\
        \hline

        512
        & W/A grid search
        & 17.72
        & 0.548 & 0.302 & 0.671 & 0.317 & 0.523 & 0.529 & 0.539
        & 0.490
        & 10.9 min \\
        & \methodname{}
        & \textbf{12.26}
        & \textbf{0.611} & \textbf{0.368} & \textbf{0.713} & \textbf{0.376}
        & \textbf{0.630} & \textbf{0.648} & \textbf{0.651}
        & \textbf{0.571}
        & \textbf{3.0 min} \\

        \hdashline
        1024
        & W/A grid search
        & 18.46
        & 0.586 & 0.310 & 0.655 & 0.300 & 0.497 & 0.530 & 0.539
        & 0.488
        & 18.4 min \\
        & \methodname{}
        & \textbf{12.20}
        & \textbf{0.638} & \textbf{0.372} & \textbf{0.731} & \textbf{0.386}
        & \textbf{0.660} & \textbf{0.653} & \textbf{0.645}
        & \textbf{0.584}
        & \textbf{3.1 min} \\

        \hdashline
        2048$^{*}$
        & W/A grid search
        & 19.12
        & 0.566 & 0.294 & 0.672 & 0.317 & 0.515 & 0.528 & 0.525
        & 0.488
        & 32.9 min \\
        & \methodname{}
        & \textbf{12.27}
        & \textbf{0.622} & \textbf{0.368} & \textbf{0.722} & \textbf{0.367}
        & \textbf{0.665} & \textbf{0.630} & \textbf{0.651}
        & \textbf{0.575}
        & \textbf{3.4 min} \\

        \hdashline
        & FP16 baseline
        & 7.81
        & 0.694 & 0.408 & 0.781 & 0.463 & 0.742 & 0.721 & 0.741
        & 0.650
        & -- \\

        \hline
        \multicolumn{12}{l}{\scriptsize $^{*}$ denotes default configuration.}
    \end{tabular}
\end{table*}

Finally, we vary the calibration domain among WikiText-2, C4 \citep{raffel2020exploring}, and Pile \citep{gao2020pile} while keeping the number of samples and sequence length fixed. As shown in Table~\ref{tab:calib-domain-sweep}, \methodname{} produces similar results across all three datasets, with perplexity between 12.27 and 12.32 and average zero-shot accuracy between 0.575 and 0.581. It also consistently outperforms W/A Grid Search across calibration domains, indicating that the calibrated clipping factors are not sensitive to the particular calibration corpus.

\begin{table*}[t]
    \centering
    \caption{LLaMA-3.2-3B results across calibration domains.}
    \label{tab:calib-domain-sweep}
    \scriptsize
    \setlength{\tabcolsep}{3.5pt}
    \begin{tabular}{llr:rrrrrrrr}
        \hline
        Domain & Method
        & PPL ($\downarrow$)
        & Wino & OBQA & PIQA & ARC-C & BoolQ & ARC-E & Hella
        & Avg. ($\uparrow$) \\
        \hline

        WikiText-2$^{*}$
        & W/A grid search
        & 19.12
        & 0.566 & 0.294 & 0.672 & 0.317 & 0.515 & 0.528 & 0.525
        & 0.488 \\
        & \methodname{}
        & \textbf{12.27}
        & \textbf{0.622} & \textbf{0.368} & \textbf{0.722} & \textbf{0.367}
        & \textbf{0.665} & \textbf{0.630} & \textbf{0.651}
        & \textbf{0.575} \\

        \hdashline
        C4
        & W/A grid search
        & 17.73
        & 0.568 & 0.316 & 0.676 & 0.306 & 0.542 & 0.546 & 0.559
        & 0.502 \\
        & \methodname{}
        & \textbf{12.32}
        & \textbf{0.627} & \textbf{0.376} & \textbf{0.720} & \textbf{0.381}
        & \textbf{0.642} & \textbf{0.643} & \textbf{0.646}
        & \textbf{0.576} \\

        \hdashline
        Pile
        & W/A grid search
        & 17.88
        & 0.566 & 0.316 & 0.655 & 0.331 & 0.553 & 0.555 & 0.552
        & 0.504 \\
        & \methodname{}
        & \textbf{12.31}
        & \textbf{0.623} & \textbf{0.362} & \textbf{0.729} & \textbf{0.373}
        & \textbf{0.685} & \textbf{0.647} & \textbf{0.652}
        & \textbf{0.581} \\

        \hdashline
        & FP16 baseline
        & 7.81
        & 0.694 & 0.408 & 0.781 & 0.463 & 0.742 & 0.721 & 0.741
        & 0.650 \\

        \hline
        \multicolumn{11}{l}{\scriptsize $^{*}$ denotes default configuration.}
    \end{tabular}
\end{table*}

\subsection{Optimization Convergence}
\label{appx:optimization-convergence}

\begin{figure*}[t]
    \centering
    \includegraphics[width=\textwidth]{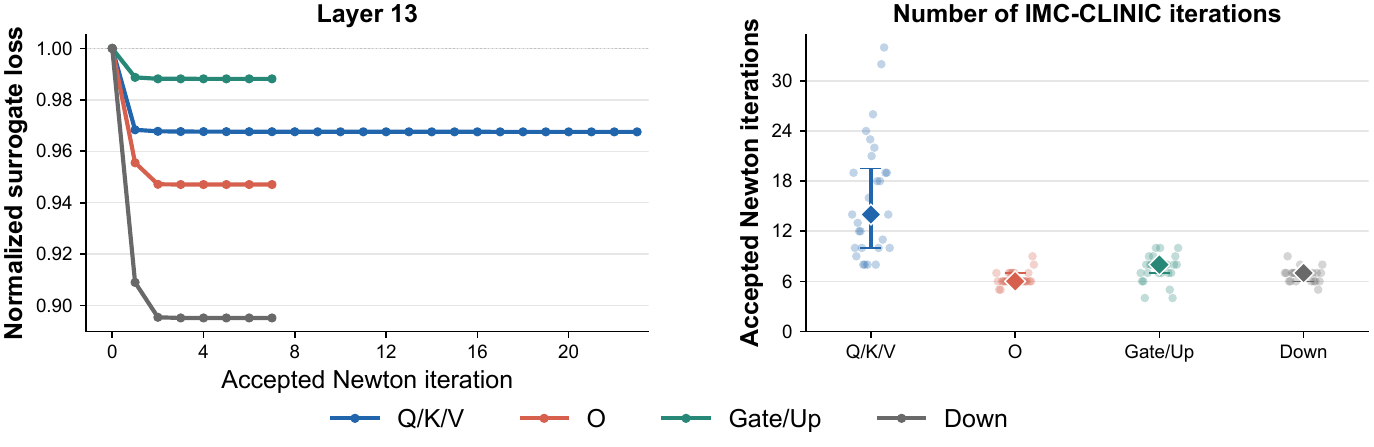}
    \caption{Optimization convergence of \methodname{} on LLaMA-3.2-3B. \textbf{Left:} Surrogate loss versus accepted Newton iteration for decoder block 13, normalized by the loss at the selected initialization, $\mathcal{L}^{(t)}/\mathcal{L}^{(0)}$. \textbf{Right:} Number of accepted Newton iterations at optimizer return across all 28 decoder blocks. Faint points denote individual decoder blocks, diamonds denote the median, and vertical ranges indicate the 25th--75th percentiles.}
    \label{fig:optimization-convergence}
\end{figure*}

Figure~\ref{fig:optimization-convergence} characterizes the convergence of \methodname{} on LLaMA-3.2-3B. The normalized surrogate loss decreases sharply within the first few accepted Newton iterations and quickly reaches a plateau across all optimization groups (Fig.~\ref{fig:optimization-convergence}, left). Across the 28 decoder blocks, the median number of accepted iterations until optimizer return ranges from approximately $6$ to $14$ across projection groups (Fig.~\ref{fig:optimization-convergence}, right). These results show that the safeguarded Newton updates reach near-final loss values rapidly in practice.

\subsection{Method Ablations}
\label{appx:method-ablations}

We ablate the main components of \methodname{} on LLaMA-3.2-3B under the standard 9-bit ADC setting. All variants use the same calibration data, quantization configuration, and default initialization. For \textbf{w/o ADC term}, we remove $\mathcal{L}_{\mathrm{ADC}}$ from the calibration objective while retaining ADC quantization during final IMC inference. For \textbf{w/o signed-bias term}, we remove $\mathcal{L}_{\mathrm{bias}}$ while keeping the remaining objective unchanged. For \textbf{Independent W/A}, activation clipping $(\gamma,\beta)$ is optimized with $\alpha=1$, and weight clipping $\alpha$ is optimized separately with $\gamma=\beta=1$; the independently obtained clipping factors are then combined for inference.

\begin{table}[t]
    \centering
    \caption{Method ablations on LLaMA-3.2-3B under the 9-bit ADC setting. All variants use the same default initialization, and ADC quantization remains enabled during inference.}
    \label{tab:method-ablations}
    \small
    \begin{tabular}{lccc r}
        \hline
        Method & Joint W/A & Signed bias & ADC term & PPL ($\downarrow$) \\
        \hline
        \methodname{}             & \checkmark & \checkmark & \checkmark & \textbf{12.27} \\
        w/o ADC term              & \checkmark & \checkmark &            & 127.91 \\
        w/o signed-bias term      & \checkmark &            & \checkmark & 12.76 \\
        Independent W/A           &            & \checkmark & \checkmark & 15.80 \\
        \hline
    \end{tabular}
\end{table}

As shown in Table~\ref{tab:method-ablations}, removing the ADC term causes a severe degradation in perplexity, demonstrating that clipping must explicitly account for ADC quantization under this hardware setting. Removing the signed-bias term also degrades perplexity, while independently optimizing activation and weight clipping increases PPL to 15.80. These results support all three components of the proposed objective: ADC-aware calibration, signed-bias modeling, and joint activation--weight clipping optimization.

\subsection{Per-Channel Weight Clipping}
\label{appx:per-channel-alpha}

Our main experiments use a shared scalar weight-clipping factor $\alpha$ per projection to provide a controlled and compact setting for analyzing the clipping objective. However, \methodname{} is not restricted to this granularity: the analytical surrogate and safeguarded Newton updates naturally extend to a larger set of clipping variables. We therefore evaluate a per-channel variant in which each output channel has its own weight-clipping factor $\alpha_o$, while the activation clipping factors remain shared as in the default configuration. As shown in Table~\ref{tab:alpha-granularity}, the finer-grained parameterization provides a modest improvement in both perplexity and average downstream accuracy, indicating that \methodname{} can also benefit from more granular quantization schemes.

\begin{table*}[t]
    \centering
    \caption{LLaMA-3.2-3B results across weight-clipping granularities.}
    \label{tab:alpha-granularity}
    \scriptsize
    \setlength{\tabcolsep}{3.5pt}
    \begin{tabular}{lr:rrrrrrrr}
        \hline
        Method
        & PPL ($\downarrow$)
        & Wino & OBQA & PIQA & ARC-C & BoolQ & ARC-E & Hella
        & Avg. ($\uparrow$) \\
        \hline

        \methodname{} + per-channel $\alpha$
        & \textbf{12.02}
        & 0.617 & \textbf{0.370} & 0.721 & \textbf{0.393}
        & \textbf{0.668} & \textbf{0.649} & 0.650
        & \textbf{0.581} \\


        \methodname{} (shared scalar $\alpha$)
        & 12.27
        & \textbf{0.622} & 0.368 & \textbf{0.722} & 0.367
        & 0.665 & 0.630 & \textbf{0.651}
        & 0.575 \\

        \hline
    \end{tabular}
\end{table*}

\subsection{Empirical-Fisher Weighting for Shared-Input Projections}
\label{appx:weighted-loss-input-divergent-projections}

For projections that share the same input activation, such as the $q/k/v$ and $up/gate$ branches, our default calibration objective simply sums their surrogate losses, assigning equal weight to each branch. We additionally evaluate whether weighting the branches by their task sensitivity improves calibration. We estimate this sensitivity using empirical Fisher (EF) information, computed from squared gradients of the task loss with respect to each branch output, and aggregate the resulting values into a scalar importance $F_m$ for branch $m$. We then optimize
\begin{equation*}
\mathcal{L}_{\mathrm{group}}^{\mathrm{EF}}(\boldsymbol{\theta})
=
\sum_{m=1}^{M}
\omega_m
\mathcal{L}_m(\gamma,\beta,\alpha_m),
\qquad
\omega_m
=
\frac{F_m}{\sum_{j=1}^{M} F_j}.
\end{equation*}
Thus, branches that are more sensitive to the task loss contribute more strongly when calibrating the shared activation clipping factors.

As shown in Table~\ref{tab:ef-weighted-loss}, empirical-Fisher weighting improves WikiText-2 perplexity from 12.27 to 11.88, but slightly decreases average zero-shot accuracy from 0.575 to 0.568. Since the improvement does not consistently transfer across downstream tasks, we retain uniform weighting as the default setting, which also keeps the calibration objective and its analysis simpler.

\begin{table*}[t]
    \centering
    \caption{LLaMA-3.2-3B results with empirical-Fisher weighting for input-divergent projections.}
    \label{tab:ef-weighted-loss}
    \scriptsize
    \setlength{\tabcolsep}{3.5pt}
    \begin{tabular}{lr:rrrrrrrr}
        \hline
        Method
        & PPL ($\downarrow$)
        & Wino & OBQA & PIQA & ARC-C & BoolQ & ARC-E & Hella
        & Avg. ($\uparrow$) \\
        \hline

        \methodname{}+EF-weighted loss
        & \textbf{11.88}
        & 0.619 & 0.364 & 0.716 & \textbf{0.370}
        & 0.622 & \textbf{0.631} & \textbf{0.657}
        & 0.568 \\


        \methodname{} (unweighted loss)
        & 12.27
        & \textbf{0.622} & \textbf{0.368} & \textbf{0.722} & 0.367
        & \textbf{0.665} & 0.630 & 0.651
        & \textbf{0.575} \\

        \hline
    \end{tabular}
\end{table*}

\section{Optimality Validation}
\label{appx:optimality-validation}

\begin{figure}[h]
    \centering
    \includegraphics[width=0.82\textwidth]{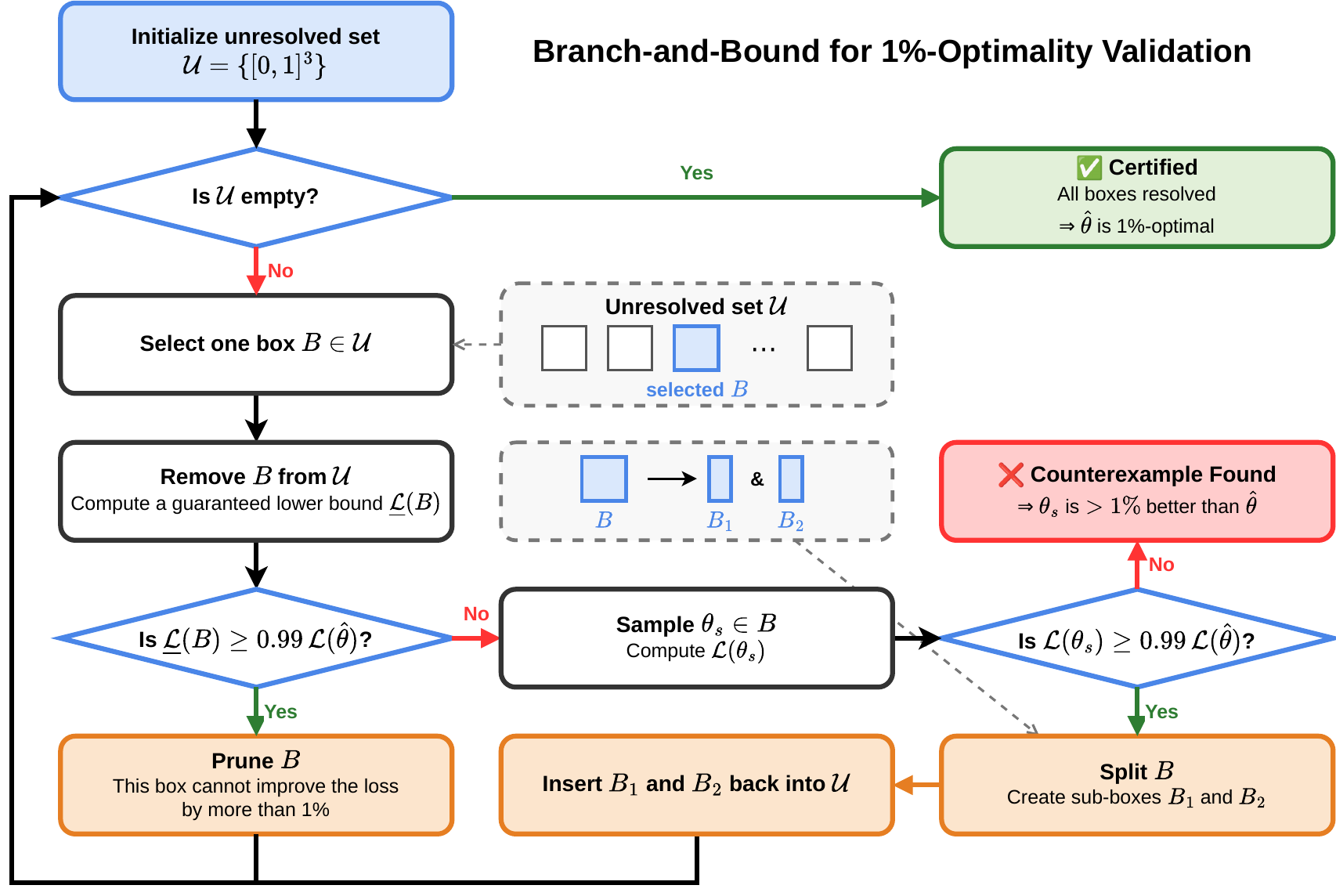}
    \caption{Branch-and-bound procedure for validating global $1\%$-optimality of the calibrated clipping factors under the surrogate objective.}
    \label{fig:optimality-validation}
\end{figure}

\subsection{Optimality Criterion and Certification Strategy}
\label{appx:optimality-criterion}

We investigate whether the analytical surrogate in Eq.~\eqref{eq:surrogate-objective} yields a sufficiently benign optimization landscape for \methodname{} to reach a near-global solution.

One sufficient, but not necessary, route to certifying global optimality of the candidate would be to establish convexity of the objective over the feasible domain, for example by verifying that its Hessian is positive semidefinite throughout the domain \citep{ConvexOptimization}. Alternatively, verified root-finding methods such as interval Newton and the Krawczyk operator can exclude or certify stationary points by applying interval arithmetic to the gradient and Hessian \citep{IntervalAnalysis}. Both routes, however, require reliable global or box-wise enclosures of the full Hessian. In our objective, several second-order terms depend on activation and weight densities evaluated at moving clipping boundaries, making such enclosures difficult to obtain from finite calibration data.

Rather than pursuing an exact global-optimality proof, we construct a numerical certificate of $\epsilon$-optimality by lower-bounding the surrogate objective over the entire feasible domain and verifying that no feasible clipping configuration can improve upon the solution found by \methodname{} by more than $\epsilon$.

For each projection, let $\widehat{\boldsymbol{\theta}}=(\widehat{\gamma},\widehat{\beta},\widehat{\alpha})\in\Theta=[0,1]^3$ denote the clipping factors returned by \methodname{}, let $U=\mathcal{L}(\widehat{\boldsymbol{\theta}})$, and define the unknown global minimum as
\begin{equation*}
\mathcal{L}^{\star}
=
\min_{\boldsymbol{\theta}\in\Theta}\mathcal{L}(\boldsymbol{\theta}).
\end{equation*}
We call $\widehat{\boldsymbol{\theta}}$ $\epsilon$-optimal if
\begin{equation}
\frac{U-\mathcal{L}^{\star}}{U}\leq\epsilon
\qquad\Longleftrightarrow\qquad
\mathcal{L}^{\star}\geq(1-\epsilon)U.
\label{eq:epsilon-optimality}
\end{equation}
Throughout our validation, $\epsilon=1\%$, so it suffices to prove that no feasible clipping configuration achieves an objective value below $0.99U$.

We certify this condition using branch-and-bound~\citep{lawler1966branch}. We partition $\Theta$ into parameter boxes $B$ and construct, for each box, a valid lower bound
\begin{equation*}
\underline{\mathcal{L}}(B)
\leq
\min_{\boldsymbol{\theta}\in B}\mathcal{L}(\boldsymbol{\theta}).
\end{equation*}
Any box satisfying $\underline{\mathcal{L}}(B)\geq0.99U$ can therefore be safely pruned. If the complete feasible domain can be covered by such pruned boxes, the candidate is certified to be $1\%$-optimal. The next subsection derives the box lower bounds used for this certification.

\subsection{Dependency-Preserving Lower Bounds}
\label{appx:box-lower-bounds}
We perform post-hoc certification independently for each projection, so the validation problem has three clipping variables, \(\boldsymbol{\theta}=(\gamma,\beta,\alpha)\). Although calibration shares activation clipping factors across projection groups, this per-projection validation imposes a stricter requirement: if the shared calibrated factors are within $1\%$ of the optimum for every constituent projection, then the summed grouped objective is also within $1\%$ of its optimum. We use this formulation because it reduces the certification dimensionality and substantially lowers branch-and-bound validation cost.

For a parameter box \(B=[\gamma_L,\gamma_U]\times[\beta_L,\beta_U]\times[\alpha_L,\alpha_U]\), our goal is to construct a guaranteed lower bound \(\underline{\mathcal L}(B)\leq \min_{\boldsymbol{\theta}\in B}\mathcal L(\boldsymbol{\theta})\). We follow the surrogate decomposition in Eq.~\eqref{eq:surrogate-objective},

\begin{equation*}
\mathcal L = \mathcal L_{\mathrm{diag}} + \mathcal L_{\mathrm{bias}} + \mathcal L_{\mathrm{ADC}}.
\end{equation*}

and lower-bound each component separately before combining them into a bound for the complete objective. Importantly, these terms share the same clipping factors: independently minimizing their intermediate quantities could implicitly use inconsistent values of \((\gamma,\beta,\alpha)\) and produce an unnecessarily loose bound. We therefore preserve their dependence on the shared clipping variables until the component-wise bounds are combined.

\paragraph{Operand Quantization Error Bounds.} We first consider the activation contribution to the diagonal loss,
\begin{equation*}
\mathcal{L}_{\mathrm{diag},x}(\gamma,\beta)
=
\sum_i w_i^2
\left(
\mathbb{E}[e_{x,\mathrm{round}}^2]
+
\mathbb{E}[e_{x,\mathrm{clip},i}^2]
\right).
\end{equation*}
This term is convex in $(\gamma,\beta)$: the rounding component is quadratic in the retained activation range, while the clipping component is convex in the clipping thresholds. Therefore, letting $\boldsymbol{\phi}=(\gamma,\beta)$, the tangent plane at any reference point $\boldsymbol{\phi}_r$ is a global lower bound,
\begin{equation*}
\mathcal{L}_{\mathrm{diag},x}(\boldsymbol{\phi})
\geq
\mathcal{L}_{\mathrm{diag},x}(\boldsymbol{\phi}_r)
+
\nabla_{\boldsymbol{\phi}}
\mathcal{L}_{\mathrm{diag},x}(\boldsymbol{\phi}_r)^\top
(\boldsymbol{\phi}-\boldsymbol{\phi}_r).
\end{equation*}
We retain this tangent plane and combine it with the bounds of the remaining loss terms before minimizing over the box. For the weight contribution, changes between rounded and clipped regimes make the elementwise error non-smooth in $\alpha$, so we instead use a valid interval lower bound over $[\alpha_L,\alpha_U]$.

\paragraph{Accumulated Bias Error Bound.} The bias term is more involved because the signed MAC error depends jointly on activation and weight clipping. Let
\begin{equation*}
\widetilde{x}_i(\gamma,\beta)
=
\operatorname{clip}
\!\left(
x_i,c_{x,\mathrm{down}},c_{x,\mathrm{up}}
\right),
\qquad
\widetilde{w}_i(\alpha)
=
\operatorname{clip}
\!\left(
w_i,-c_w,c_w
\right)
\end{equation*}
denote the activation and weight after clipping but before rounding. The signed-error expression in Eq.~\eqref{eq:mac-signed-mean} can equivalently be rewritten as
\begin{equation*}
\mathbb{E}[\delta y_i]
=
\widetilde{w}_i\,\mathbb{E}[\widetilde{x}_i]
-
w_i\mathbb{E}[x_i],
\end{equation*}
since $\widetilde{x}_i=x_i+e_{x,\mathrm{clip},i}$ and $\widetilde{w}_i=w_i+e_{w,\mathrm{clip},i}$. Thus, bounding $\mathbb{E}[\delta y_i]$ reduces to bounding the coupled product $\widetilde{w}_i\mathbb{E}[\widetilde{x}_i]$.

We first construct affine upper and lower bounds for the two factors. For a convex function, a tangent is a lower bound and a secant is an upper bound; for a concave function, the roles are reversed,
\begin{equation*}
\text{convex:}\quad
\operatorname{tan} f \leq f \leq \operatorname{sec} f,
\qquad
\text{concave:}\quad
\operatorname{sec} f \leq f \leq \operatorname{tan} f.
\end{equation*}
The positive contribution to $\mathbb{E}[\widetilde{x}_i]$ is concave in $\gamma$, while the negative contribution is convex in $\beta$; clipped weights are similarly concave or convex in $\alpha$ depending on their sign. This gives affine envelopes
\begin{equation*}
\ell_i^x(\gamma,\beta)
\leq
\mathbb{E}[\widetilde{x}_i]
\leq
u_i^x(\gamma,\beta),
\qquad
\ell_i^w(\alpha)
\leq
\widetilde{w}_i
\leq
u_i^w(\alpha).
\end{equation*}

We then propagate these bounds through the product using McCormick envelopes~\citep{mccormick1976computability}. For example, if $a\in[a_L,a_U]$ and $b\in[b_L,b_U]$, two valid affine lower bounds are
\begin{equation*}
ab
\geq
a_L b+b_L a-a_Lb_L,
\qquad
ab
\geq
a_U b+b_U a-a_Ub_U.
\end{equation*}
These planes lie below the bilinear surface $ab$ throughout the box. Applying them to $\widetilde{w}_i\mathbb{E}[\widetilde{x}_i]$ therefore gives affine bounds on $\mathbb{E}[\delta y_i]$ while retaining its dependence on the shared clipping factors.

Finally, we propagate these bounds through
\begin{equation*}
\mathcal{L}_{\mathrm{bias}}
=
\left(
\sum_i \mathbb{E}[\delta y_i]
\right)^2
-
\sum_i \mathbb{E}[\delta y_i]^2.
\end{equation*}
The negative squares are lower-bounded by secants of the concave function $-z^2$, while the positive square is lower-bounded using $s^2\geq 2\rho s-\rho^2$. This yields an affine lower bound for $\mathcal{L}_{\mathrm{bias}}$ in $(\gamma,\beta,\alpha)$.

\paragraph{ADC Error Bound.} The ADC term in Eq.~\eqref{eq:adc-loss} also couples activation and weight clipping, but has a simpler structure than the bias term:
\begin{equation*}
\mathcal{L}_{\mathrm{ADC}}
=
K\frac{\Delta_{\mathrm{ADC}}^2}{12}(s_xs_w)^2.
\end{equation*}
Substituting the activation and weight scales, its clipping-dependent part is proportional to
\begin{equation*}
\left[
\frac{1}{T}\sum_{t=1}^{T}
\left(\gamma M_t^+-\beta M_t^-\right)^2
\right]\alpha^2,
\end{equation*}
where $T$ is the number of activation samples in the calibration set, and all remaining factors are fixed by the quantization and hardware configuration. The activation-range term and $\alpha^2$ are both nonnegative and convex, so we lower-bound each using a tangent plane. Their product remains nonlinear, and we therefore apply the same McCormick relaxation as above to obtain an affine lower bound for $\mathcal{L}_{\mathrm{ADC}}$ in $(\gamma,\beta,\alpha)$.

\paragraph{Combining the Box Bounds.} After constructing lower bounds for $\mathcal{L}_{\mathrm{diag}}$, $\mathcal{L}_{\mathrm{bias}}$, and $\mathcal{L}_{\mathrm{ADC}}$, we combine them before minimizing over the parameter box. Because each component has been relaxed to an affine function of the same clipping variables, their sum has the form
\begin{equation*}
\ell(\gamma,\beta,\alpha)
=
a_0+a_\gamma\gamma+a_\beta\beta+a_\alpha\alpha,
\end{equation*}
with
\begin{equation*}
\ell(\gamma,\beta,\alpha)
\leq
\mathcal{L}(\gamma,\beta,\alpha),
\qquad
\forall\,(\gamma,\beta,\alpha)\in B.
\end{equation*}
Since $\ell$ is affine and $B$ is a rectangular box, its minimum is attained at a corner: for each clipping factor, we choose the lower endpoint if its coefficient is nonnegative and the upper endpoint otherwise. This gives the guaranteed box lower bound
\begin{equation*}
\underline{\mathcal{L}}(B)
=
\min_{\boldsymbol{\theta}\in B}
\ell(\boldsymbol{\theta}),
\end{equation*}
which can be evaluated exactly and is then used by the branch-and-bound procedure.


\subsection{Adaptive Branch-and-Bound Certification}
\label{appx:branch-and-bound}

Figure~\ref{fig:optimality-validation} illustrates the adaptive branch-and-bound certification procedure. We maintain a set $\mathcal{U}$ of unresolved boxes, initialized as $\mathcal{U}=\{\Theta\}$, and process the box with the smallest current lower bound. If $\underline{\mathcal{L}}(B)\geq0.99U$, the box is safely pruned because it cannot contain a solution that improves upon the calibrated objective by more than $1\%$.

Otherwise, we evaluate a point within $B$ to search for a counterexample. If its objective is below $0.99U$, the calibrated solution is not $1\%$-optimal. If no counterexample is found, we bisect $B$ along its longest dimension and add the two child boxes to $\mathcal{U}$. This process continues until either a counterexample is found or no unresolved boxes remain.

For each child $B'\subseteq B$, any valid lower bound for the parent remains valid:
\begin{equation*}
\min_{\boldsymbol{\theta}\in B'}\mathcal{L}(\boldsymbol{\theta})
\geq
\min_{\boldsymbol{\theta}\in B}\mathcal{L}(\boldsymbol{\theta}).
\end{equation*}
We therefore strengthen each newly computed child bound by
\begin{equation*}
\underline{\mathcal{L}}(B')
\leftarrow
\max\!\left\{
\underline{\mathcal{L}}(B'),
\underline{\mathcal{L}}(B)
\right\}.
\end{equation*}
This monotonic inheritance prevents the lower bound from weakening during refinement.

Throughout the procedure, the union of pruned and unresolved boxes remains a complete cover of the feasible domain $\Theta$. Hence, when $\mathcal{U}=\emptyset$, every feasible point belongs to a box certified to have objective at least $0.99U$, establishing $1\%$-optimality.

\subsection{Certificate Validation and Results}
\label{appx:optimality-results}

\paragraph{Numerical Safeguards.} This test is performed in double precision. To make the certification decisions conservative under floating-point evaluation, we expand relevant interval endpoints outward using \texttt{nextafter}, reduce each computed lower bound by $10^{-10}$ times its numerical scale, and increase candidate and feasible upper values by an absolute $10^{-10}$. These conservative quantities are used for pruning, stopping, relative-gap tests, and counterexample detection. We additionally audit the adopted margins against extended-precision recomputation.

\paragraph{Certificate Validation.} We perform several independent numerical checks on the resulting branch-and-bound certificates. We verify that the terminal boxes provide a complete, non-overlapping cover of the feasible domain, recompute the saved box lower bounds, and reconstruct the final domain-wide certificate from the saved artifacts. We additionally sample points within terminal boxes and confirm that direct objective evaluations do not fall below their corresponding conservative lower bounds. These checks validate both the domain bookkeeping and the numerical implementation of the lower-bound procedure.

\paragraph{Optimality Results.} Table~\ref{tab:optimality-validation} summarizes the validation results for LLaMA-3.2-3B and Qwen3-4B. All evaluated projections are certified to be within $1\%$ of the global minimum of the calibration objective, with no counterexamples or unresolved regions. For LLaMA-3.2-3B, this covers all $28$ decoder blocks and all seven projection types, for a total of $196/196$ projection-level optimization problems. The same validation covers all evaluated projections of Qwen3-4B.

\begin{table}[t]
\centering
\caption{Optimality-validation results. Each entry under a projection type reports the number of projection-level clipping solutions certified to be within $1\%$ of the global minimum of the calibration objective. Validation time is the median wall-clock time per projection and is incurred only for the optimality check, not during clipping calibration.}
\label{tab:optimality-validation}
\resizebox{\linewidth}{!}{
\begin{tabular}{lccccccccc}
\toprule
Model
& Q & K & V & O & Gate & Up & Down
& Total & Median Time \\
\midrule
LLaMA-3.2-3B
& 28/28 & 28/28 & 28/28 & 28/28
& 28/28 & 28/28 & 28/28
& 196/196 & 16.88 min \\
Qwen3-4B
& 36/36 & 36/36 & 36/36 & 36/36
& 36/36 & 36/36 & 36/36
& 252/252 & 18.58 min \\
\bottomrule
\end{tabular}}
\end{table}

The validation time should be distinguished from the calibration cost reported in the main experiments: branch-and-bound certification is performed only as an offline optimality check and is not required to obtain or deploy the clipping factors. These results therefore provide evidence that the clipping factors found by \methodname{} are consistently near-global solutions of the calibration objective across different projection types and model families.

\end{document}